\documentclass[runningheads]{llncs}
\usepackage[utf8]{inputenc}
\usepackage{graphicx}
\usepackage{comment}
\usepackage{mathtools}
\usepackage{amsmath,amssymb} 
\usepackage{xcolor}
\usepackage{mynotation}
\usepackage{algorithm}
\usepackage[noend]{algpseudocode}
\usepackage{booktabs}
\usepackage[caption=false]{subfig}
\usepackage{pifont}
\usepackage{arydshln}
\usepackage{verbatim}
\usepackage[width=122mm,left=12mm,paperwidth=146mm,height=193mm,top=12mm,paperheight=217mm]{geometry}

\algrenewcommand{\algorithmicrequire}{\textbf{Input:}}
\algrenewcommand{\algorithmicensure}{\textbf{Output:}}

\newcommand{\algcomment}[1]{\hspace{3mm}{\footnotesize\# #1}}

\newcommand{\parsection}[1]{\vspace{0.5mm}\noindent\textbf{#1:}~}

\begin{document}
\pagestyle{headings}
\mainmatter
\title{Learning What to Learn for Video Object Segmentation} 


\newcommand{\asep}{\hspace{6mm}}
\newcommand{\aand}{\hspace{3mm}}

\author{Goutam Bhat$^{*,1}$ \aand Felix Järemo Lawin$^{*,2}$ \aand Martin Danelljan$^{1}$ \\ Andreas Robinson$^{2}$ \aand Michael Felsberg$^2$ \aand Luc Van Gool$^{1}$ \aand Radu Timofte$^{1}$ \vspace{1.5mm}}
\institute{$^1$ CVL, ETH Z\"urich, Switzerland \asep $^2$ CVL, Link\"oping University, Sweden}
\authorrunning{Bhat, Lawin, Danelljan, Robinson, Felsberg, Van Gool, Timofte}

\maketitle

\begin{abstract}
Video object segmentation (VOS) is a highly challenging problem, since the target object is only defined during inference with a given first-frame reference mask. The problem of how to capture and utilize this limited target information remains a fundamental research question. We address this by introducing an end-to-end trainable VOS architecture that integrates a differentiable few-shot learning module. This internal learner is designed to predict a powerful parametric model of the target by minimizing a segmentation error in the first frame.
We further go beyond standard few-shot learning techniques by learning what the few-shot learner should learn. This allows us to achieve a rich internal representation of the target in the current frame, significantly increasing the segmentation accuracy of our approach. 
We perform extensive experiments on multiple benchmarks. Our approach sets a new state-of-the-art on the large-scale YouTube-VOS 2018 dataset by achieving an overall score of $81.5$, corresponding to a $2.6\%$ relative improvement over the previous best result.

\end{abstract}
{\let\thefootnote\relax\footnote{{$^*$Both authors contributed equally.}}}

\section{Introduction}

\begin{figure}[t]
    \centering%
    \includegraphics[width=\columnwidth]{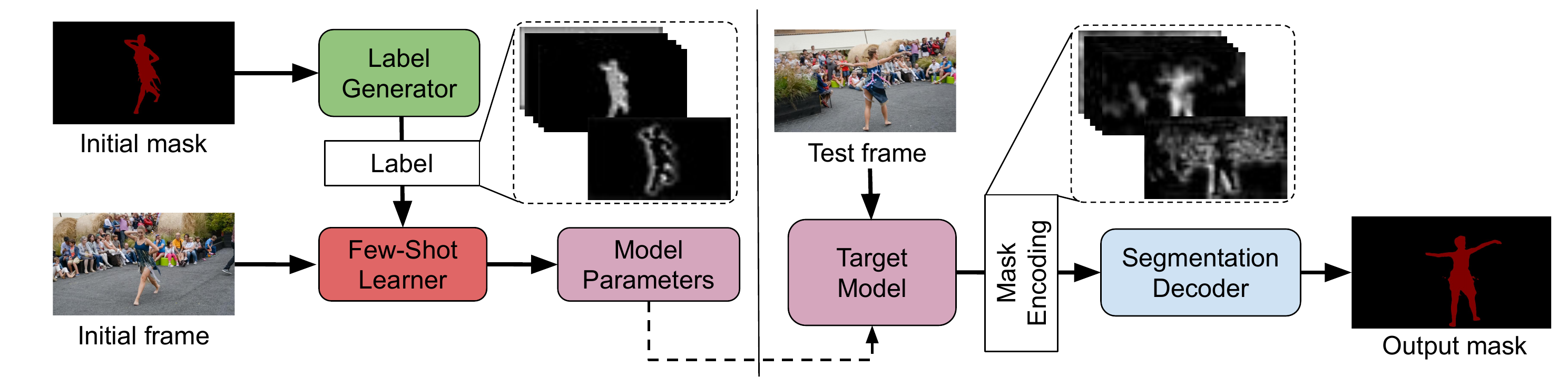}\vspace{-4mm}
    \caption{An overview of our VOS approach. Given the annotated first frame, our few-shot learner constructs a target model that outputs an encoding of the target mask (left). In subsequent test frames, the mask encoding predicted by the target model is utilized by the segmentation decoder to generate the final segmentation (right). Importantly, our approach learns how to generate the ground-truth labels for the few-shot learner. This allows the target model to output rich mask encoding (top-right).}\vspace{-2mm}
    \label{fig:internal}
\end{figure}

Semi-supervised Video Object Segmentation (VOS) is the problem of performing pixels-wise classification of a set of target objects in a video sequence. With numerous applications in e.g.\ autonomous driving~\cite{saleh2016kangaroo,ros2015vision}, surveillance~\cite{cohen1999detecting,erdelyi2014adaptive} and video editing~\cite{oh2018rgmp}, it has received significant attention in recent years. VOS is an extremely challenging problem, since the target objects are only defined by a reference segmentation in the first video frame, with no other prior information assumed. The VOS method therefore must utilize this very limited information about the target in order to perform segmentation in the subsequent frames. 
In this work, we therefore address the key research question of how to capture the scarce target information in the video.

While most state-of-the-art VOS approaches employ similar image feature extractors and segmentation heads, the advances in how to capture and utilize target information has led to much improved performance~\cite{johnander2018generative,voigtlaender2018feelvos,oh2019video,robinson2020learning}. A promising direction is to employ feature matching techniques \cite{hu2018videomatch,johnander2018generative,voigtlaender2018feelvos,oh2019video} in order to compare the reference frame with new images to segment. Such feature feature-matching layers greatly benefit from their efficiency and differentiability. This allows the design of fully end-to-end trainable architectures, which has been shown to be important for segmentation performance~\cite{johnander2018generative,voigtlaender2018feelvos,oh2019video}. On the other hand, feature matching relies on a powerful and generic feature embedding, which may limit its performance in challenging scenarios. In this work, we instead explore an alternative direction.

We propose an approach to capture the target object information in a compact parametric model. 
To this end, we integrate a differentiable few-shot learner module, which predicts the target model parameters using the first frame annotation. Our learner is designed to explicitly optimize an error between target model prediction and a ground truth label, which ensures a powerful model of the target object. Given a new frame, our target model predicts an intermediate representation of the target mask, which is input to the segmentation decoder to generate the final prediction. By employing an efficient and differentiable few-shot learner, our approach learns a robust target model using limited annotation, while being end-to-end trainable.

We further address the problem of what the internal few-shot learner should learn about the target.
The standard optimization-based few-shot learning strategy forces the target model to only learn to generate an object mask output. However, directly learning to predict the segmentation mask from a single sample is difficult. More importantly, this approach limits the target-specific information sent to the segmentation network to be a single channel mask. To address this important issue, we further propose to \emph{learn what to learn}. That is, our approach learns the ground-truth labels used by the few-shot learner to train the target model. This enables our segmentation network to learn a rich target representation, which is encoded by the learner and predicted by the target model in novel frames. Furthermore, in order to guide the learner to focus on the most crucial aspect of the target, we also learn to predict spatial importance weights for different elements in the few-shot learning loss. Since our optimization-based learner is differentiable, all modules in our architecture can be trained end-to-end by maximizing segmentation accuracy on annotated VOS videos. An overview of our video object segmentation approach is shown in Fig.\ \ref{fig:internal}.

\parsection{Contributions} Our main contributions are listed as follows. \textbf{(i)} We propose a novel VOS architecture, based on an optimization-based few-shot learner. \textbf{(ii)} We go beyond standard few-shot learning approaches, to learn what the learner should learn in order to maximize segmentation accuracy. \textbf{(iii)} Our learner predicts the target model parameters in an efficient and differentiable manner, enabling end-to-end training. \textbf{(iv)} We utilize our learned mask representation to design a light-weight bounding box initialization module, allowing our approach to generate target segmentations masks in the weakly supervised setting.

We perform comprehensive experiments on the YouTube-VOS \cite{xu2018youtube2} and DAVIS \cite{perazzi2016davis} benchmarks. Our approach sets a new state-of-the-art on the large-scale YouTube-VOS 2018 dataset, achieving an overall score of $81.5$ ($+2.6\%$ relative). We further provide detailed ablative analyses, showing the impact of each component in the proposed method.

\section{Related Work}
In recent years, progress within video object segmentation has surged, leading to rapid performance improvements. Benchmarks such as DAVIS \cite{perazzi2016davis} and YouTube-VOS \cite{xu2018youtube2} have had a significant impact on this development.

\parsection{Target Models in VOS}
Early works mainly adapted semantic segmentation networks to the VOS task through online fine-tuning~\cite{perazzi2017masktrack,caelles2017osvos,maninis2017osvos_s,xu2018youtube}. However, this strategy easily leads to overfitting to the initial target appearance and impractically long run-times.
More recent methods \cite{vondrick2018tracking,hu2018videomatch,voigtlaender2018feelvos,oh2018rgmp,wang2019ranet,oh2019video,lin2019agss} therefore integrate target-specific appearance models into the segmentation architecture. In addition to improved run-times, many of these methods can also benefit from full end-to-end learning, which has been shown to have a crucial impact on performance~\cite{voigtlaender2018feelvos,johnander2018generative,oh2019video}. Generally, these works train a target-agnostic segmentation network that is conditioned on a target model. The latter captures information about the target object, deduced from the initial image-mask pair. The generated target-aware representation is then provided to the target-agnostic segmentation network, which outputs the final prediction. 
Crucially, in order to achieve end-to-end training of the entire network, the target model needs to be differentiable.

While most VOS methods share similar feature extractors and segmentation heads, several different strategies for encoding and exploiting the target information have been proposed. In RGMP~\cite{oh2018rgmp}, a representation of the target is generated by encoding the reference frame along with its ground-truth target mask. This representation is then concatenated with the current-frame features, before being input to the segmentation head. The approach \cite{lin2019agss} extends RGMP to jointly process multiple targets using an instance specific attention generator. In \cite{johnander2018generative}, a light-weight generative model is learned from embedded features corresponding to the initial target labels. The generative model is then used to classify features from the incoming frames. The target models in \cite{hu2018videomatch,voigtlaender2018feelvos} directly store foreground features and classify pixels in the incoming frames through feature matching. The recent STM approach~\cite{oh2019video} performs feature matching within a space-time memory network. It implements a read operation, which retrieves information from the encoded memory through an attention mechanism. This information is then sent to the segmentation network to predict the target mask. The method \cite{wang2019ranet} predicts template correlation filters given the input target mask. Target classification is then performed by applying the correlation filters on the the test frame. Lastly, the recent method~\cite{robinson2020learning} trains a target model consisting of a two-layer segmentation network using the Conjugate Gradient method.

\parsection{Meta-learning for VOS}
Since the VOS task itself includes a few-shot learning problem, it can be addressed with techniques developed for meta-learning~\cite{finn2017model,bertinetto2018metalearning,lee2019meta}. A few recent attempts follow this direction~\cite{liu2019meta,behl2018meta}. The method~\cite{behl2018meta} learns a classifier using k-means clustering of segmentation features in the train frame.  In~\cite{liu2019meta}, the final layer of a segmentation network is predicted by closed-form ridge regression~\cite{bertinetto2018metalearning}, using the reference example pair.
Meta-learning based techniques have been more commonly adopted in the related field of visual tracking~\cite{park2018meta,choi2019deep,bhat2019learning}. The method in~\cite{park2018meta} performs gradient based adaptation to the current target, while~\cite{choi2019deep} learns a target specific feature space online which is combined with a Siamese-based matching network. The recent work in~\cite{bhat2019learning} propose an optimization-based meta-learning strategy, where the target model directly generates the output classifications scores.
In contrast to these previous approaches, we integrate a differentiable optimization-based few-shot learner to capture target information for the VOS problem. Furthermore, we go beyond standard few-shot and meta-learning techniques by learning what the target model should learn in order to generate accurate segmentations.

\section{Method}
In this section, we present our method for video object segmentation (VOS). First, we describe our few-shot learning formulation for VOS in Sec.~\ref{sec:vos}. In Sec.~\ref{sec:learn_wtl} we then describe our approach to learn what the few-shot learner should learn. Sec.~\ref{sec:meta_learning} details our target module and the internal few-shot learner. Our segmentation architecture is described next in Sec.~\ref{sec:vos_arch}. The inference and training procedures are detailed in Sec.~\ref{sec:inference} and \ref{sec:training_details}, respectively. Finally, Sec.~\ref{sec:bbinit} describes how our approach can be easily extended to perform VOS with only a bounding box initialization.

\subsection{Video Object Segmentation as Few-shot Learning}
\label{sec:vos}
In video object segmentation (VOS), the target object is only defined by a reference target mask given in the first frame. No other prior information about the test object is assumed. The VOS method therefore needs to exploit the given first-frame annotation in order to segment the target in each subsequent frame of the video. To address this core problem in VOS, we first consider a general class of VOS architectures formulated as $\segnet(\image, \targetlayer(\image))$, where $\netparam$ denotes the learnable parameters. The network $\segnet$ takes the current image $\image$ along with the output of a target module $\targetlayer$. While $\segnet$ itself is target-agnostic, it is conditioned on $\targetlayer$, which exploits information about the target object, encoded in its parameters $\targetparam$. It generates a target-aware output that is used by $\segnet$ to predict the final segmentation. The target model parameters $\targetparam$ needs to be obtained from the initial image $\image_0$ and its given mask $\maskgt_0$, which defines the target object itself. We denote this as a function $\targetparam = \modelpredictor(\image_0, \maskgt_0)$. The key challenge in VOS, to which most of the research effort has been directed, is in the design of $\targetlayer$ and $\modelpredictor$.

It is important to note that the pair $(\image_0, \maskgt_0)$ constitutes a training sample for learning to segment the desired target. However, this training sample is only given during inference. Hence, a few-shot learning problem naturally arises within VOS. In this work, we adopt this view to develop our approach. In relation to few-shot learning, $\modelpredictor$ constitutes the internal learning method, which generates the parameters $\targetparam$ of the predictor $\targetlayer$ from a single example pair $(\image_0, \maskgt_0)$. While there exist a diverse set of few-shot learning methodologies, we aim to find the target model parameters $\targetparam$ that minimizes a supervised learning objective $\internalloss$,
\begin{equation}
\label{eq:online_loss}
\targetparam = \modelpredictor(x_0, \maskgt_0) = \argmin_{\targetparam'} \, \internalloss (T_{\targetparam'}(x_0), \maskgt_0) \,.
\end{equation}
Here, the target module $\targetlayer$ is learned to output the segmentation of the target object in the initial frame. In general, we operate on a deep representation of the input image $x = \featextractor(\image)$, generated by e.g.\ a ResNet architecture. Given a new frame $\image$ during inference, the object is segmented as $\segnet(\image, \targetlayer(\featextractor(\image))$. In other words, the target module is applied to the new frame to generate a first segmentation. This output is further refined by $\segnet$, which can additionally integrate powerful pre-learned knowledge from large VOS datasets. 

The main advantage of the optimization-based formulation~\eqref{eq:online_loss} is that the target model is predicted by directly minimizing the segmentation error in the first frame. It is thus capable of learning a powerful model, which can generate a robust prediction of the target in the coming frames. However, for practical purposes, the target model prediction \eqref{eq:online_loss} needs to be efficient. Equally important, to enable end-to-end training of the entire VOS architecture, we wish the learner $\modelpredictor$ to be \emph{differentiable}. While this is challenging in general, different strategies have been proposed in the literature~\cite{lee2019meta,bertinetto2018metalearning,bhat2019learning}, mostly in the context of meta-learning. We detail the employed approach in Sec.~\ref{sec:meta_learning}. In the next section, we first address another fundamental limitation of the formulation in Eq.~\eqref{eq:online_loss}.

\subsection{Learning What to Learn}
\label{sec:learn_wtl}
In our first approach, discussed in the previous section, the target module $\targetlayer$ learns to predict a first segmentation mask of the target object from the initial frame. This mask is then refined by the segmentation network $\segnet$, which possess learned strong segmentation priors. However, the segmentation network $\segnet$ is not limited to input an approximate target mask in order to perform target-conditional segmentation. In contrast, any information that alleviates the task of the network $\segnet$ to identify and accurately segment the target object is beneficial.  
Predicting only a single-channel mask thus severely limits the amount of target-specific information that can be passed to the segmentation network $\segnet$. Moreover, it is difficult for the internal few-shot learner $\modelpredictor$ to generate a target model $\targetlayer$ capable of performing a full segmentation of the object. Ideally, the target model should predict a rich representation of the target in the current frame, which provides strong target-aware cues that alleviate the task of the segmentation network $\segnet$. However, this is not possible in the standard few-shot learning setting \eqref{eq:online_loss}, since the output of the target module $\targetlayer$ is directly defined by the available ground-truth mask $\maskgt_0$. In this work, we address this issue by learning what our internal few-shot learner should learn.

Instead of directly employing the first-frame mask $\maskgt_0$ in our few-shot learner \eqref{eq:online_loss}, we propose to \emph{learn} the ground-truth, i.e.\ the labels of the few-shot learner. To this end, we introduce a trainable convolutional neural network $\labelenc(\maskgt)$ that takes a ground-truth mask $\maskgt$ as input and predicts ground-truth for the few-shot learner. The target model is thus predicted as,
\begin{equation}
    \label{eq:online_loss_enc}
    \targetparam = \modelpredictor(x_0, \maskgt_0) = \argmin_{\targetparam'} \, \internalloss \big(T_{\targetparam'}(x_0), \labelenc(\maskgt_0)\big) \,.
\end{equation}
Unlike the formulation in \eqref{eq:online_loss}, the encoded ground-truth mask $\labelenc(\maskgt_0)$ can be multi-dimensional. This allows the target module $\targetlayer$ to predict a richer representation of the object, providing powerful cues to the segmentation network.

While the label generator $\labelenc(\maskgt)$ predicts what the few-shot learner should learn, it does not handle the issue of data imbalance in our training set. For instance, a channel in the few-shot learner label $\labelenc(\maskgt)$ might encode object boundaries. However, as only a few pixels in the image belong to object boundary, it can be difficult to learn a target model which can perform such a task. We address this issue by  proposing a network module $\weightpredictor(\maskgt)$, called the weight predictor. Similar to $\labelenc$, it consists of a convolutional neural network taking a ground-truth mask $\maskgt$ as input. This module predicts the importance weight for each element in the loss $\internalloss \big(T_{\targetparam}(x_0), \labelenc(\maskgt_0)\big)$. It therefore has the same output dimensions as $\targetlayer$ and $\labelenc$. Importantly, our weight predictor can guide the few-shot learner to focus on the most crucial aspects of the ground truth label $\labelenc(\maskgt)$.

The formulation~\eqref{eq:online_loss_enc} has the potential of generating a more powerful target module $\targetlayer$. However, we have not yet fully addressed the question of \emph{how} to learn the ground-truth generating network $\labelenc$, and the weight predictor $\weightpredictor$. As previously discussed, we desire a rich encoding of the ground-truth that is also easy for the few-shot learner to learn. Ideally, we wish to train all parameters $\netparam$ in our segmentation architecture in an end-to-end manner on annotated VOS datasets. Indeed, we may back-propagate the error measured between the final segmentation output $\maskpred_t = \segnet(\image_t, \targetlayer(\featextractor(\image_t))$ and the ground truth $\maskgt_t$ on a test frame $\image_t$. However, this requires the internal learner \eqref{eq:online_loss_enc} to be efficient and differentiable w.r.t.\ both the underlying features $x$ \emph{and} the parameters of the label generator $\labelenc$ and weight predictor $\weightpredictor$. We address these open questions in the next section, to achieve an efficient and end-to-end trainable VOS architecture.

\subsection{Internal Learner}
\label{sec:meta_learning}

In this section, we detail our target module $\targetlayer$ and internal few-shot learner $\modelpredictor$. The target module $\targetlayer : \reals^{H \times W \times C} \rightarrow \reals^{H \times W \times D}$ maps a $C$-dimensional deep feature representation to a $D$-dimensional target-aware encoding of the same spatial dimensions $H \times W$. We require $\targetlayer$ to be efficient and differentiable. To ensure this, we employ a linear target module $\targetlayer(x) = x \conv \targetparam$, where $\targetparam \in \reals^{K \times K \times C \times D}$ constitutes the weights of a convolutional layer with kernel size $K$. Note that, the target module is linear and operates directly on a high-dimensional deep representation. As our results in Section~\ref{sec:experiments} clearly demonstrate, it learns to output powerful activations that encodes the target mask, leading to improved segmentation performance. Moreover, while a more complex target module has larger capacity, it is also prone to overfitting and is computationally more costly to learn. 

We design our internal learner to minimize the squared error between the output of the target module $\targetlayer(x)$ and the generated ground-truth labels $\labelenc(\maskgt)$, weighted by the element-wise importance weights $\weightpredictor(\maskgt_t)$,
\begin{equation}
    \label{eq:internal-loss}
    L(\targetparam) = \frac{1}{2}\sum_{(x_t, \maskgt_t) \in \basedset} \big\| \weightpredictor(\maskgt_t) \cdot \big(\targetlayer(x_t) - \labelenc(\maskgt_t)\big) \big\|^2 + \frac{\lambda}{2}\|\targetparam\|^2 \,.
\end{equation}
Here, $\basedset$ is the few-shot training set of the internal learner (i.e.\ support set). While it usually only contains one ground-truth annotated frame, it is often useful to include additional frames by, for instance, self-annotating new images in the video. The scalar $\lambda$ is a learned regularization parameter.

As a next step, we need to design a differentiable and efficient few-shot learner module that minimizes \eqref{eq:internal-loss} as $\targetparam = \modelpredictor(\basedset) = \argmin_{\targetparam'} L(\targetparam')$. The properties of the linear least squares loss \eqref{eq:internal-loss} aid us in achieving this goal. Note that \eqref{eq:internal-loss} is a convex quadratic objective in $\targetparam$. 
It hence has a well-known closed-form solution, which can be expressed in either primal or dual form. However, both options lead to computations that are intractable when aiming for acceptable frame-rates, requiring extensive matrix multiplications and solutions to linear systems. Moreover, these methods cannot directly utilize the convolutional structure of the problem. In this work, we therefore find an approximate solution of \eqref{eq:internal-loss} by applying steepest descent iterations, previously also used in \cite{bhat2019learning}. Given a current estimate $\targetparam^i$, it finds the step-length $\alpha^i$ that minimizes the loss in the gradient direction $\alpha^i = \argmin_\alpha L(\targetparam^i - \alpha g^i)$. Here, $g^i = \nabla L(\targetparam^i)$ is the gradient of \eqref{eq:internal-loss} at $\targetparam^i$. The optimization iteration can then be expressed as,
\begin{align}
    \label{eq:sd-step}
    \targetparam^{i+1} = \targetparam^i - \alpha^i g^i \,, \quad \alpha^i &= \frac{\|g^i\|^2}{\sum_t \| \weightpredictor(\maskgt_t) \cdot (x_t \conv g^i) \|^2 + \lambda \|g^i\|^2} \,, \nonumber \\
    g^i &= \sum_t x_t \conv\tp \Big(\weightpredictor^2(\maskgt_t) \cdot \big(x_t \conv \targetparam^i - \labelenc(\maskgt_t)\big)\Big) + \lambda \targetparam^i \,.
\end{align}
Here, $\conv\tp$ denotes the transposed convolution operation.
A detailed derivation is provided in Appendix \ref{sec:derivation}. 

Note that all computations in \eqref{eq:sd-step} are easily implemented using standard neural network operations. Moreover, all operations are differentiable, therefore the resulting target model parameters $\targetparam^i$ after $i$ iterations is differentiable w.r.t.\ all neural network parameters $\netparam$. Our internal few-shot learner is implemented as a neural network module $\modelpredictor(\basedset, \targetparam^0) = \targetparam^N$, by performing $N$ iterations of steepest descent \eqref{eq:internal-loss}, starting from a given initialization $\targetparam^0$. Thanks to the rapid convergence of steepest descent, we only need to perform a handful of iterations $N$ during training and inference. Moreover, our optimization based formulation allows the target model parameters $\targetparam$ to be efficiently updated with new samples by simply adding them to $\basedset$ and applying a few iterations \eqref{eq:sd-step} starting from the current parameters $\targetparam^0 = \targetparam$. We integrate this efficient, flexible and differentiable few-shot learner module into our VOS architecture, providing a powerful integration of target information.

\subsection{Video Object Segmentation Architecture}
\label{sec:vos_arch}

\begin{figure}[t]
    \centering
    \includegraphics[width=\columnwidth]{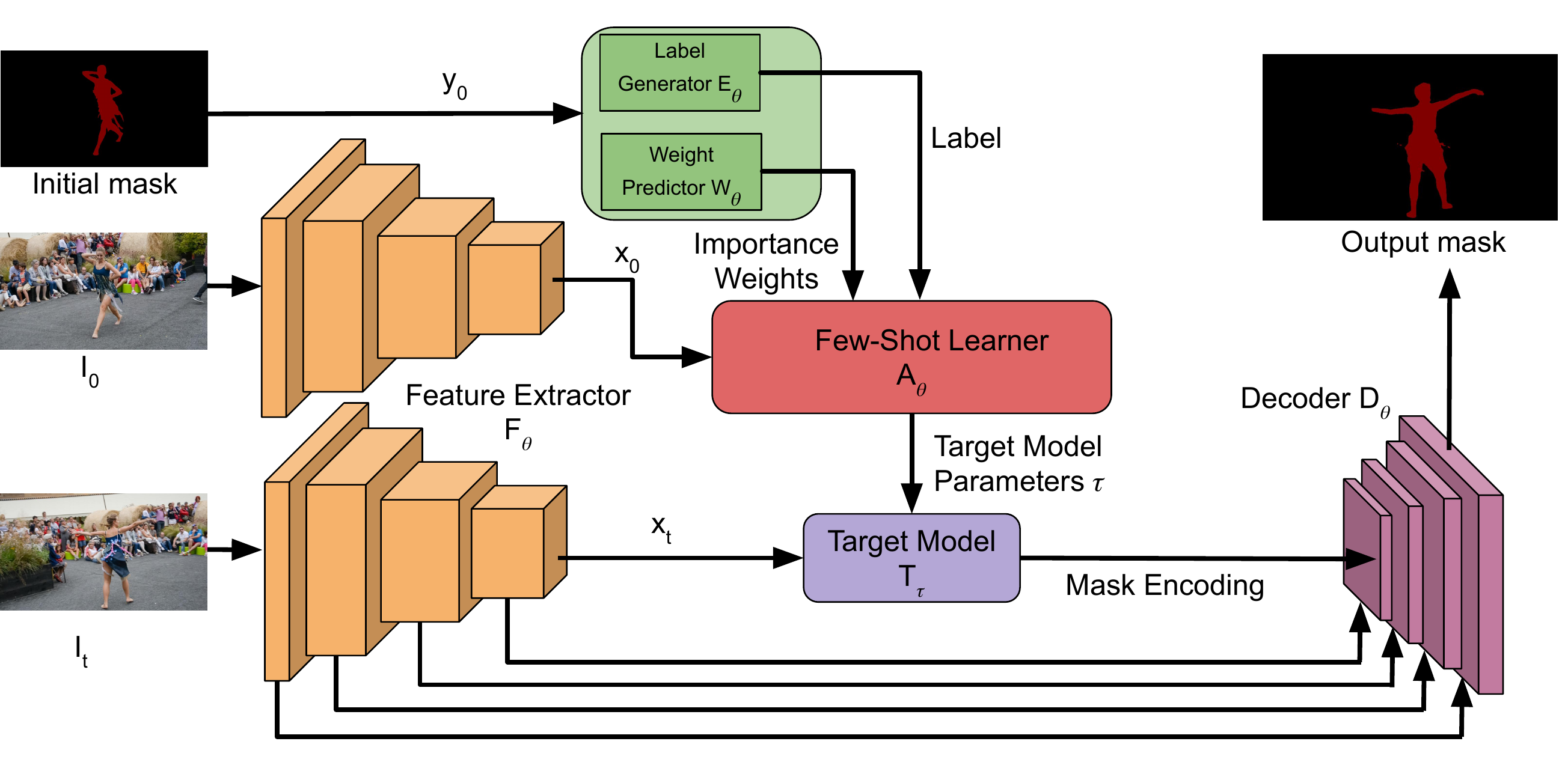}
    \vspace{-2mm}
    \caption{An overview of our segmentation architecture. It contains a few-shot learner, which generates a parametric target model $\targetlayer$ from the initial frame information. The parameters $\targetparam$ are computed by minimizing the loss in \eqref{eq:internal-loss}, with labels predicted by the label generator $\labelenc$. The elements of the loss are assigned importance weights predicted by the weight predictor $\weightpredictor$. In the incoming frames, the target model predicts the current mask encoding, which is processed along with image features by our decoder module $\decoder$ to produce the final segmentation mask.} 
    \vspace{-2mm}
    \label{fig:architecture}
\end{figure}

Our VOS method is implemented as a single end-to-end network architecture, illustrated in Fig.~\ref{fig:architecture}. The proposed architecture is composed of the following network modules: a deep feature extractor $\featextractor$, label generator $\labelenc$, loss weight predictor $\weightpredictor$, target module $\targetlayer$, few-shot learner $\modelpredictor$ and the segmentation decoder $\decoder$. As previously mentioned, $\netparam$ denotes the network parameters learned during the offline training, while $\targetparam$ are the target module parameters that are \emph{predicted} by the few-shot learner module.
The following sections detail the individual modules in our architecture.

\parsection{Feature extractor $\featextractor$} 
We employ a ResNet-50 network as backbone feature extractor $\featextractor$. Features from $\featextractor$ are input to both the decoder module $\decoder$ and the target module $\targetlayer$. For the latter, we employ the third residual block, which has a spatial stride of $s=16$. These features are first fed through an additional convolutional layer that reduces the dimension to $C=512$, before given to $\targetlayer$. 

\parsection{Few-shot label generator $\labelenc$}
Our ground-truth label generator $\labelenc$ predicts a rich representation by extracting useful visual cues from the target mask.
 The latter is mapped to the resolution of the deep features as $\labelenc: \reals^{sH\times sW \times 1} \rightarrow \reals^{H\times W \times D}$, where $H$, $W$ and $D$ are the height, width and dimensionality of the target model features and $s$ is the feature stride. We implement the proposed mask encoder $\labelenc$ as a convolutional network,
decomposed into a generic mask feature extractor for processing the raw mask $y$ and a prediction layer for generating the final label encoding. More details are provided in Appendix \ref{sec:supp_arch_label}.

\parsection{Weight predictor $\weightpredictor$} 
The weight predictor $\weightpredictor: \reals^{sH\times sW \times 1} \rightarrow \reals^{H\times W \times D}$ generates weights for the internal loss \eqref{eq:internal-loss}. It is implemented as a convolutional network that takes the target mask $y$ as input. In our implementation, $\weightpredictor$ shares the mask feature extractor with $\labelenc$. The weights are then predicted from the extracted mask features with a separate conv-layer.  

\parsection{Target module $\targetlayer$ and few-shot learner $\modelpredictor$}
We implement our target module $\targetlayer$ as convolutional filter with a kernel size of $K=3$. The number of output channels $D$ is set to $16$.
Our few-shot learner $\modelpredictor$ (see Sec.~\ref{sec:meta_learning}) predicts the target module parameters $\targetparam$ by recursively applying steepest descent iterations \eqref{eq:sd-step}. On the first frame in the sequence, we start from a zero initialization $\targetparam^0 = 0$. On the test frames, we apply the predicted target model parameters $\targetlayer(x)$ to predict the target activations, which are provided to the segmentation decoder.

\parsection{Segmentation decoder $\decoder$}
This module takes the output of the target module $\targetlayer$ along with backbone features in order to predict the final accurate segmentation mask.
Our approach can in principle be combined with any decoder architecture. For simplicity, we employ a decoder network similar to the one used in~\cite{robinson2020learning}. We adapt this network to process a multi-channel target mask encoding as input. More details are provided in Appendix \ref{sec:supp_arch_decoder}. 

\subsection{Inference}
\label{sec:inference}
In this section, we describe our inference procedure.
Given a test sequence $\video = \{\image_t\}_{t=0}^Q$, along with the first frame annotation $\maskgt_0$, we first create an initial training set $\basedset_0 = \{(x_0, \maskgt_0)\}$ for the few-shot learner, consisting of the single sample pair. Here, $x_0 = \featextractor(\image_0)$ is the feature map extracted from the first frame. The few-shot learner then predicts the parameters $\targetparam_0 = \modelpredictor(\basedset_0, \targetparam^0)$ of the target module by minimizing the internal loss \eqref{eq:internal-loss}. We set the initial estimate of the target model $\targetparam^0 = 0$ to all zeros. Note that the ground-truth $\labelenc(\maskgt_0)$ and importance weights $\weightpredictor(\maskgt_0)$ for the minimization problem \eqref{eq:internal-loss} are predicted by our network. 

The learned model $\targetparam_0$ is then applied on the subsequent test frame $\image_1$ to obtain an encoding $T_{\tau_0}(x_1)$ of the segmentation mask. This mask encoding is then processed by the decoder module, along with the image features, to generate the mask prediction $\maskpred_1 = \decoder(x_1, T_{\tau_0}(x_1))$. In order to adapt to the changes in the scene, we further update our target model using the information from the processed frame. This is achieved by extending the few-shot training set $\basedset_0$ with the new training sample $(x_1, \maskpred_1)$, where the predicted mask $\maskpred_1$ serves as the pseudo-label for the frame $\image_1$. The extended training set $\basedset_1$ is then used to obtain a new target module parameters $\targetparam_1 = \modelpredictor(\basedset_1, \targetparam_0)$. Note that instead of predicting the parameters $\targetparam_1$ from scratch, our optimization based learner allows us to update the previous target model $\targetparam_0$, which increases the efficiency of our approach. Specifically, we apply additional $N_\text{update}^\text{inf}$ steepest-descent iterations \eqref{eq:sd-step} with the new training set $\basedset_1$. The updated model $\targetparam_1$ is then applied on the next frame $\image_2$. This process is repeated till the end of the sequence.

\parsection{Details} 
Our internal few-shot learner $\modelpredictor$ employs  $N_\text{init}^\text{inf} = 20$ iterations in the first frame and $N_\text{update}^\text{inf} = 3$ iterations in each subsequent frame. 
Our few-shot learner formulation \eqref{eq:internal-loss} allows an easy integration of a global importance weight for each frame in the training set $\basedset$. We exploit this flexibility to integrate an exponentially decaying weight $\eta^{-t}$ to reduce the impact of older frames. We set $\eta = 0.9$ and ensure the weights sum to one. 
We ensure a maximum $K_{\mathrm{max}}=32$ samples in the few-shot training dataset $\basedset$, by removing the oldest. We always keep the first frame since it has the reference target mask $\maskgt_0$.
Each frame in the sequence is processed by first cropping a patch that is $5$ times larger than the previous estimate of target, while ensuring the maximal size to be equal to the image itself. The cropped region is resized to $832 \times 480$ with preserved aspect ratio. If a sequence contains multiple targets, we independently process each in parallel, and merge the predicted masks using the soft-aggregation operation~\cite{oh2018rgmp}.

\subsection{Training}
\label{sec:training_details}
To train our end-to-end network architecture, we aim to simulate the inference procedure employed by our approach, described in Section \ref{sec:inference}. This is achieved by training the network on mini-sequences $\video = \{(\image_t, \maskgt_t)\}_{t=0}^{Q-1}$ of length $Q$. These are constructed by sampling frames from annotated VOS sequences. In order to induce robustness to fast appearance changes, we randomly sample frames in temporal order from a larger window consisting of $Q'$ frames. As during inference, we create the initial few-shot training set from the first frame $\basedset_0 = \{(x_0, \maskgt_0)\}$. This is used to learn the initial target module parameters $\targetparam_0 = \modelpredictor(\basedset_0, 0)$ by performing $N_\text{init}^\text{train}$ steepest descent iterations. In subsequent frames, we use $N_\text{update}^\text{train}$ iterations to update the model as $\targetparam_t = \modelpredictor(\basedset_t, \targetparam_{t-1})$. The final prediction $\maskpred_t = \decoder(x_t, T_{\targetparam_{t-1}}(x_t))$ in each frame is added to the few-shot train set $\basedset_t = \basedset_{t-1} \cup \{(x_t,\maskpred_t)\}$. All network parameters $\netparam$ are trained by minimizing the per-sequence loss,
\begin{equation}
    \label{eq:train-loss}
    \segloss_\text{seq}(\netparam; \video) = \frac{1}{Q-1}\sum_{t=1}^{Q-1} \segloss\Big(\decoder\big(\featextractor(\image_t), T_{\targetparam_{t-1}}(\featextractor(\image_t))\big), \, \maskgt_t\Big) \,.
\end{equation}
Here, $\segloss(\maskpred, \maskgt)$ is the employed segmentation loss between the prediction $\maskpred$ and ground-truth $\maskgt$. 
We compute the gradient of the final loss \eqref{eq:train-loss} by averaging over multiple mini-sequences in each batch. Note that the target module parameters $\targetparam_{t-1}$ in \eqref{eq:train-loss} are predicted by our differentiable few-shot learner $\modelpredictor$, and therefore depend on the network parameters of the label generator $\labelenc$, weight predictor $\weightpredictor$, and feature extractor $\featextractor$. These modules can therefore be trained end-to-end thanks to the differentiability of our learner $\modelpredictor$.

\parsection{Details} Our network is trained using the YouTube-VOS \cite{xu2018youtube2} and DAVIS \cite{perazzi2016davis} datasets. We use mini-sequences of length $Q=4$ frames, generated using a video segment of length $Q'=100$. We employ random flipping, rotation, and scaling for data augmentation. We then sample a random $832 \times 480$ crop from each frame. The number of steepest-descent iterations in the few-shot learner $\modelpredictor$ is set to $N_\text{init}^{train}=5$ for the first frame and $N_\text{update}^\text{train}=2$ in subsequent frames. We use the Lovasz~\cite{berman2018lovasz} segmentation loss in \eqref{eq:train-loss}. 
To avoid using combinations of additional still-image segmentation datasets as performed in e.g.~\cite{oh2019video,oh2018rgmp}, we initialize our backbone ResNet-50 with the Mask R-CNN~\cite{He2017MaskR} weights from \cite{massa2018mrcnn} (see Appendix \ref{sec:sup_abl} for analysis). All other modules are initialized using~\cite{KaimingInit}. 
Our network is trained using ADAM~\cite{kingma2014adam} optimizer. We first train our network for $70$k iterations with the backbone weights fixed. The complete network, including the backbone feature extractor, is then trained for an additional $100$k iterations. The entire training takes $48$ hours on 4 Nvidia V100 GPUs. Further details about our training is provided in Appendix \ref{sec:sup_training}.

\subsection{Bounding Box Initialization}
\label{sec:bbinit}
In many practical applications, it is too costly to generate an accurate reference-frame annotation in order to perform video object segmentation. In this work, we therefore investigate using weaker supervision in the first frame. In particular, we follow the recent trend of only assuming the target bounding box in the first frame of the sequence~\cite{wang2019fast,voigtlaender2019siam}. 
By exploiting our learned ground-truth encoding, we show that our architecture can accommodate this important setting with only a minimal addition. Analogously to the label generator $\labelenc$, we introduce a bounding box encoder module $\boxenc(b_0, x_0)$. It takes a mask-representation $b_0$ of the initial box along with backbone features $x_0$. The bounding box encoder $\boxenc$ then predicts a target mask representation in the same $D$-dimensional output space of $\labelenc$ and $\targetlayer$. This allows us to exploit our existing decoder network in order to predict the target segmentation in the initial frame as $\maskpred_0 = \decoder(x_0, \boxenc(b_0, x_0))$. VOS is then performed using the same procedure as described in Sec.~\ref{sec:inference}, by simply replacing the ground-truth reference mask $\maskgt_0$ with the predicted mask $\maskpred_0$. Our box encoder $\boxenc$, consisting of only a linear layer followed by two residual blocks, is easily trained by freezing the other parameters in the network, which ensures preserved segmentation performance. Thus, we only need to sample single frames during training and minimize the segmentation loss $\segloss(\decoder(x_0, \boxenc(b_0, x_0)), \maskgt_0)$. As a result, we gain the ability to perform VOS with bounding box initialization by only adding the small box encoder $\boxenc$ to our overall architecture, without loosing any performance in the standard VOS setting.

\parsection{Details} We train the box encoder on images from MSCOCO~\cite{lin2014microsoft} and YouTube-VOS for $50,000$ iterations, while freezing the pre-trained components of the network. During inference we reduce the impact of the first frame annotation by setting $\eta=0.8$ and remove it from the memory after $K_{\mathrm{max}}$ frames. For best performance, we only update the target model every fifth frame with $N_\text{update}^\text{inf}=5$.

\section{Experiments}
\label{sec:experiments}
We evaluate our approach on the two standard VOS benchmarks: YouTube-VOS and DAVIS 2017. Detailed results are provided in the appendix. Our approach operates at  $6$ FPS on single object sequences. \textbf{Code:} Train and test code along with trained models will be released upon publication.

\subsection{Ablative Analysis}
\begin{figure}[t]
\tabcolsep=0.01cm
\renewcommand{\arraystretch}{0.06}
\renewcommand{\image}{\includegraphics[width=0.166\columnwidth]}
\centering
\begin{tabular}{cccccc}
\image{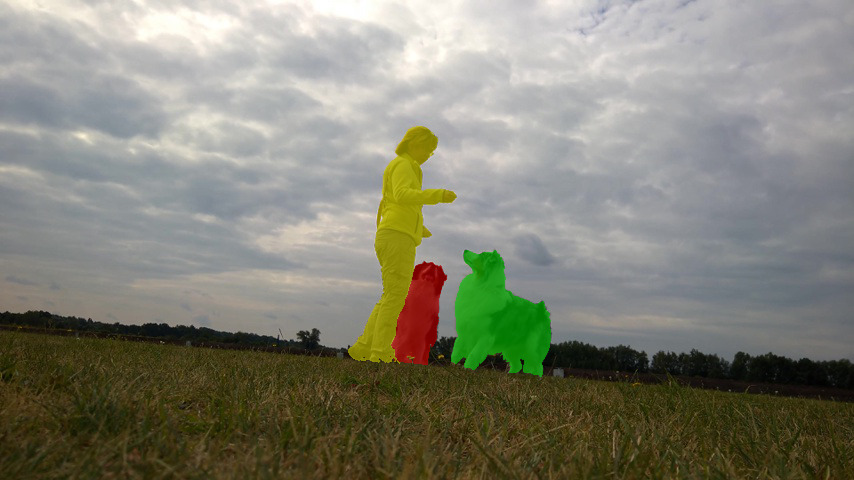} &
\image{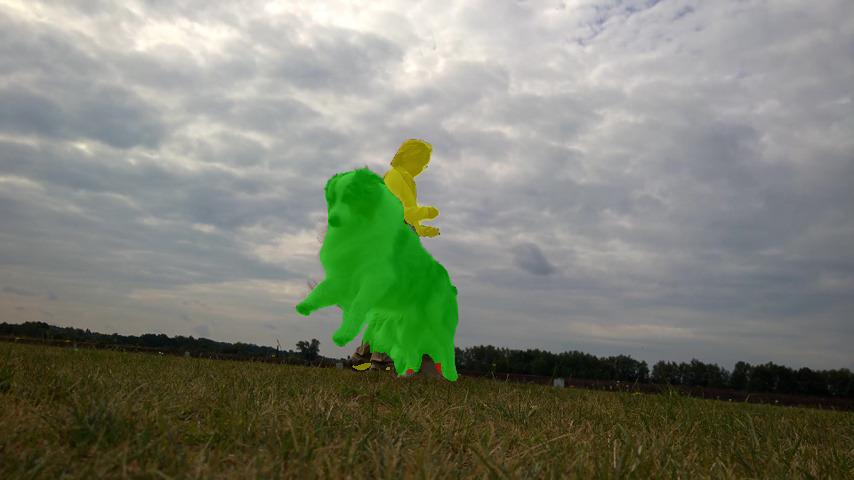} &
\image{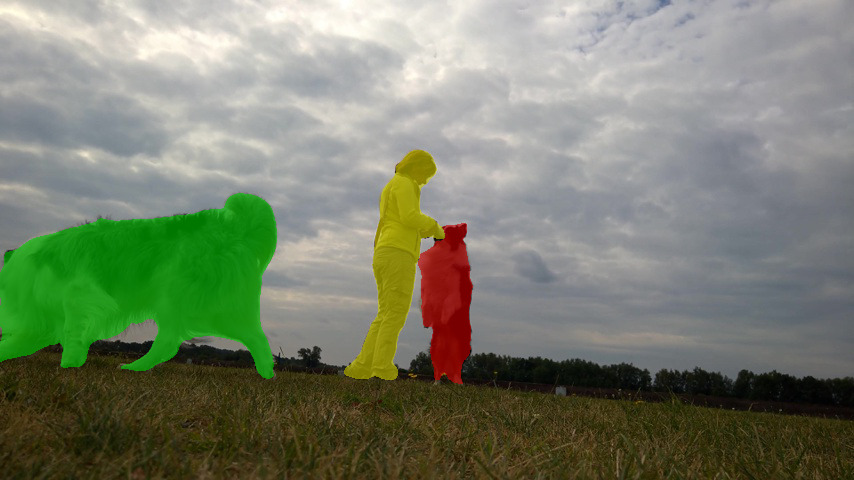} &
\image{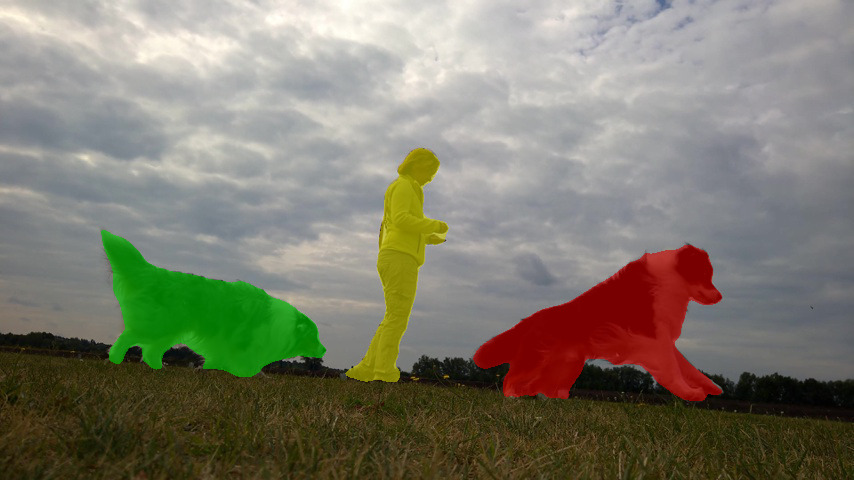} &
\image{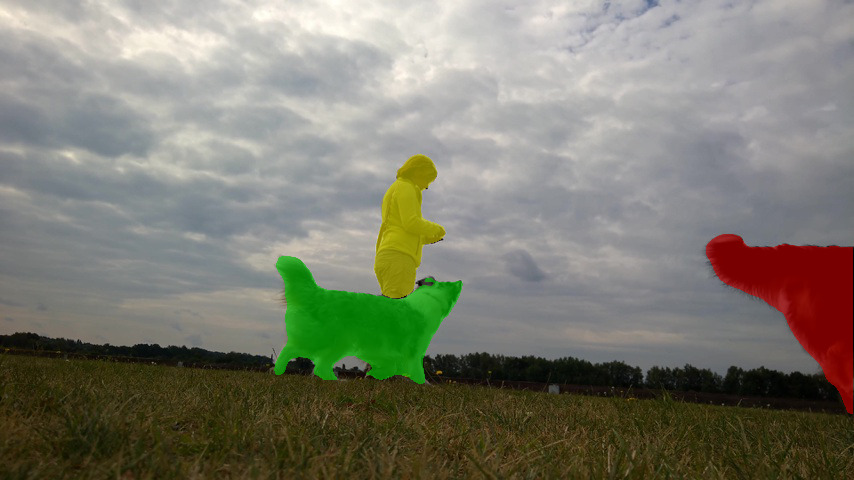} &
\image{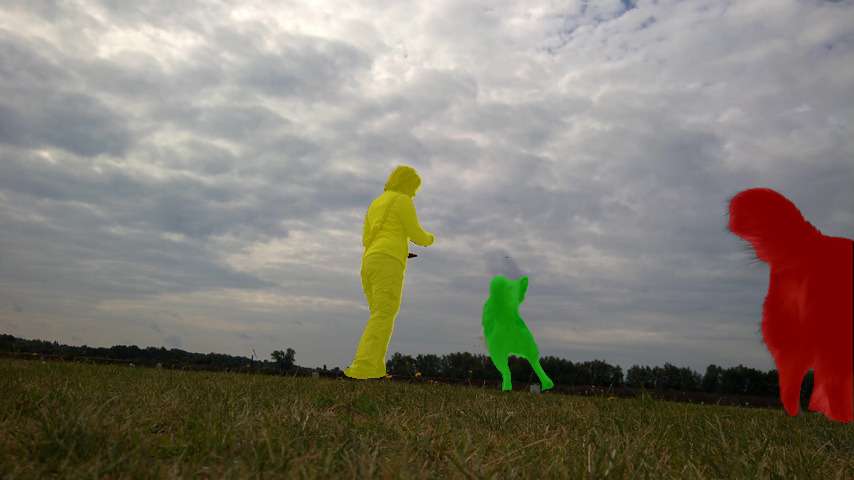} \\
\image{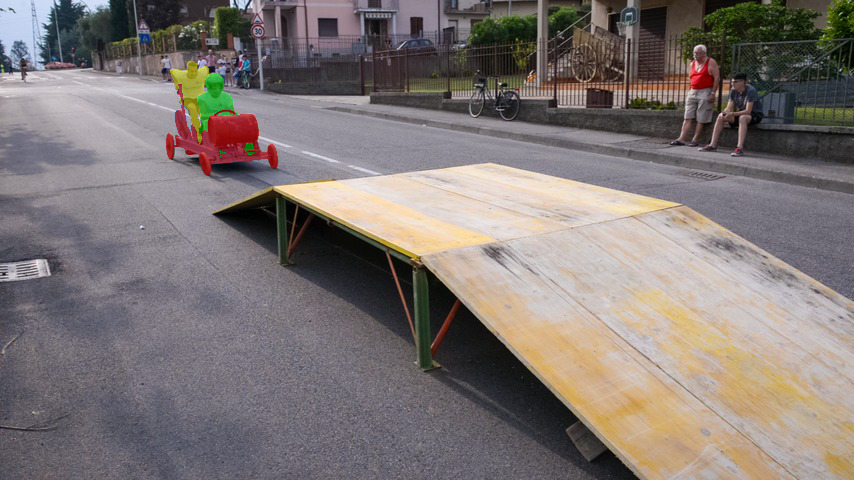} &
\image{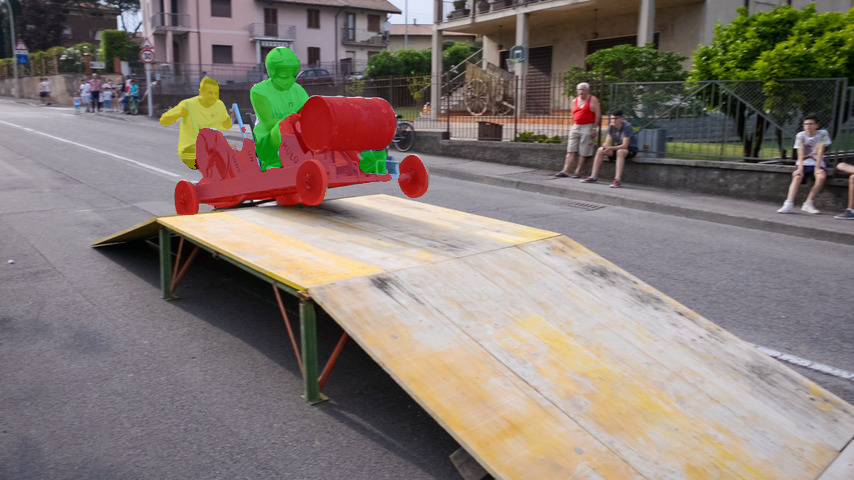} &
\image{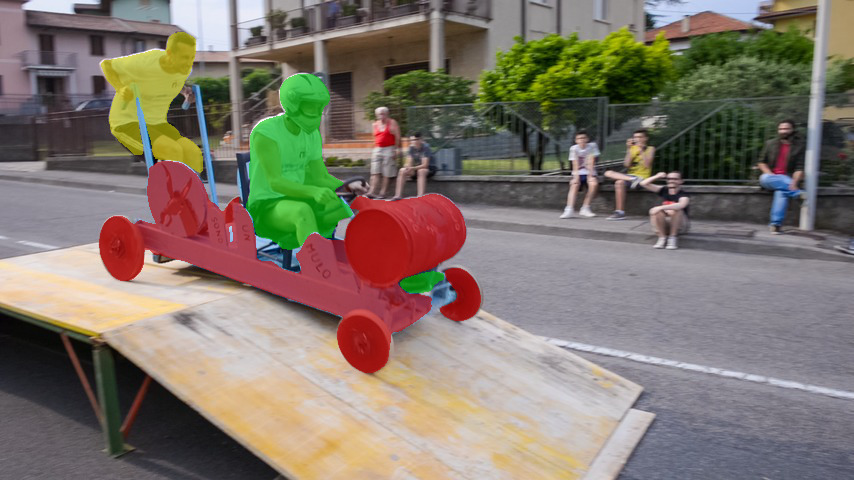} &
\image{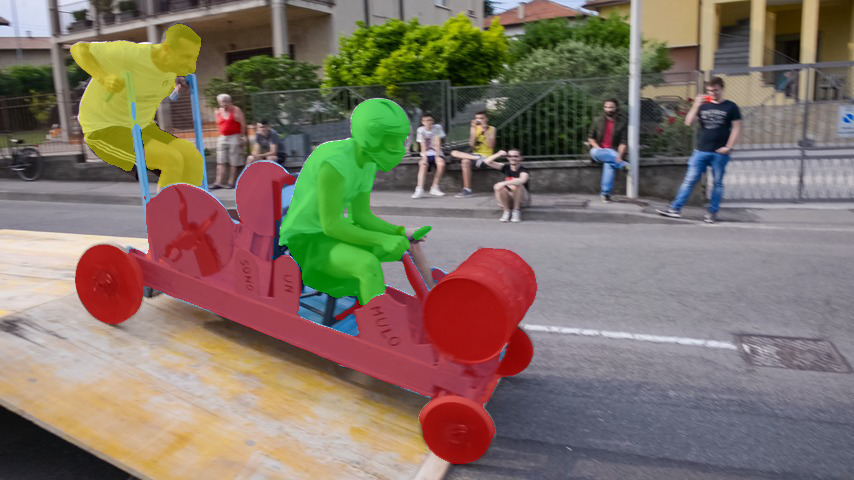} &
\image{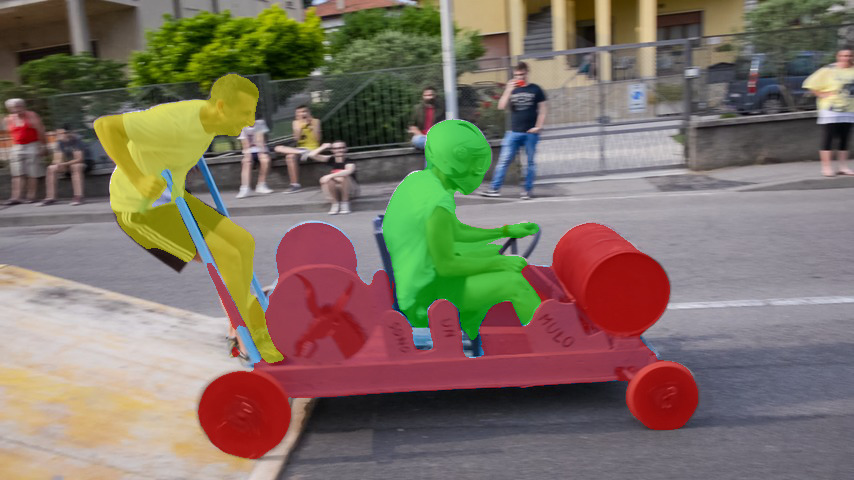} &
\image{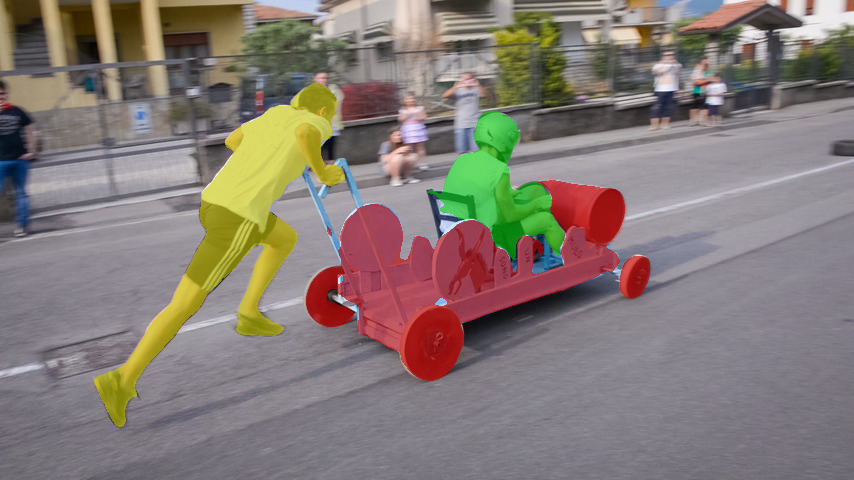}  \\
\image{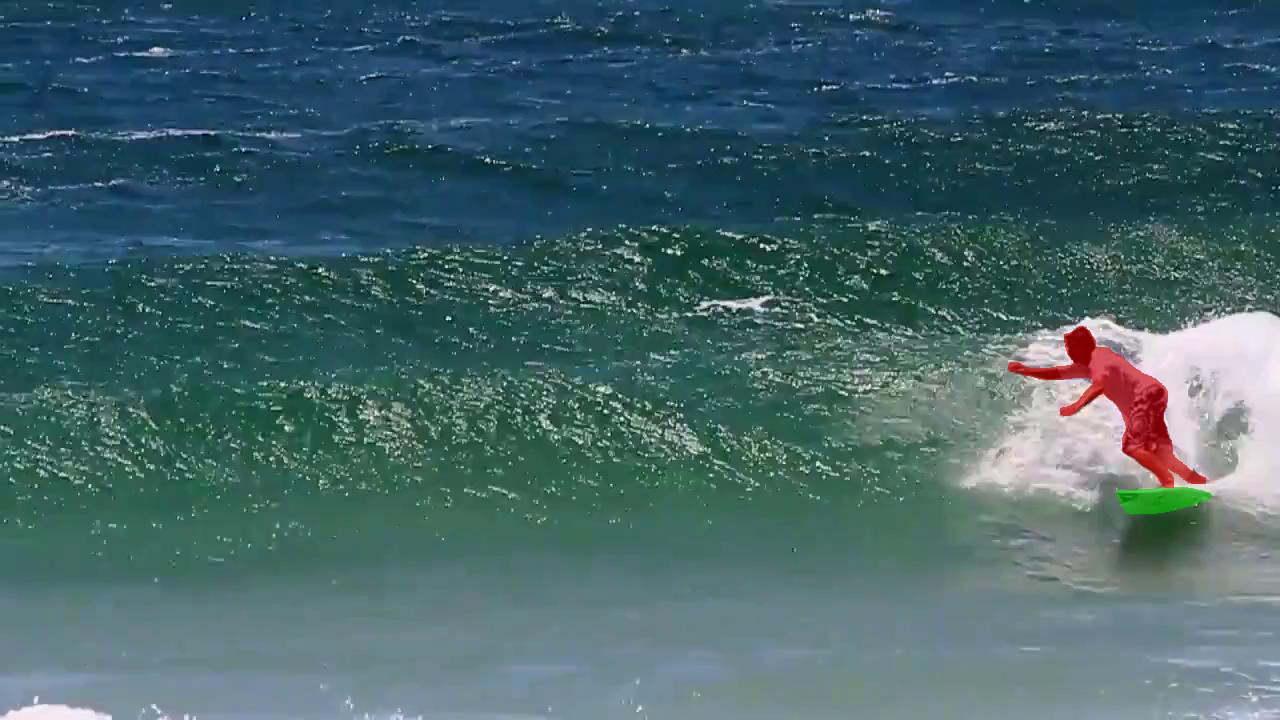} &
\image{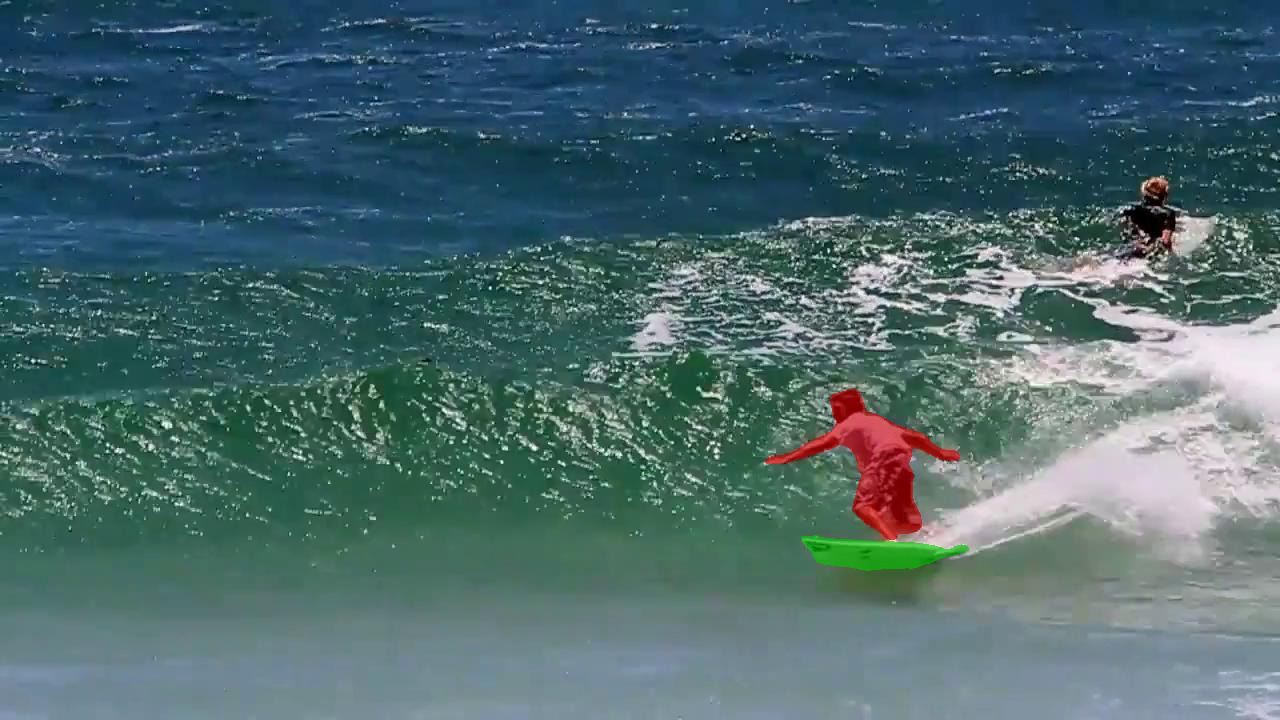} &
\image{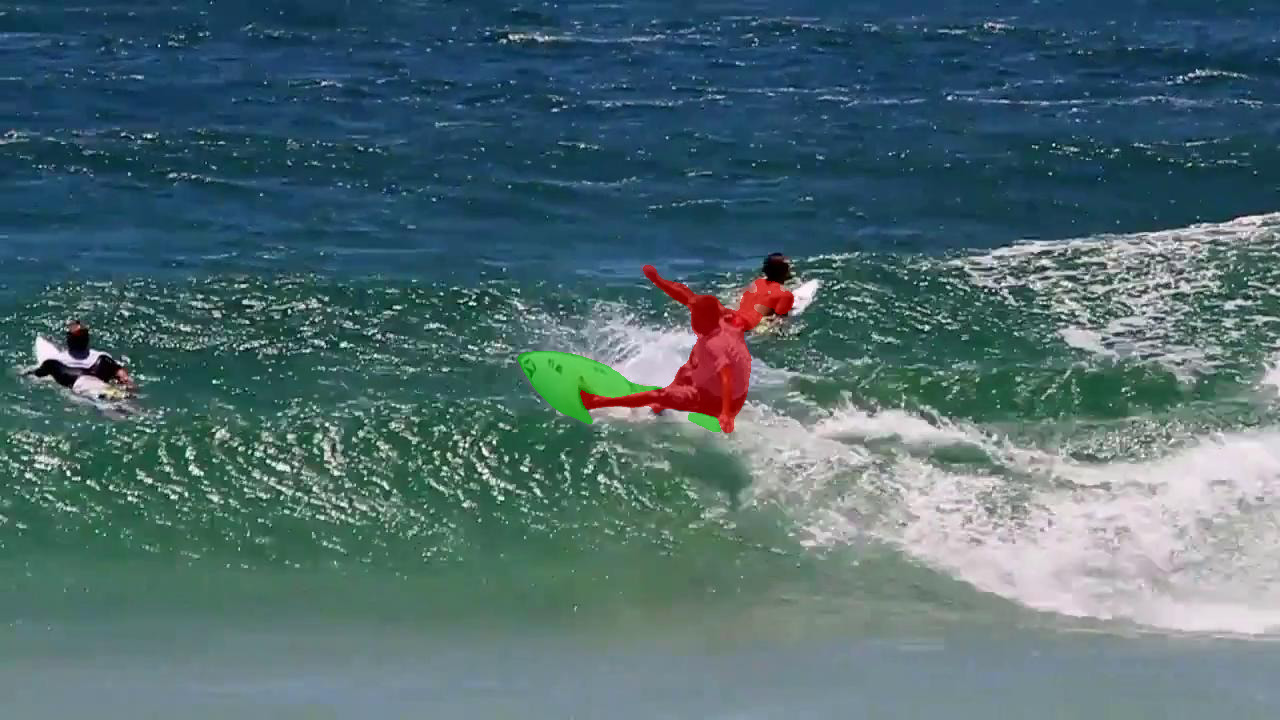} &
\image{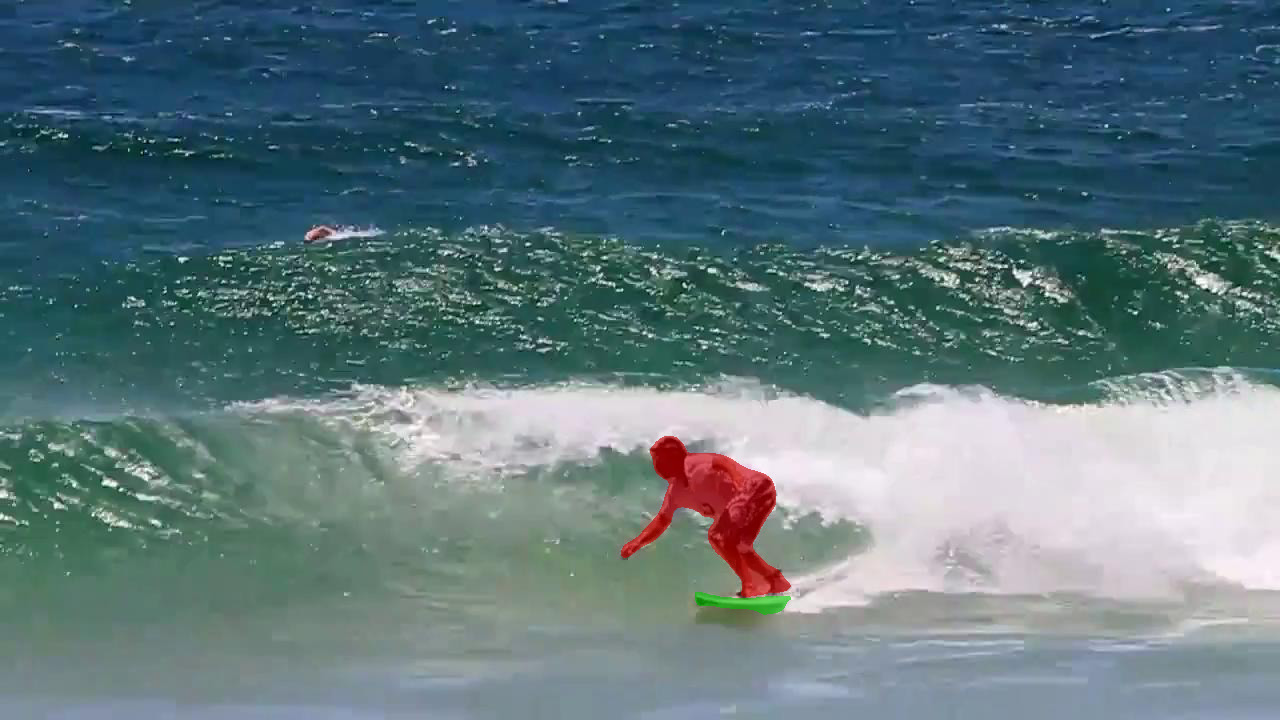} &
\image{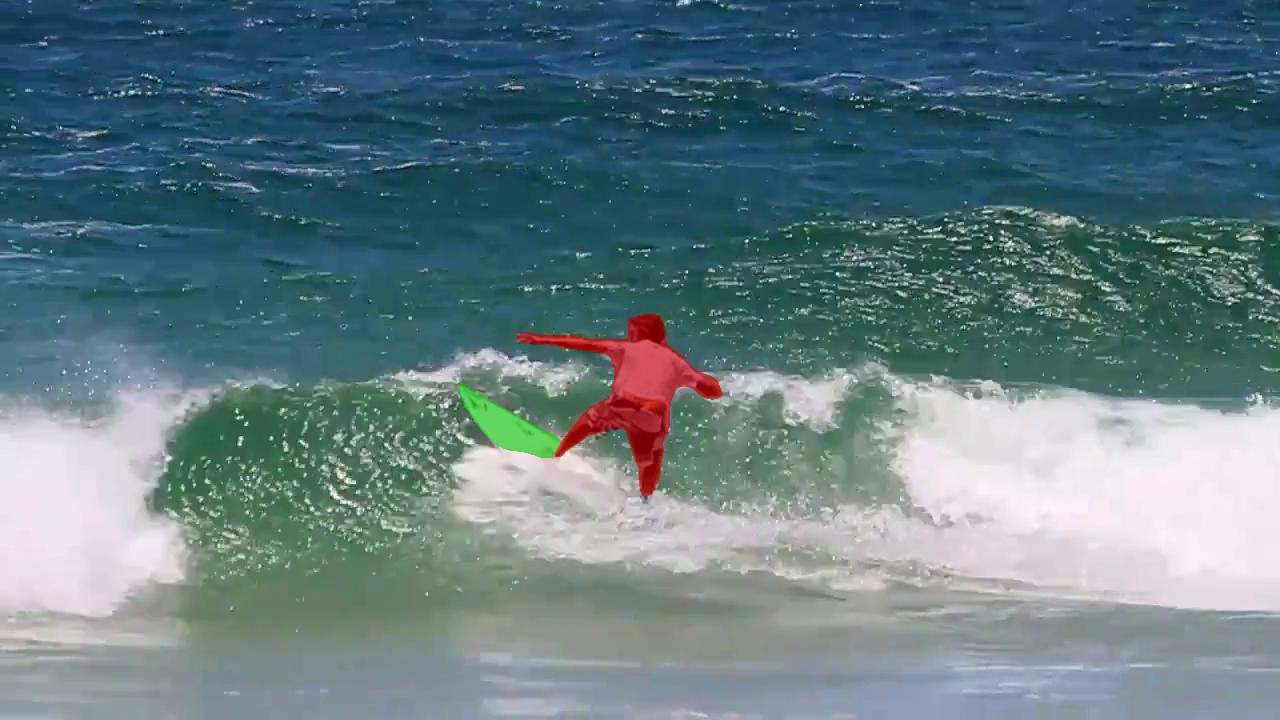} &
\image{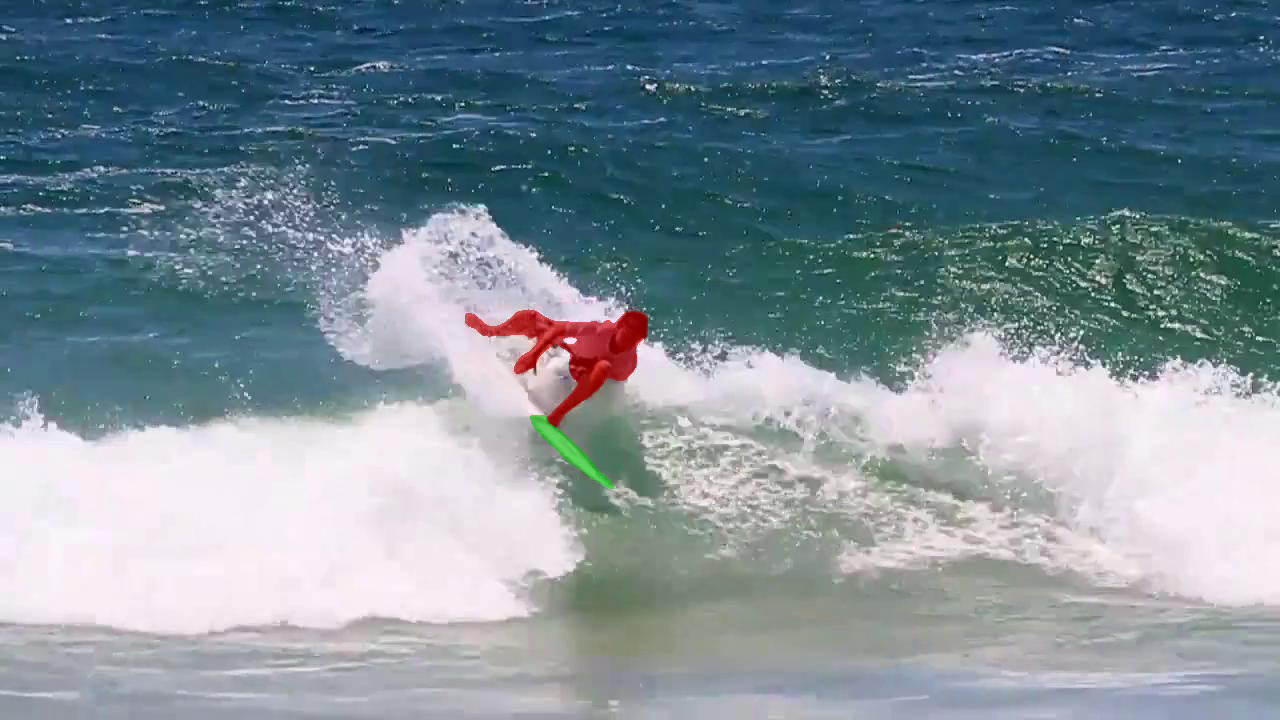} \\
\image{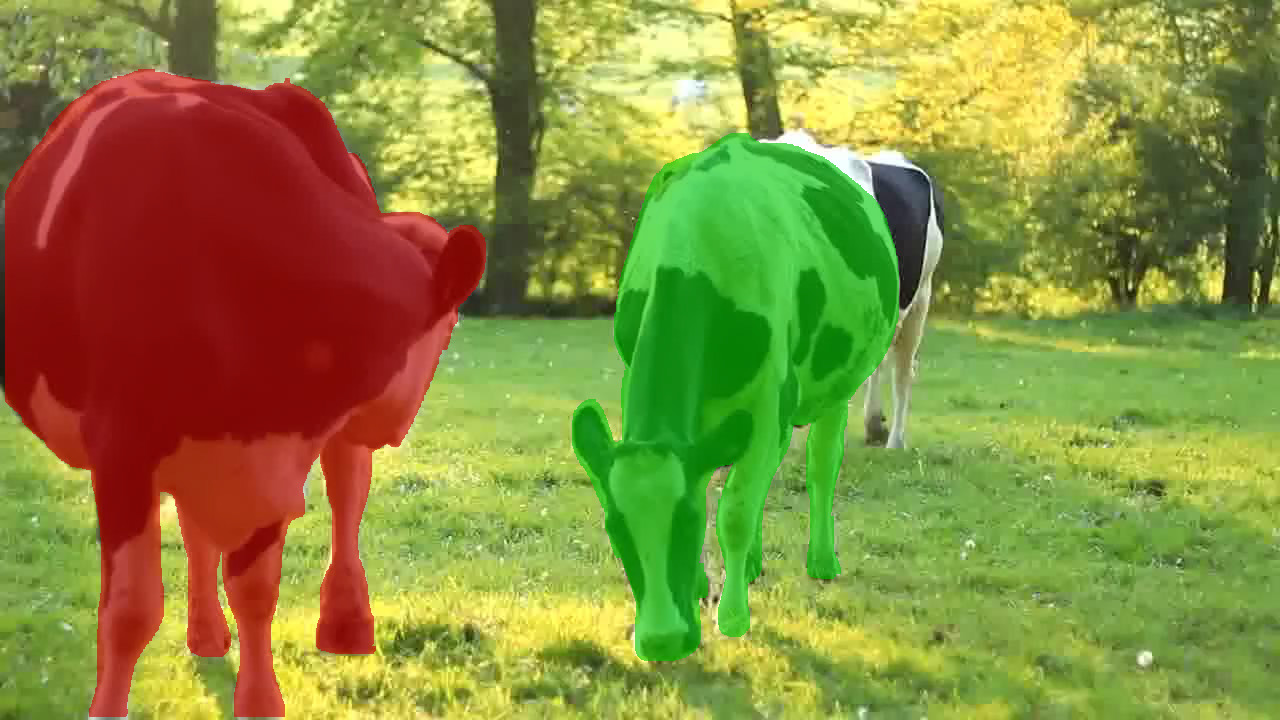} &
\image{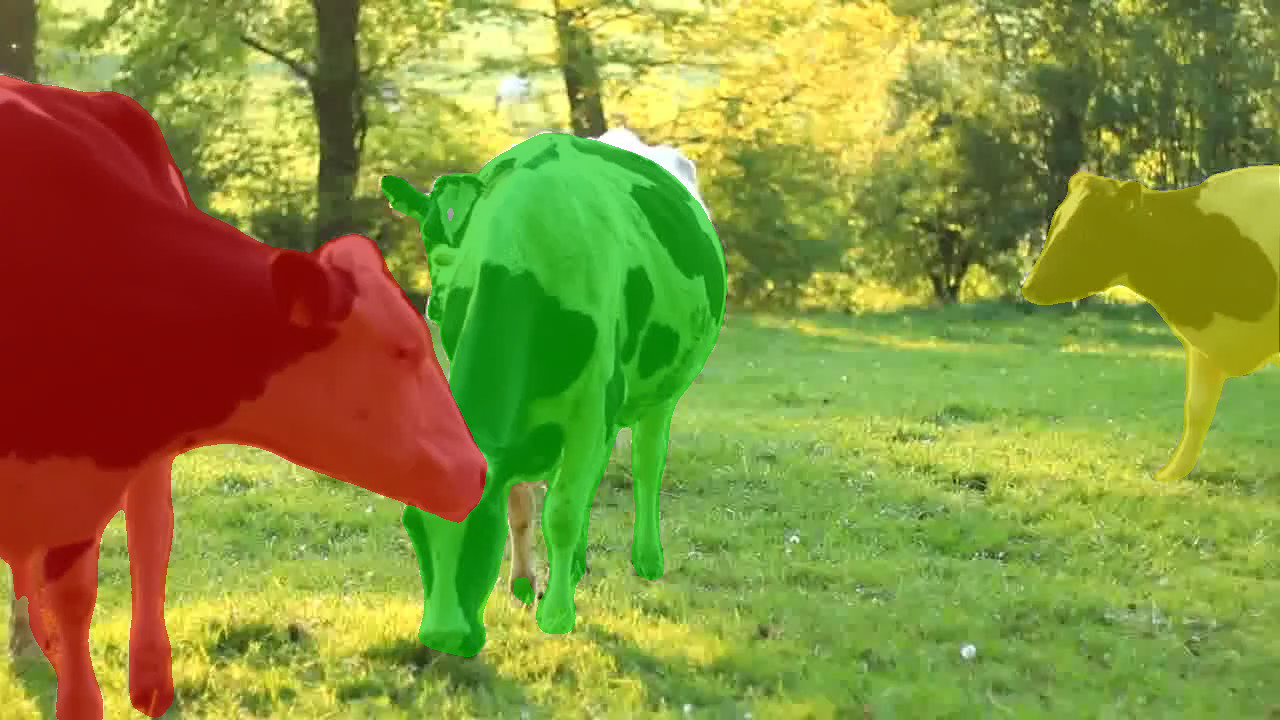} &
\image{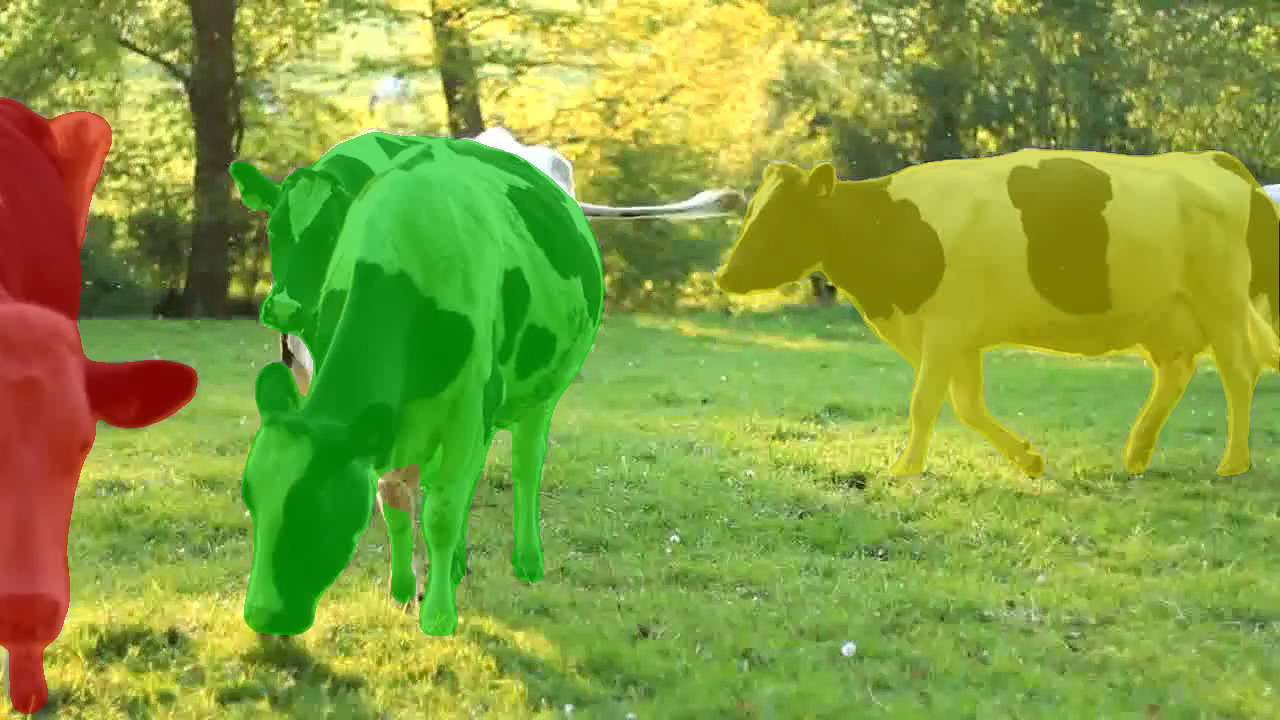} &
\image{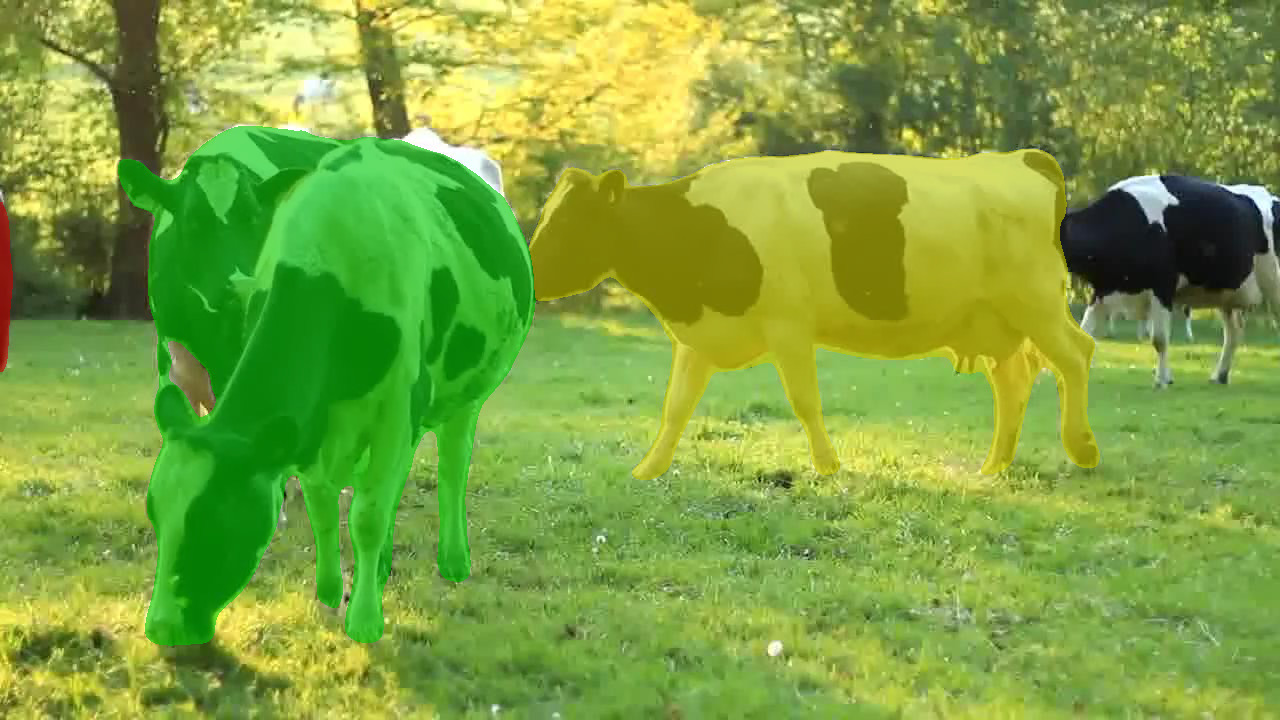} &
\image{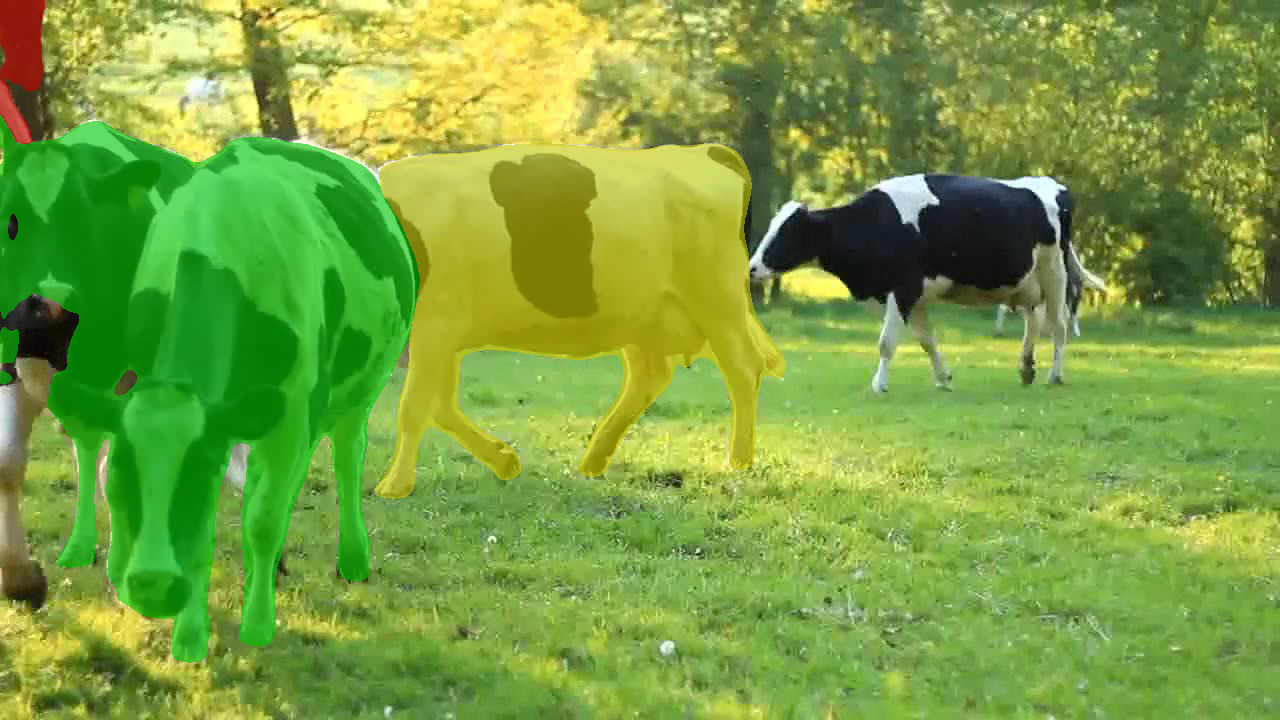} &
\image{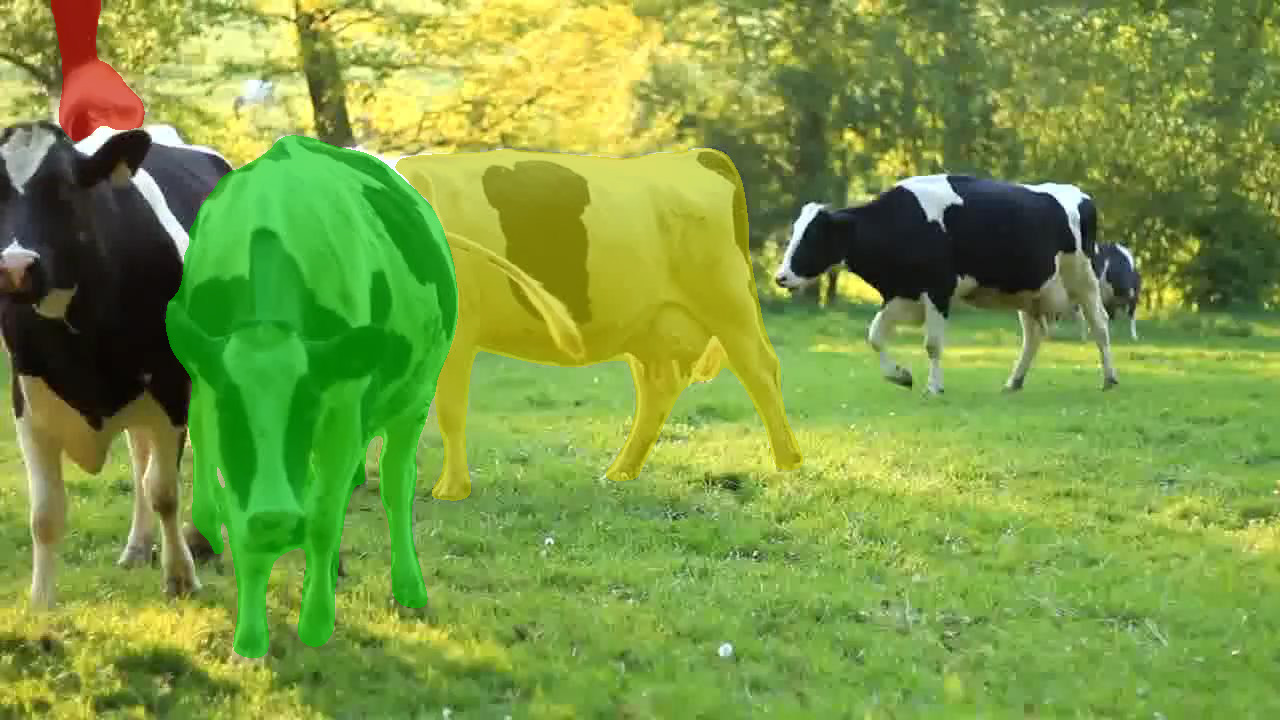}
\end{tabular}\vspace{0mm}
\caption{Qualitative results of our VOS method. Our approach provides accurate segmentations in very challenging scenarios, including occlusions (row 1 and 3), distractor objects (row 1, and 2), and appearance changes (row 1, 2 and 3). Row 4 shows an example failure case, due to severe occlusions and very similar objects.}\vspace{0mm}
\label{fig:qualitative}
\end{figure}

Here, we analyze the impact of the key components in the proposed VOS architecture. Our analysis is performed on a validation set consisting of 300 sequences randomly sampled from the YouTube-VOS 2019 training set. For simplicity, we do not train the backbone ResNet-50 weights in the networks of this comparison. The networks are evaluated using the mean Jaccard $\mcJ$ index (IoU). Results are shown in Tab.~\ref{tab:ablation}. Qualitative examples are visualized in Fig.~\ref{fig:qualitative}.

\parsection{Baseline} Our baseline constitutes a version where the target model is trained to directly predict an initial mask, which is subsequently refined by the decoder $\decoder$. That is, the ground-truth employed by the few-shot learner is set to the reference mask. Further, we do not back-propagate through the learning of the target model during offline training and instead only train the decoder module $\decoder$. Thus, it does not perform end-to-end training through the learner. 

\parsection{End-to-end Training} Here, we exploit the differentiablity of our few-shot learner to train the underlying features used by the target model in an end-to-end manner. 
Learning the specialized features for the target model provides a substantial improvement of $+3.0$ in $\mcJ$ score. This clearly demonstrates the importance of end-to-end learning capability provided by our few-shot learner.

\parsection{Label Generator $\labelenc$} Instead of training the target model to predict an initial segmentation mask, we here employ the proposed label generator module $\labelenc$ to learn what the target model should learn. 
This allows training the target model to output richer representation of the target, leading to an improvement of $+1.4$ in $\mcJ$ score over the version which does not employ the label generator.

\parsection{Weight Predictor $\weightpredictor$} In this version, we additionally include the proposed weight predictor $\weightpredictor$ to obtain the importance weights for the internal loss \eqref{eq:internal-loss}. Using the importance weights leads to an improvement of an additional $+0.9$ in $\mcJ$ score. This shows that our weight predictor learns to predict what the internal learner should focus on, in order to generate a robust target model.

\begin{table}[t!]
\caption{Ablative analysis of our approach on a validation set consisting of $300$ videos sampled from the YouTube-VOS 2019 training set. We analyze the impact of {\bf end-to-end training}, the {\bf label generator} module  and the  {\bf weight predictor}.}\vspace{-2mm}
\centering
\resizebox{0.80\columnwidth}{!}{%
\renewcommand{\yes}{\text{\ding{51}}}
\renewcommand{\no}{-}
\centering
\begin{tabular}{lc@{~~}c@{~~}c@{~~}c}
\toprule
 &End-to-end&Label Generator&Weight Predictor& $\mathcal{J}$ \\
\midrule
Baseline&\no&\no&\no&74.5 \\
+ End-to-end training&\yes&\no&\no&77.5\\
+ Label Encoder ($\labelenc$) &\yes&\yes&\no&78.9 \\
+ Weight Predictor ($\weightpredictor$)&\yes&\yes&\yes&79.8 \\
\bottomrule
\end{tabular}

}\vspace{0mm}
\label{tab:ablation}
\end{table}

\subsection{State-of-the-art Comparison}
Here, we compare our method with state-of-the-art. Since many approaches employ additional segmentation datasets during training, we always indicate whether additional data is used. 
We report results for two versions of our approach: the standard version which employs additional data (as described in Sec.~\ref{sec:training_details}), and one that is only trained on the train split of the specific dataset. For the latter version, we initialize the backbone ResNet-50 with ImageNet pre-training instead of the MaskRCNN backbone weights (see Sec.~\ref{sec:training_details}).

\begin{table}[t!]
\caption{State-of-the-art comparison on the large-scale YouTube-VOS 2018 validation dataset.
Our approach outperforms all previous methods, both when comparing with additional training data and when training only on YouTube-VOS 2018 train split.}\vspace{-2mm}
\centering
\resizebox{0.99\columnwidth}{!}{%
\newcommand{\first}[1]{\textbf{{#1}}}
\centering
\begin{tabular}{l@{~~}cccccc:cccccc}
\toprule
 & \multicolumn{6}{c}{\textbf{Additional Training Data}} & \multicolumn{6}{c}{\textbf{Only YouTube-VOS training}} \\
 &  {\bf Ours} & STM & SiamRCNN &  PreMVOS & OnAVOS &  OSVOS~ &  ~{\bf Ours} &  STM &  FRTM &  AGAME & AGSSVOS  & S2S  \\
 && \cite{oh2019video} & \cite{voigtlaender2019siam} & \cite{luiten2018premvos}& \cite{voigtlaender2017onavos} & \cite{caelles2017osvos} & & \cite{oh2019video} & \cite{robinson2020learning} & \cite{johnander2018generative} & \cite{lin2019agss} & \cite{xu2018youtube}  \\
\midrule
$\mathcal{G}$ (overall) & \first{81.5} & 79.4 & 73.2 & 66.9 & 55.2 & 58.8 & \first{80.2} & 68.2 & 71.3 & 66.1 & 71.3 & 64.4  \\ \midrule
$\mathcal{J}_\text{seen}$ & \first{80.4} & 79.7 & 73.5 & 71.4 & 60.1 & 59.8 & \first{78.3} & - & 72.2 & 67.8& 71.3 & 71.0 \\ 
$\mathcal{J}_\text{unseen}$ & \first{76.4} & 72.8 & 66.2 & 56.5 & 46.1 & 54.2 & \first{75.6} & - & 64.5 & 61.2& 65.5 & 55.5  \\ \midrule
$\mathcal{F}_\text{seen}$ & \first{84.9} & 84.2 & - & - &  62.7 &  60.5 & \first{82.3} & - & 76.1 & 69.5&75.2 & 70.0  \\
$\mathcal{F}_\text{unseen}$ & \first{84.4} & 80.9 & - & - & 51.4 & 60.7 & \first{84.4} & - & 72.7 & 66.2 &73.1& 61.2  \\
\bottomrule
\end{tabular}
}
\vspace{0mm}
\label{tab:ytvos_results_dset}
\end{table}

\parsection{YouTube-VOS~\cite{xu2018youtube2}}
We evaluate our approach on the YouTube-VOS 2018 validation set, containing 474 sequences and 91 object categories. Out of these, 26 are \emph{unseen} in the training dataset. The benchmark reports Jaccard $\mcJ$ and boundary $\mcF$ scores for \emph{seen} and \emph{unseen} categories. Methods are ranked by the overall $\mcG$-score, obtained as the average of all four scores. 
The results are obtained through the online evaluation server.
Results are reported in Tab.~\ref{tab:ytvos_results_dset}. Among previous approaches, STM~\cite{oh2019video} obtains the highest overall $\mcG$-score of $79.4$. Our approach significantly outperforms STM with a relative improvement of over $2.6\%$, achieving an overall $\mcG$-score of $81.5$. Without the use of additional training data, the performance of STM is notably reduced to an overall $\mcG$-score of $68.2$. FRTM~\cite{robinson2020learning} and AGSS-VOS~\cite{lin2019agss} achieve relatively strong performance of $71.3$ in overall $\mcG$-score, despite only employing YouTube-VOS data for training. Sill, our approach outperforms all previous methods by a $12.5\%$ margin in this setting. Remarkably, this version even outperforms all previous methods trained with additional data, achieving a $\mcG$-score of $80.2$. This clearly demonstrates the strength of our optimization-based few-shot learner.
Furthermore, our approach demonstrates superior generalization capability to object classes that are unseen during training. Our method achieves a relative improvement over AGSS-VOS of $15.4\%$ and $15.5\%$ on the $\mcJ_\text{unseen}$ and $\mcF_\text{unseen}$ scores respectively. This demonstrates that our internal few-shot learner can effectively adapt to novel classes.

\parsection{DAVIS 2017~\cite{perazzi2016davis}}
The DAVIS 2017 validation set contains 30 videos. In addition to our standard training setting (see Sec.~\ref{sec:training_details}), we provide results of our approach when using only the DAVIS 2017 training set and ImageNet initialization for the ResNet-50 backbone.
Methods are evaluated in terms of mean Jaccard $\mcJ$ and boundary $\mcF$ scores, along with the overall score $\mcJF$. 
Results are reported in Tab.~\ref{tab:dv17_results}. 
Our approach achieves similar performance to STM, with only a marginal $0.2$ lower overall score. 
When employing only DAVIS 2017 training data, our approach outperforms all previous approaches, with a relative improvement of $7.4\%$ over the second best method FRTM in terms of $\mcJF$.
In contrast to STM and AGAME~\cite{johnander2018generative}, our approach remains competitive to methods that are trained on large amounts of additional data.

\begin{table}[t!]
\caption{State-of-the-art comparison on the DAVIS 2017 validation dataset.
Our approach is almost on par with the best performing method STM, while significantly outperforming all previous methods with only the DAVIS 2017 training data.}\vspace{-2mm}
\centering
\resizebox{0.99\columnwidth}{!}{%
\centering
\begin{tabular}{l@{~~}cccccccc:ccccccc}
\toprule
 & \multicolumn{8}{c}{\textbf{Additional Training Data}} & \multicolumn{7}{c}{\textbf{Only DAVIS 2017 training}} \\
 &  \textbf{Ours} & STM &  SiamRCNN &  PreMVOS & FRTM & AGAME & FEELVOS &AGSSVOS~ & ~\textbf{Ours}&  STM & FRTM &  AGAME & AGSSVOS \\
 && \cite{oh2019video} & \cite{voigtlaender2019siam} & \cite{luiten2018premvos}& \cite{robinson2020learning} & \cite{johnander2018generative}& \cite{voigtlaender2018feelvos}&\cite{lin2019agss}& & \cite{oh2019video} & \cite{robinson2020learning} & \cite{johnander2018generative} & \cite{lin2019agss}  \\
\midrule
$\mathcal{J \& F}$ & 81.6 & \textbf{81.8} & 74.8 & 77.8 & 76.4 & 70.0 & 71.5&67.4 &\textbf{74.3} & 43.0 & 69.2 & 63.2&66.6 \\ 
$\mathcal{J}$ & 79.1 & \textbf{79.2} & 69.3 &  73.9 &  73.7 & 67.2 &  69.1 & 64.9&\textbf{72.2} & 38.1 & 66.8 & - &63.4 \\ 
$\mathcal{F}$ & 84.1 & \textbf{84.3} & 80.2 & 81.7 & 79.1 & 72.7 & 74.0 & 69.9& \textbf{76.3} & 47.9 &77.9 & -&69.8  \\ 
\bottomrule
\end{tabular}
}
\label{tab:dv17_results}
\vspace{0mm}
\end{table}

\parsection{Bounding Box Initialization}
\begin{table}[t!]
\caption{State-of-the-art comparison with bounding box initialization on YouTube-VOS 2018 and DAVIS 2017 validation. 
Our approach outperforms existing methods on YouTube-VOS, while achieving a $\mcJF$ score on par with state-of-the art on DAVIS.}\vspace{0mm}
\centering
\resizebox{0.7\columnwidth}{!}{%
\newcommand{\first}[1]{\textbf{{#1}}}
\centering
\begin{tabular}{l@{~~~}c@{~~}c@{~~}c@{~~}c@{~~}c|c@{~~}c@{~~}c}
\toprule
& \multicolumn{5}{c}{\textbf{YouTube-VOS 2018}} & \multicolumn{3}{c}{\textbf{DAVIS 2017}} \\
Method &  $\mcG$  & $\mcJ_\text{seen}$  & $\mcJ_\text{unseen}$ & $\mcF_\text{seen}$ & ~$\mcF_\text{unseen}$~~ & ~~$\mathcal{J \& F}$~ & $\mathcal{J}$  & $\mathcal{F}$ \\
\midrule
{\bf Ours}  & \first{70.2} & \first{72.7} & \first{62.5} & \first{75.1} & \first{70.4} & \first{70.6} & \first{67.9} & 73.3\\
Siam-RCNN~\cite{voigtlaender2019siam} & 68.3 & 69.9 & 61.4 & - & - &\first{70.6} & 66.1 & \first{75.0}\\
Siam-Mask~\cite{wang2019fast} & 52.8 & 62.2 & 45.1 & 58.2 & 47.7  &56.4 & 54.3 & 58.5\\  
\bottomrule
\end{tabular}
}
\label{tab:bb_results}
\vspace{0mm}
\end{table}
Finally, we evaluate our approach on VOS with bounding box initialization on YouTube-VOS 2018 and DAVIS 2017 validation sets. Results are reported in Tab.~\ref{tab:bb_results}. We compare with the recent Siam-RCNN~\cite{voigtlaender2019siam} and Siam-Mask~\cite{wang2019fast}. Our approach achieves a relative improvement of $2.8\%$ in terms of $\mcG$-score over the previous best method Siam-RCNN on YouTube-VOS. Despite only using bounding-box supervision, our approach remarkably outperform several recent methods in Tab.~\ref{tab:ytvos_results_dset} employing mask initialization. On DAVIS 2017, our approach is on par with Siam-RCNN with a $\mcJF$-score of $70.6$.
These results demonstrate that our approach can readily generalize to the box-initialization setting thanks to the flexible internal target representation.
\section{Conclusions}
We present a novel VOS approach by integrating a optimization-based few-shot learner. Our internal learner is differentiable, ensuring an end-to-end trainable VOS architecture. Moreover, we propose to learn what the few-shot learner should learn. This is achieved by designing neural network modules that predict the ground-truth label and importance weights of the few-shot objective.  This allows the target model to predict a rich target representation, guiding our VOS network to generate accurate segmentation masks.

\noindent\textbf{Acknowledgments}: This work was partly supported by the ETH Z\"urich Fund (OK), a Huawei Technologies Oy (Finland) project, an Amazon AWS grant, and Nvidia.
\bibliographystyle{splncs04}
\bibliography{references}

\clearpage

\setcounter{section}{0}

\renewcommand{\thesection}{\Alph{section}}

\begin{center}
	\textbf{\large Appendix}
\end{center}

In this appendix, we provide additional results and further details about our method. First, in Section~\ref{sec:derivation}, we provide a derivation of the steepest descent iterations in Eq.~(4). Next, we present more details about our label generator, weight predictor, box decoder and decoder modules in Section~\ref{sec:supp_arch}. Section~\ref{sec:sup_inference} and Section~\ref{sec:sup_training} provide additional details about our inference and training procedures, respectively. A comparison of our approach on the YouTube-VOS 2019 validation set is provided in Section~\ref{sec:sup_sota}. We further provide detailed ablative analysis of our training and inference parameters in Section~\ref{sec:sup_abl}. Finally, in Section~\ref{sec:sup_qual}, we provide additional qualitative results, including outputs generated with our box initialization setting and visualization of the mask encoding outputs on a few example sequences. 

\section{Derivation of Internal Learner Iteration Steps}
\label{sec:derivation}
In this section we derive the steepest decent iterations in Eq.~(4) used in our few-shot learner to minimize the loss in Eq.~(3). To simplify the derivation, we first convert the loss into a matrix formulation. We then derive expressions for the vectorized gradient $\bar{g}$ and step-length $\alpha$, showing that these can be computed using simple neural network operations.

We use the fact that the convolution between the feature map $x_t \in \reals^{H \times W \times C}$ and weights $\targetparam \in \reals^{K \times K \times C \times D}$ can be written in matrix form as \mbox{$\vecop(x_t \conv \targetparam) = X_t \targetparamvec$}. Here, $\vecop$ is the vectorization operator, $\targetparamvec=\vecop(\targetparam)\in \reals^{K^2CD}$ and $X_t \in \reals^{H W D \times K^2 C D}$ is a matrix representation of $[x_t \conv]$. We further define $e_t = \vecop(\labelenc(\maskgt_t))\in \reals^{HWD}$ as a vectorization of the label encoding and $W_t = \diag ( \vecop( \weightpredictor(\maskgt_t)))\in \reals^{HWD \times HWD}$ is a diagonal matrix corresponding to the point-wise multiplication of the importance weights $\weightpredictor(\maskgt_t)$. We can now write Eq.~(3) in matrix form as,
\begin{equation}
\label{eq:L2_mat}
L(\targetparamvec)= \frac{1}{2}\sum_{t}\big\|W_t (X_t \targetparamvec - e_t)\big\|^2 +  \frac{\lambda}{2}\big\|\targetparamvec\big\|^2\,.
\end{equation}

In the steepest descent algorithm, we update the parameters by taking steps $\targetparamvec^{i+1} = \targetparamvec^i - \alpha^i \bar{g}^i$ in the gradient direction $\bar{g}^i$ with step length $\alpha^i$. Setting $\resvec_t(\targetparamvec)=W_t (X_t \targetparamvec - e_t)$, the gradient is obtained using the chain rule,
\begin{equation}
\bar{g}=\nabla L(\targetparamvec_d) = \sum_t \Big(\frac{\partial \resvec_t}{\partial \targetparamvec}\Big)\tp \resvec_t(\targetparamvec) + \lambda\targetparamvec = \sum_t X_t\tp W_t^2 \big(X_t \targetparamvec - e_t\big) + \lambda\targetparamvec\,.
\label{eq:grad_mat}
\end{equation}
We see that the gradient can be computed as, 
\begin{align}
\bar{g}=&\sum_t X_t\tp W_t^2 \big(\vecop(x_t \conv \targetparam) - e_t\big) + \lambda \vecop(\targetparam) \nonumber \\
=&\vecop\left(\sum_t  x_t \conv\tp \weightpredictor(\maskgt_t)^2 \cdot \big(x_t \conv \targetparam - \labelenc(\maskgt_t)) + \lambda \targetparam\right) \,,
\label{eq:grad_mat2}
\end{align}
where the transposed convolution $x_t\conv\tp$ corresponds to the matrix multiplication with $X_t\tp$. Thus,
\begin{equation}
\label{eq:sd-g}
g = \sum_t x_t \conv\tp \Big(\weightpredictor(\maskgt_t)^2 \cdot \big(x_t \conv \targetparam - \labelenc(\maskgt_t)\big) \Big) + \lambda \targetparam \,.
\end{equation}

We compute the step length $\alpha^i$ that minimizes $L$ in the current gradient direction $g^i$
\begin{equation}
\label{eq:L_alpha}
\alpha^i=\argmin_\alpha L(\targetparam^i - \alpha \bar{g}^i)\,. 
\end{equation}
Since the loss is convex, it has an unique global minimum obtained by solving for the stationary point \mbox{$\frac{\diff L(\targetparamvec^i - \alpha \bar{g}^i)}{\diff \alpha} = 0$}. We set $v=\targetparamvec^i - \alpha \bar{g}^i$, and use \eqref{eq:grad_mat} with the chain rule to obtain,
\begin{align}
0 = \frac{\diff L(v)}{\diff \alpha} &= \Big(\frac{\diff v}{\diff \alpha}\Big)\tp \nabla_v L(v) \nonumber \\ 
& = (\bar{g}^i)\tp \bigg( \sum_t X_t\tp W_t^2 \big(X_t (\targetparamvec^i - \alpha \bar{g}^i) - e_t\big) + \lambda(\targetparamvec^i - \alpha \bar{g}^i) \bigg) \nonumber \\
&=(\bar{g}^i)\tp \bar{g}^i - \alpha (\bar{g}^i)\tp \bigg(\sum_t X_t\tp W_t^2 X_t \bar{g}^i + \lambda \bar{g}^i\bigg) \nonumber \\ 
&= \|\bar{g}^i\|^2 - \alpha \bigg(\sum_t \| W_t X_t \bar{g}^i \|^2 + \lambda \|\bar{g}^i\|^2\bigg)\,.
\end{align}
Thus, the step length is obtained as, 
\begin{equation}
\alpha = \frac{\big\|\bar{g}^i\big\|^2}{\sum_t\big\|W_t X_t \bar{g}^i\big\|^2+\lambda \big\|\bar{g}^i\big\|^2}\,.
\end{equation}
We note that, $\big\|\bar{g}^i\big\|^2 = \big\|g^i\big\|^2$ and $\big\|W_t X_t \bar{g}^i\big\|^2 = \big\|\vecop(\weightpredictor(\maskgt_t) \cdot x_t \conv g^i)\big\|^2=\big\|\weightpredictor(\maskgt_t) \cdot (x_t \conv g^i)\big\|^2$. The step length can therefore be computed as follows, 
\begin{equation}
\label{eq:sd-steplength}
\alpha = \frac{\big\|g^i\big\|^2}{\sum_t\big\|\weightpredictor(\maskgt_t) \cdot (x_t \conv g^i)\big\|^2+\lambda \big\|g^i\big\|^2}\,.
\end{equation}
\section{Architecture Details}
\label{sec:supp_arch}
\subsection{Few-shot Label Generator $\labelenc$ and Weight Predictor $\weightpredictor$}
\label{sec:supp_arch_label}
\begin{figure}[t]
    \centering
    \includegraphics[width=0.8\columnwidth]{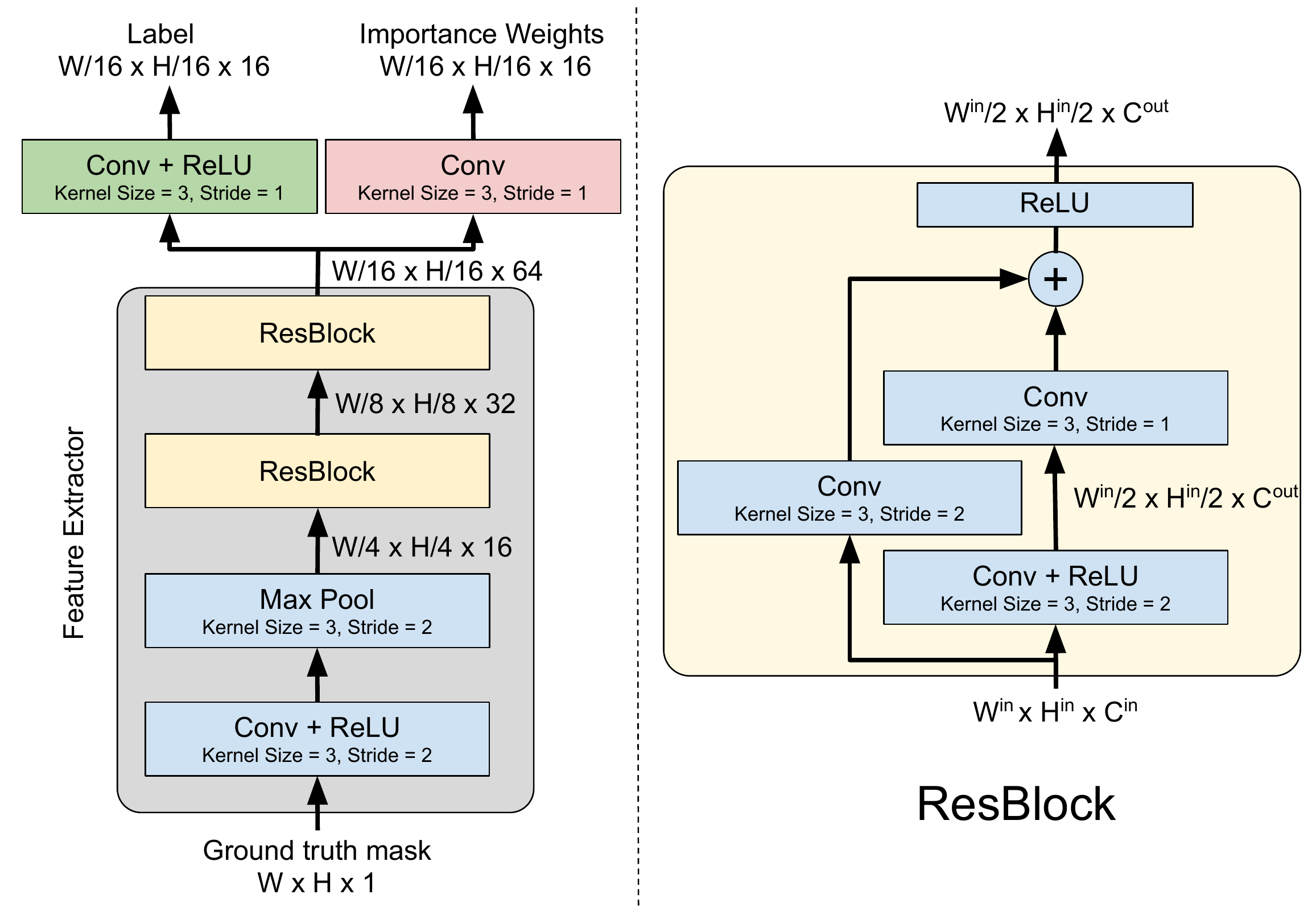}
    \vspace{-3mm}
    \caption{The network architecture employed for the label generator $\labelenc$ and the importance weight predictor $\weightpredictor$ modules. Both modules share a common feature extractor (gray) consisting of a convolutional layer followed by two residual blocks (yellow). The feature extractor takes as input the ground truth segmentation mask, and outputs a deep representation of the mask containing $64$ channels. The label generator $\labelenc$ (green), consisting of a single convolutional layer followed by ReLU activation, generates the ground truth label for the few-shot learner using the mask features as input. Similarly, the importance weight predictor $\weightpredictor$ (red), which consists of a single convolutional layer, predicts the importance weights using the mask features as input.} 
    \vspace{0mm}
    \label{fig:label_enc}
\end{figure}

Here, we describe in detail the network architecture employed for the label generator $\labelenc$ and the importance weight predictor $\weightpredictor$. The network architecture is visualized in Figure~\ref{fig:label_enc}. The label generator $\labelenc$ and the importance weight predictor $\weightpredictor$ share a common feature extractor consisting of a convolutional layer followed by two residual blocks. The feature extractor takes as input the ground truth segmentation mask, and outputs a deep representation of the mask. The mask features contain $64$ channels, and have a spatial resolution $16$ times lower than the input mask. The label generator module $\labelenc$, which consists of a single convolutional layer followed by a ReLU activation, operates on the mask features to predict the ground truth label for the few-shot learner. Similarly, the importance weight predictor $\weightpredictor$ consists of a single convolutional layer and predicts the importance weights using the mask features as input. All the convolutional layers in the network employ $3\times 3$ kernels. Note that the importance weights $\weightpredictor(\maskgt)$ are squared when computing the squared error between the target module output and the ground truth label $\labelenc(\maskgt)$ predicted by the label generator $\labelenc$. Thus, we thus allow $\weightpredictor(\maskgt)$ to take negative values.

\subsection{Segmentation Decoder $\decoder$}
\label{sec:supp_arch_decoder}
\begin{figure}[t]
    \centering
    \includegraphics[width=\columnwidth]{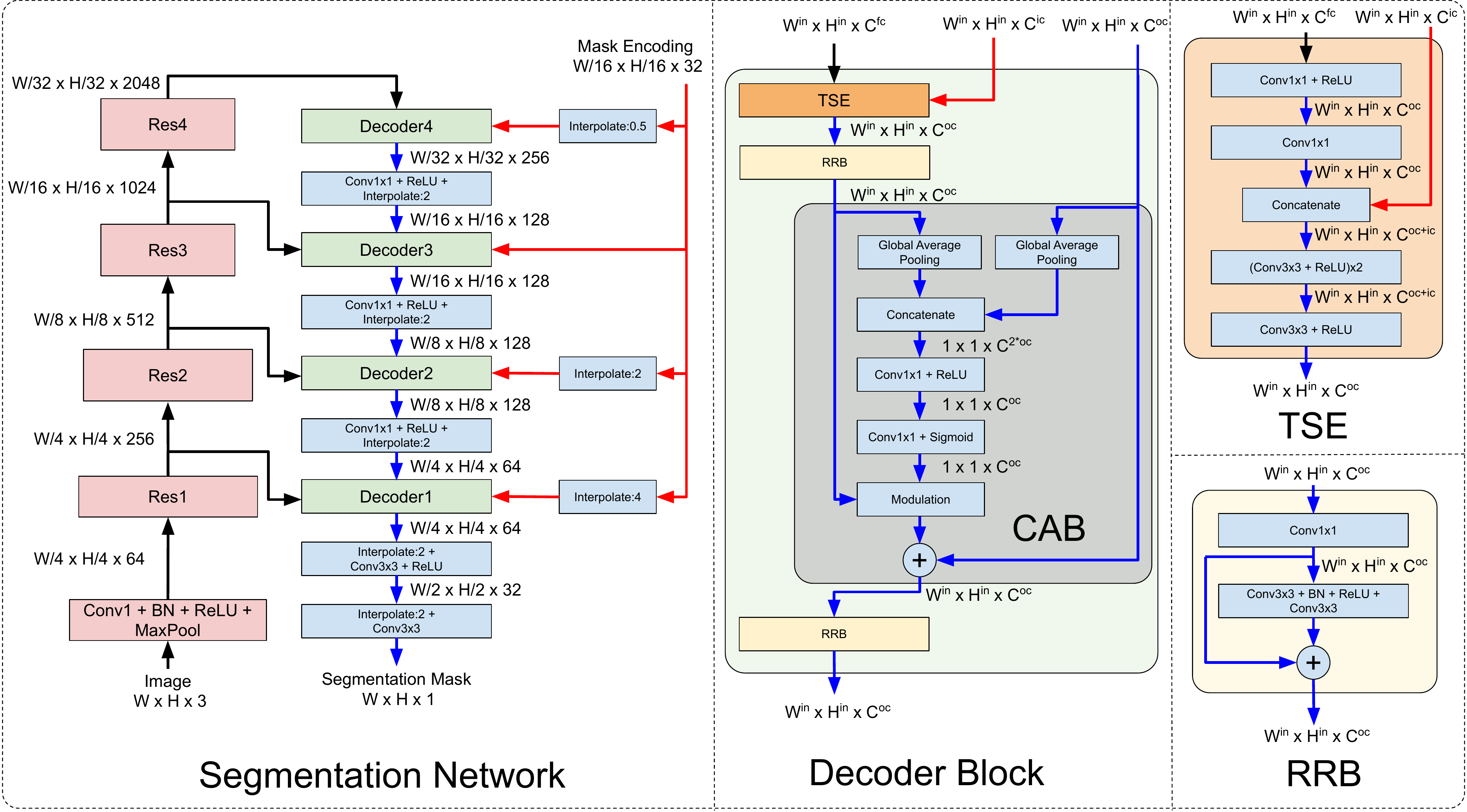}
    \vspace{-4mm}
    \caption{The network architecture employed by our segmentation decoder $\decoder$ module. The segmentation decoder takes as input the mask encoding output by the target module $\targetlayer$ (red arrow), along with the output of the four residual blocks of the backbone ResNet-50 (black arrow). The network has a U-Net based structure containing four decoder blocks (green) corresponding to the residual blocks in the ResNet-50 feature extractor. The TSE module (orange) in each decoder block first projects the backbone features to a lower-dimensional representation, which is concatenated with the mask encoding. We interpolate the mask encodings to the same spatial size as the backbone features before concatenations. The concatenated features are processed by three convolutional layers followed by a residual block (yellow). The resulting features are then merged with features from a deeper decoder module with a channel attention block (CAB)~\cite{yu2018dfn} (gray). Finally, the features are processed by another residual block before being passed to the next decoder level. The output from the final decoder block is up-sampled and processed by convolutional layer to obtain the final segmentation mask.} 
    \vspace{0mm}
    \label{fig:segdec}
\end{figure}
Here, we detail the segmentation decoder $\decoder$ architecture, visualized in Figure~\ref{fig:segdec}. We adopt a similar architecture as in~\cite{robinson2020learning}. The decoder module takes the mask encoding output by the target module $\targetlayer$, along with backbone ResNet features, in order to predict the final accurate segmentation mask. It has a U-Net based structure containing four decoder blocks corresponding to the residual blocks in the ResNet feature extractor. In each decoder block, we first project the backbone features into a lower-dimensional representation. Next, we concatenate the projected feature maps with the mask encoding output by the $\targetlayer$. These are processed by three convolutional layers followed by a residual block. The resulting features are then merged with features from a deeper decoder module with a channel attention block (CAB)~\cite{yu2018dfn}. Finally, the features are processed by another residual block before they are merged with features from a shallower level. The output from the final decoder module is up-sampled and projected to a single channel target segmentation mask.

\begin{figure}[t]
    \centering
    \includegraphics[width=0.8\columnwidth]{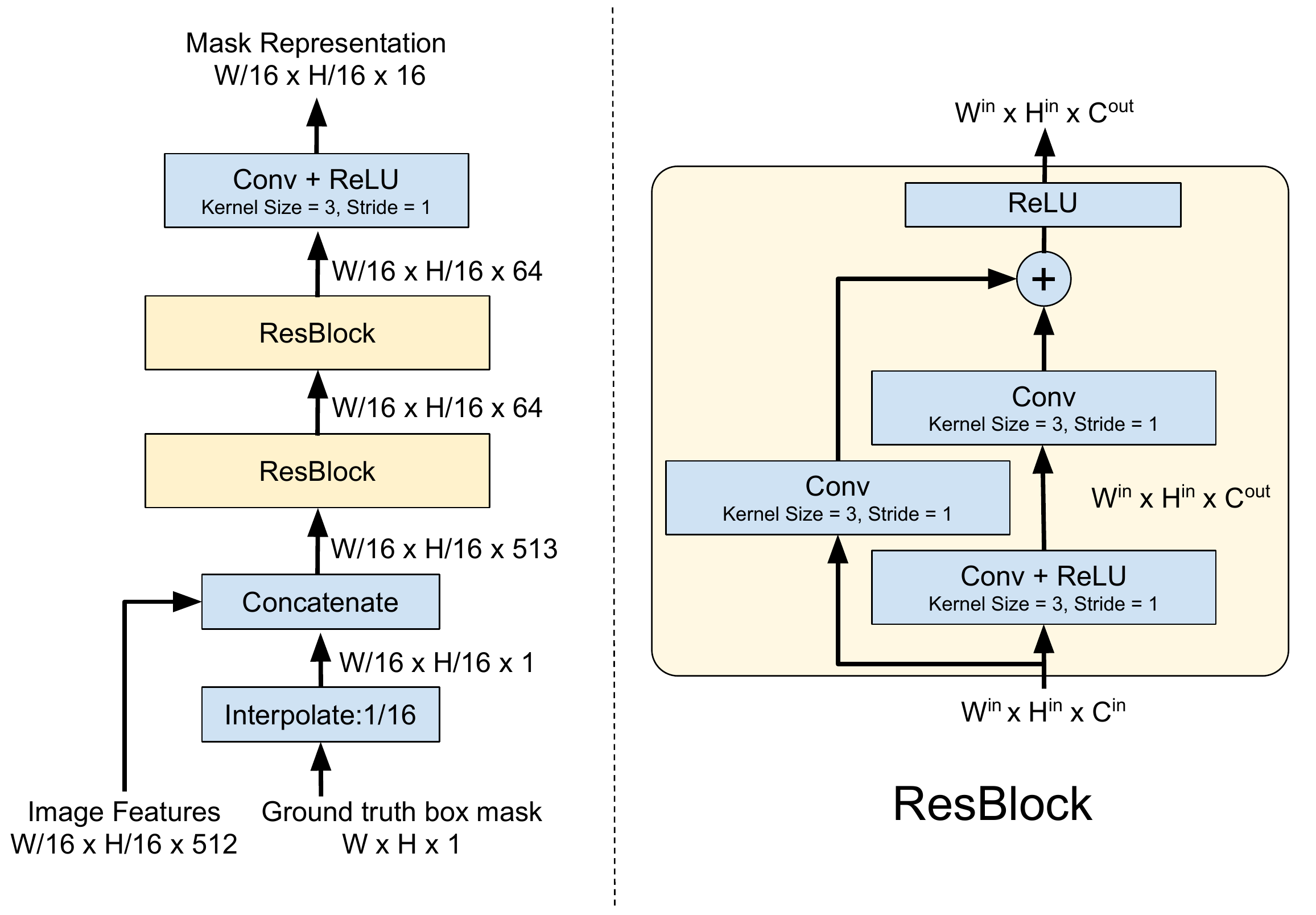}
    \vspace{-4mm}
    \caption{The network architecture employed for the bounding box encoder $\boxenc(b_0, x_0)$. The network takes as input a mask $b_0$ denoting the input ground truth box, along with a deep feature representation $x_0$ of the image. The input mask is first downsampled by a factor of $16$ and concatenated with the image features. These are then processed by two residual blocks (yellow). The output of the second residual block is passed to a convolutional block which predicts the mask representation of the target object. This mask representation is input to the segmentation decoder module $\decoder$ to obtain the segmentation mask for the target.}
    \vspace{0mm}
    \label{fig:box_enc}
\end{figure}

\subsection{Bounding Box Encoder $\boxenc$}
In this section we give a detailed description of the network architecture of the box encoder module. We provide an illustration of the architecture in Figure~\ref{fig:box_enc}. The network takes a mask representation of the bounding box along with features from layer3 from the backbone feature extractor as input. The mask is downsampled with bilinear interpolation to a 1/16th of the input resolution, to match the size of the backbone features. The backbone features are first processed by a convolutional layer that reduces the dimension to $C = 512$. Here, the weights of this convolutional layer is shared with the projection layer for the target model. The resulting features are first concatenated with the downsampled mask and then fed through a residual block, which also reduces the feature dimension to 64 channels. Next, the features are processed by an another residual block, before the final box encoding is generated by a convolutional layer. This convolutional layer reduces the number of dimensions to coincide with number of channels in the mask representation produced by the label generator module. The output of the box encoder can then be processed by the decoder network to produce a segmentation mask.

\section{Inference Details}
\label{sec:sup_inference}
In this section, we provide more details about our inference procedure. Instead of operating on the full image, we process only a local region around the previous target location in each frame. This allows us to effectively segment objects of any size. The local search region is obtained by cropping a patch that is $5$ times larger than the previous estimate of target, while ensuring the maximal size to be equal to the image itself. The cropped region is resized to $832 \times 480$ with preserved aspect ratio. An estimate of the target location and size is obtained from the predicted segmentation mask, as detailed in Algorithm~\ref{alg:mask_to_box}. The target center is determined as the center of mass of the predicted target mask, while the target size is computed using the variance of the segmentation mask. We additionally prevent drastic changes in the target size between two consecutive frames by limiting the target scale change between two frames to be in the range $[0.95, 1.1]$. This allows the inference to be robust to incorrect mask predictions in one or few frames.

\algrenewcommand{\algorithmicrequire}{\textbf{Input:}}
\algrenewcommand{\algorithmicensure}{\textbf{Output:}}
\begin{algorithm}[t]
\newcommand{\maskp}{m}
	\caption{Segmentation Mask to Target Box}
	\begin{algorithmic}[1]
    	\Require Mask prediction $\maskp(\vecn{r}) \in [0,1]$ for every pixel $\vecn{r} \in \Omega \defeq \{0,\dots,W - 1\} \times \{0,\dots,H - 1\}$ in the image, Previous target size $\vecn{\hat{b}}= (\hat{b}_w, \hat{b}_h)$
		\Ensure Current target location $\vecn{c}=(c_x, c_y)$ and size $\vecn{b}=(b_w, b_h)$
		\State $z = \sum_{\vecn{r} \in \Omega} \maskp(\vecn{r})$     \hspace{25.5mm}\algcomment{Compute normalization factor}
		\State $\vecn{c} = \frac{1}{z}\sum_{\vecn{r} \in \Omega} \vecn{r} \cdot  \maskp(\vecn{r})$  \hspace{19mm}\algcomment{Estimate target center}
		\State $\sigma^2 = \frac{1}{z}\sum_{\vecn{r} \in \Omega} (\vecn{r} - \vecn{c})^2 \cdot \maskp(\vecn{r})$  \hspace{8.0mm}\algcomment{Estimate target size}
		\State $\vecn{b} = 4\sigma$
		
		\State $\Delta_\text{size} = \sqrt{\frac{b_w b_h}{\hat{b}_w \hat{b}_h}}$ \hspace{26mm}\algcomment{Change in target size from prev. frame}
	
		\State $\Delta_\text{size} = \min(\max(\Delta_\text{size}, 0.95), 1.1)$ \hspace{0mm}\algcomment{Limit change in target size}
            \State $\vecn{b} = \Delta_\text{size}\vecn{\hat{b}}$
	\end{algorithmic}
	\label{alg:mask_to_box}
\end{algorithm}

\section{Training Details}
\label{sec:sup_training}
Here, we provide more details about our offline training procedure.  Our network is trained using the YouTube-VOS 2019 training set (excluding the 300 validation videos) and the DAVIS 2017 training set. We sample sequences from both datasets without replacement, using a 6 times higher probability for YouTube-VOS, as compared to the DAVIS 2017 training set, due to the higher number of sequences in the former dataset. 
Our network is trained using the ADAM~\cite{kingma2014adam} optimizer.

Our final networks, that are used for state-of-the-art comparisons, are trained using the \emph{long} strategy. In this setting, the networks are trained for $170k$ iterations in total, with a base learning rate of $10^{-2}$. The learning rate is reduced by a factor of $5$ after $40$k, $95$k, and $145$k iterations. For the first $70$k iterations, we freeze the weights of our backbone feature extractor, and only train the newly added layers. The complete network, excluding the first convolutional and residual blocks in the feature extractor, are then trained for the remaining $100$k iterations. We use a mini-batch size of $20$ throughout our training. For evaluation on the DAVIS dataset, we additionally fine-tune our network for $2k$ more iterations using only the DAVIS 2017 training set. The entire training takes $48$ hours on 4 Nvidia V100 GPUs.

Due to resource constraints, we use a \emph{shorter} schedule when training different versions of our proposed approach for the ablation study. Here, we train the network for $70$k iterations, using a mini-batch size of 10. We use a base learning rate of $10^{-2}$, which is reduced by a factor of $5$ after $25$k, and $50$k iterations. In this training setting, we keep the weights of the backbone feature extractor fixed and only train the newly added layers. The entire training takes $24$ hours on a single Nvidia V100 GPU.

The bounding box encoder is trained on YouTube-VOS 2019 (excluding the 300 validation videos) and MSCOCO~\cite{lin2014microsoft}. The mini-batches are constructed by sampling images with twice as high probability from MSCOCO compared to YouTube-VOS. We train the network for $50k$ iterations with a batch size of $8$, only updating the weights of the convolutional layers in the box encoder. We use a base learning rate of $10^{-2}$ and reduce it by a factor of $5$ after $20$k and $40$k iterations.

\section{YouTube-VOS 2019}
\label{sec:sup_sota}
\begin{table}[t!]
\caption{Comparison of our approach with the recently introduced STM~\cite{oh2019video} on the large-scale YouTube-VOS 2019 validation dataset. Results are reported in terms of mean Jaccard ($\mcJ$) and boundary ($\mcF$) scores for object classes that are \emph{seen} and \emph{unseen} in the training set, along with the overall mean ($\mcG$). Our approach outperforms STM with a large margin of $+1.8$ points in terms of the overall $\mcG$ score.}\vspace{0mm}
\centering
\resizebox{0.60\columnwidth}{!}{%
\centering
\begin{tabular}{l@{~~~~}ccc}
\toprule
       & $\mathcal{G} (\%)$ & $\mathcal{J} (\%)$ & $\mathcal{F} (\%)$\\
Method &  overall &   ~~~seen $|$ unseen &  ~~~seen $|$ unseen  \\
\midrule
Ours  & 81.0 & 79.6 $|$ 76.4 & 83.8 $|$ 84.2\\
STM \cite{oh2019video} & 79.2 & 79.6 $|$ 73.0 & 83.6 $|$ 80.6\\
\bottomrule
\end{tabular}

}\vspace{0mm}
\label{tab:yt2019}
\end{table}

We evaluate our approach on the YouTube-VOS 2019 validation set consisting of 507 sequences. The dataset contains 1063 unique object instances belonging to 91 object categories, of which 26 are \emph{unseen} in the training dataset. The results are obtained through the online evaluation server. The benchmark reports Jaccard $\mcJ$ and boundary $\mcF$ scores for \emph{seen} and \emph{unseen} categories. Methods are ranked by the overall $\mcG$-score, obtained as the average of all four scores. We compare our approach with results shown on the leaderboard of the evaluation server for the recently introduced STM~\cite{oh2019video} method. The results are shown in Table~\ref{tab:yt2019}. Our approach achieves at overall $\mcG$ score of $81.0$, outperforming STM with a large margin of $+1.8$.

\section{Detailed Ablative Study}
\label{sec:sup_abl}
In this section, we analyze the impact different components in our architecture. As in the main paper, our analysis is performed on the validation set of 300 sequences generated from the YouTube-VOS 2019 training set. Unless otherwise mentioned, we use the shorter training schedule (see Section~\ref{sec:sup_training}) for training the networks compared in this section and the default inference parameters (see Section~3.5) in the evaluations. The networks are evaluated using the mean Jaccard $\mcJ$ index (IoU).

\parsection{Impact of initial backbone weights} Here, we analyse the impact of using the Mask-RCNN~\cite{He2017MaskR} weights for initializing our backbone feature extractor. We compare our approach with a baseline network which uses ImageNet~\cite{imagenet_cvpr09} pre-trained weights for initializing the backbone. The results of this comparison is shown in Table \ref{tab:ablation_initw}. Using the Mask-RCNN weights for initializing the backbone feature extractor provides an improvement of $+1.2$ in $\mcJ$ score over the version which uses ImageNet trained weights.

\parsection{Impact of training loss} We investigate the impact of using the Lovasz~\cite{berman2018lovasz} loss as our segmentation loss in Eq.~(5) by evaluating a version which uses the binary cross-entropy (BCE) loss. The results in Table~\ref{tab:ablation_loss} show that using the Lovasz loss provides as improvement of $+0.6$ in $\mcJ$ score over the baseline trained using the BCE loss.

\begin{table}[t!]
\caption{Impact of the weights used for initializing the backbone feature extractor. We compare a network using Mask-RCNN weights for initializing the backbone feature extractor with a network using ImageNet pre-trained weights. The results are reported over a validation set of $300$ videos sampled from YouTube-VOS 2019 training set, in terms of mean Jaccard $\mcJ$ score.}\vspace{0mm}
\centering
\resizebox{0.35\columnwidth}{!}{%
\renewcommand{\yes}{\text{\ding{51}}}
\renewcommand{\no}{-}
\centering
\begin{tabular}{l@{~~~~}c}
\toprule
 & $\mathcal{J} (\%)$ \\
\midrule
Mask-RCNN weights&79.8 \\
ImageNet weights&78.6\\
\bottomrule
\end{tabular}
}\vspace{0mm}
\label{tab:ablation_initw}
\end{table}

\begin{table}[t!]
\caption{Impact of the segmentation loss employed during training. We compare a network trained using the Lovasz~\cite{berman2018lovasz} loss function, with a network trained using the binary cross-entropy loss.}\vspace{0mm}
\centering
\resizebox{0.40\columnwidth}{!}{%
\renewcommand{\yes}{\text{\ding{51}}}
\renewcommand{\no}{-}
\centering
\begin{tabular}{l@{~~~~}c}
\toprule
 & $\mathcal{J} (\%)$ \\
\midrule
Lovasz Loss&79.8 \\
Binary Cross-Entropy Loss&79.2 \\
\bottomrule
\end{tabular}
}\vspace{0mm}
\label{tab:ablation_loss}
\end{table}

\parsection{Impact of long training} Here, we analyze the impact of using the long training procedure, as described in Section~\ref{sec:sup_training}. Table~\ref{tab:ablation_fulltraining} shows a comparison between the shorter and longer training strategy. The long training provides an improvement of $+1.4$ in $\mcJ$ score over the version using the shorter training, albeit taking 8 times more GPU hours for training.

\begin{table}[t!]
	\caption{Comparison of a network trained using the shorter training strategy (see Section~\ref{sec:sup_training}) with the network trained using the long training strategy. In contrast to the shorter training strategy, the backbone feature extractor is also trained when using the long training strategy, while employing a larger batch size. The long training however requires 8 times more GPU hours, as compared to the shorter training.}\vspace{0mm}
	\centering
	\resizebox{0.40\columnwidth}{!}{%
		\renewcommand{\yes}{\text{\ding{51}}}
\renewcommand{\no}{-}
\centering
\begin{tabular}{l@{~~~~}c}
\toprule
 & $\mathcal{J} (\%)$ \\
\midrule
Short training strategy&79.8 \\
Long training strategy&81.2 \\
\bottomrule
\end{tabular}
	}\vspace{0mm}
	\label{tab:ablation_fulltraining}
\end{table}

\begin{figure}[t]
    \centering
    \includegraphics[width=0.6\columnwidth]{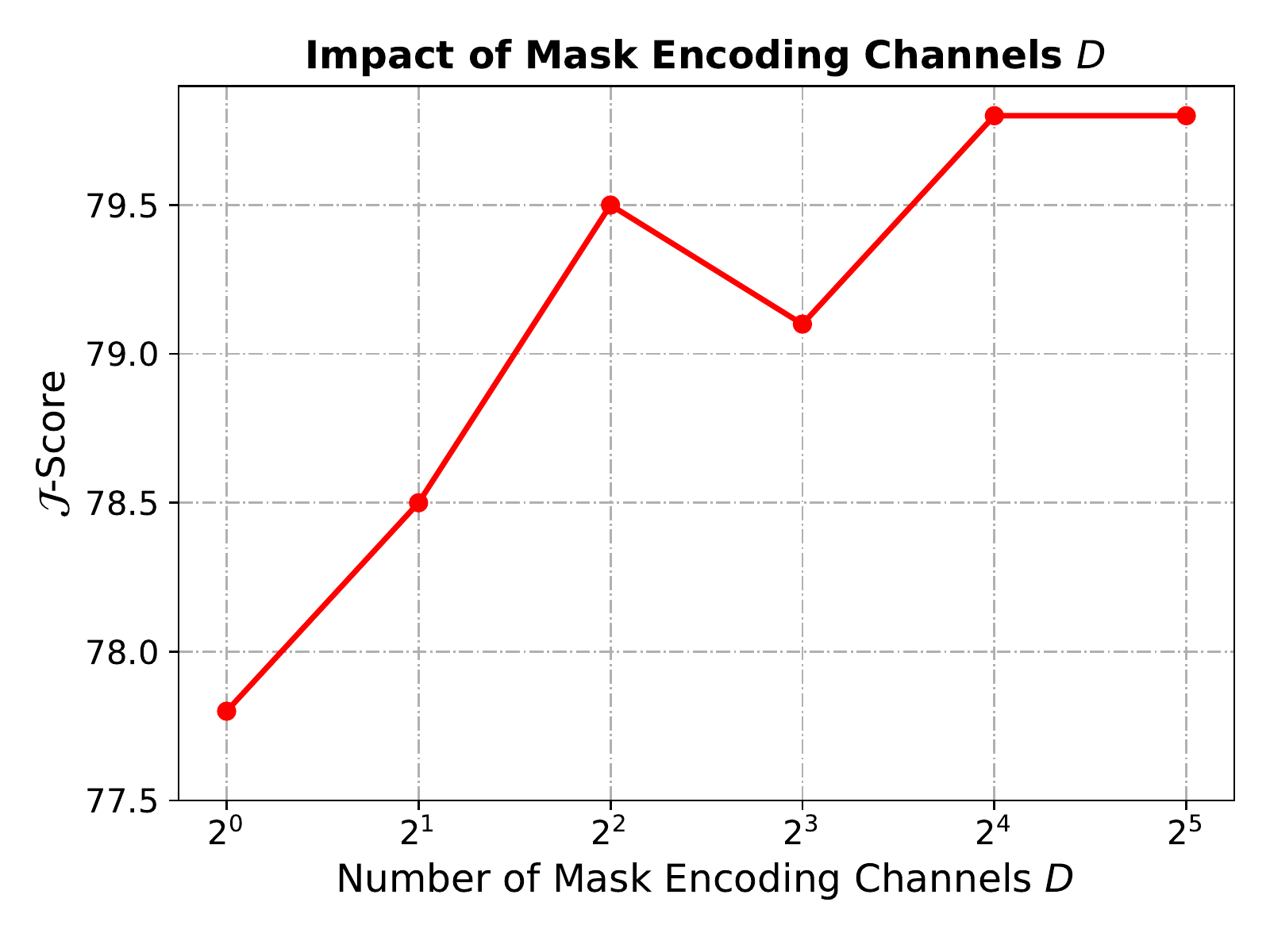}
    \vspace{-6mm}
    \caption{Impact of the number of output channels $D$ in the mask encoding $\targetlayer(x)$ predicted by the target module $\targetlayer$. We plot the $\mcJ$ score (y-axis) computed over a validation set of $300$ videos sampled from YouTube-VOS 2019 training set for different values of mask encoding channels $D$ (x-axis). Using a larger number of channels ($>= 4$) allows the target module to output a richer representation of the target mask, leading to improved results over the baseline employing a single channel.} 
    \vspace{0mm}
    \label{fig:num_filters}
\end{figure}

\parsection{Impact of number of mask encoding channels $D$} We investigate the impact of the number of output channels $D$ in the mask encoding $\targetlayer(x)$ predicted by the target module $\targetlayer$. The $\mcJ$ score for different values of mask encoding channels $D$ are plotted in Figure \ref{fig:num_filters}. Using a larger number of channels ($\geq 4$) allows the target module to output a richer representation of the target mask, leading to an improvement of $+2.0$ in $\mcJ$ score over the baseline employing a single channel.

\begin{figure*}[t]
	\newcommand{\wid}{0.5\textwidth}%
	\centering\vspace{0mm}%
	\subfloat[Impact of update iterations\label{fig:num_iter}]{\includegraphics[width = \wid]{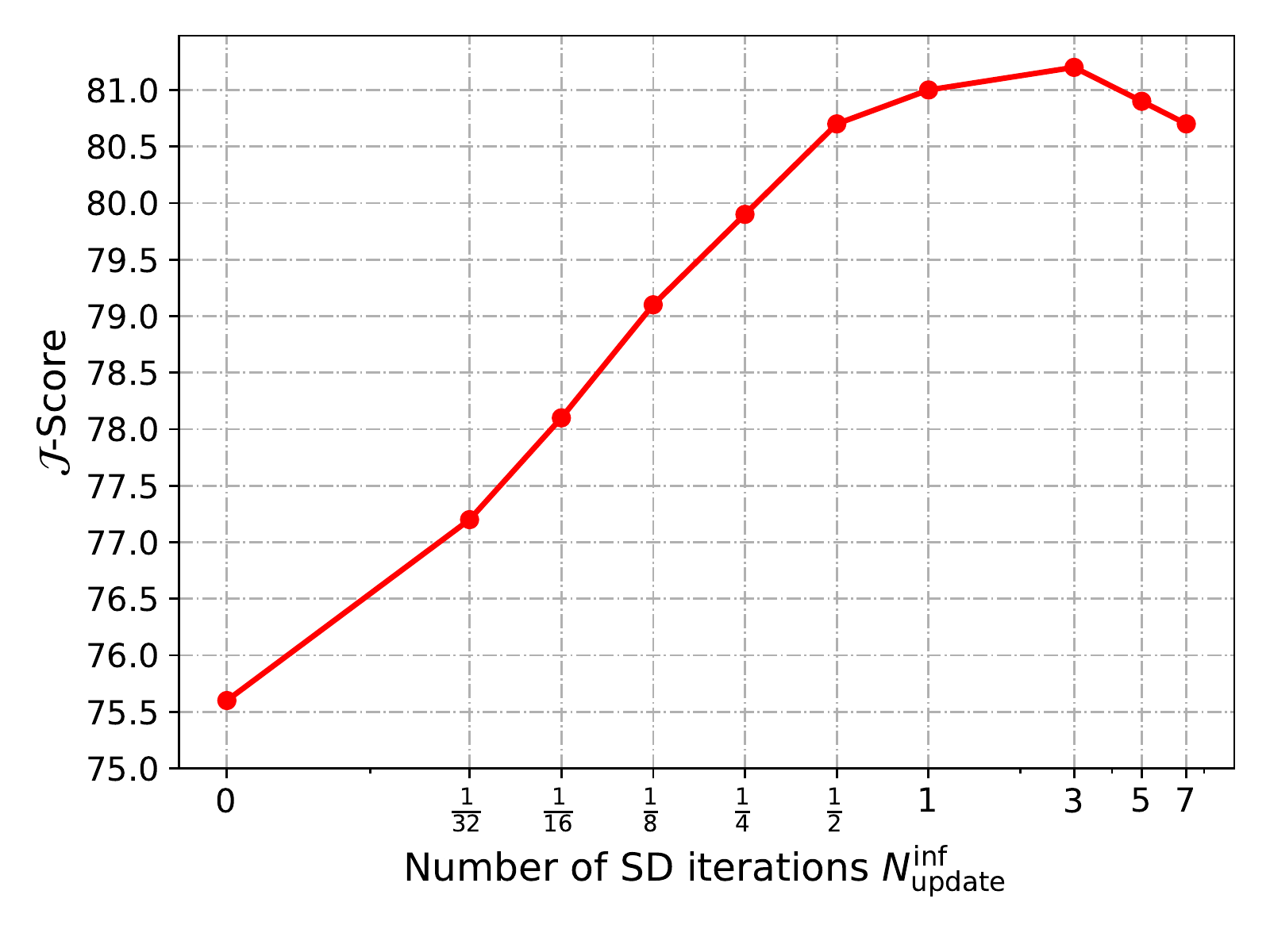}}%
	\subfloat[Impact of update rate $1 - \eta$\label{fig:lr}]{\includegraphics[width = \wid]{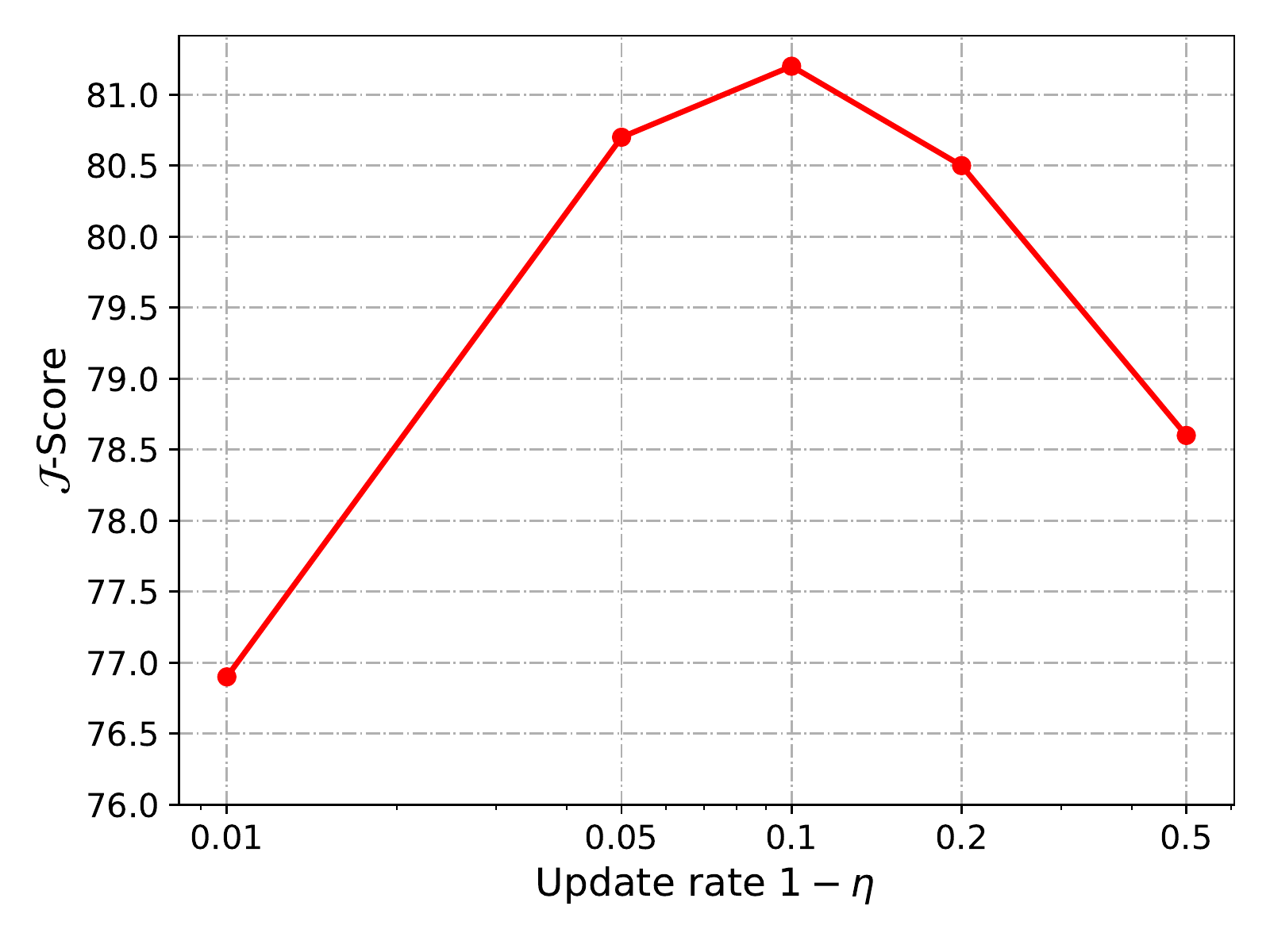}}\vspace{-2mm}%
	\caption{Impact of the number of steepest-descent update iterations (a), and the update rate $1 - \eta$ (b) employed during inference. In plot (a), fractional values $\frac{1}{n}$ for the update iterations $N_\text{update}^\text{inf}$ imply that a single steepest-descent iteration is performed after every $n$ frames. The results are shown in terms of the mean Jaccard $\mcJ$ score over a validation set of 300 videos randomly sampled from the YouTube-VOS 2019 training set.
	}
	\label{fig:ablation_inf}\vspace{-1mm}
\end{figure*}

\parsection{Impact of number of update iterations} We analyze the impact of number of steepest-descent (SD) iterations $N_\text{update}^\text{inf}$ employed during inference. The analysis is performed using the final network trained using the long training strategy. The $\mcJ$ score for different number of update iterations $N_\text{update}^\text{inf}$ are plotted in Figure \ref{fig:num_iter}. The fractional values $\frac{1}{n}$ for the update iterations $N_\text{update}^\text{inf}$ in Figure \ref{fig:num_iter} imply that a single steepest-descent iteration is performed after every $n$ frames. The setting $N_\text{update}^\text{inf}=0$ corresponds to not updating the target model parameters $\targetparam$ during inference. That is, the target model estimated using the initial frame is used for the whole sequence. We observe that performing only a single steepest-descent iteration in each frame significantly improves the performance by more than $5$ points in $\mcJ$ score, compared to the version with no model update. This demonstrates that our few-shot learner can quickly adapt the target model to the changes in the scene, thanks to the fast convergence of steepest-descent updates. The setting $N_\text{update}^\text{inf}=3$ provides the best performance with a $\mcJ$ score of $81.2$. Performing a higher number of SD iterations ($> 3$) results in overfitting of the target model to the training set, leading to a slight degradation in performance.

\parsection{Impact of update rate} Here, we investigate the impact of the update rate $1 - \eta$ employed when setting the global importance weights for the samples in the few-shot training set $\basedset$ (see Section~3.5). Figure~\ref{fig:lr} shows the $\mcJ$ score for different values of the update rate $1 - \eta$. A higher update rate implies that the recent samples get a larger global importance weights when updating the target model $\targetparam$ during inference. We observe that an update rate of $0.1$ gives the best results with a $\mcJ$ score of $81.2$. 

\begin{figure}[t]
    \centering
    \includegraphics[width=0.6\columnwidth]{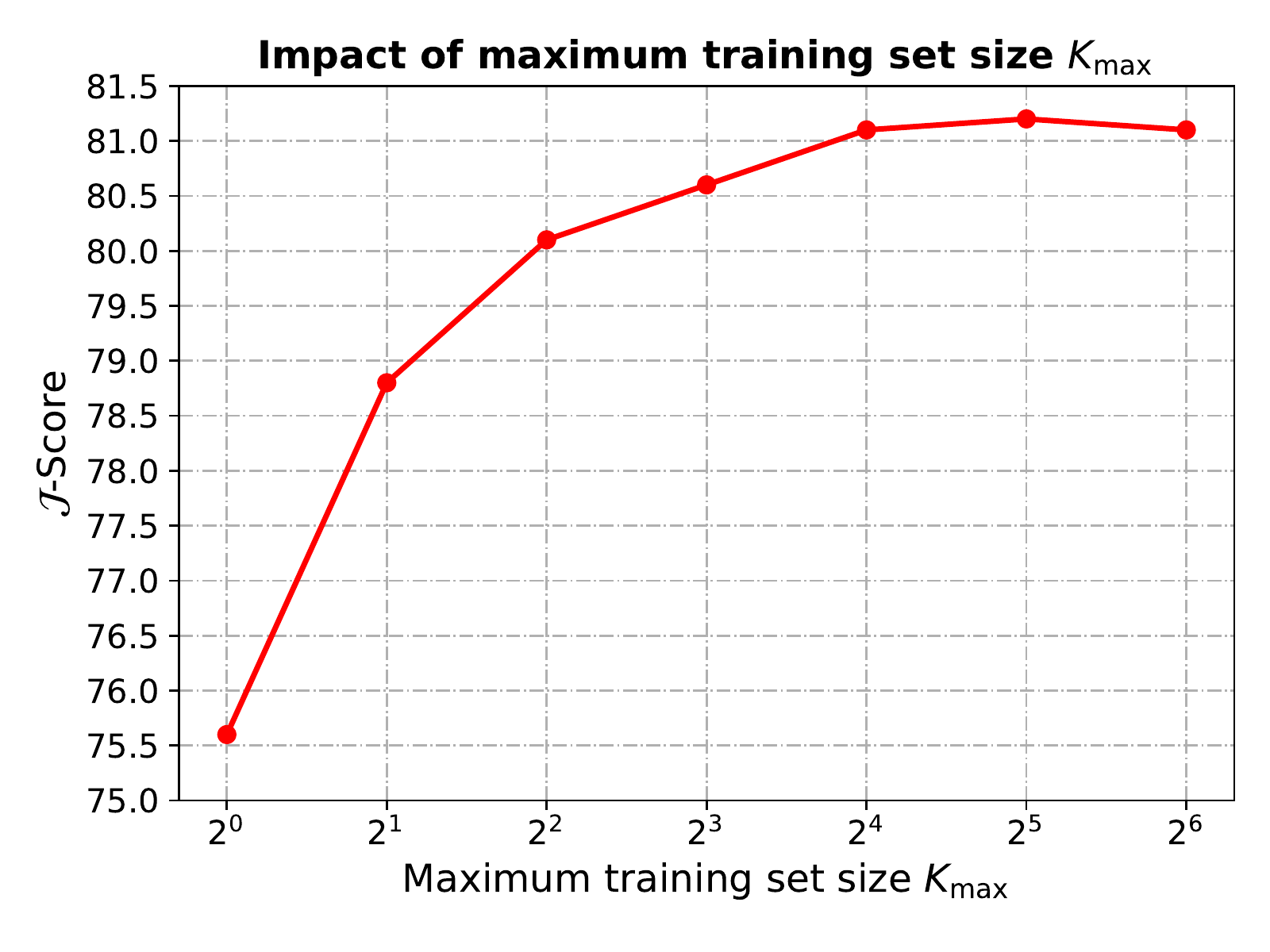}
    \vspace{-6mm}
    \caption{Impact of the maximum few-shot training set size $K_{\mathrm{max}}$ employed during inference. The results are shown in terms of the mean Jaccard $\mcJ$ score over a validation set of 300 videos randomly sampled from the YouTube-VOS 2019 training set.}
    \vspace{0mm}
    \label{fig:ablation_memsz}
\end{figure}

\parsection{Impact of maximum training set size $K_{\mathrm{max}}$} Here, we analyze the impact of the maximum few-shot training set size $K_{\mathrm{max}}$. Figure~\ref{fig:ablation_memsz} shows the $\mcJ$ score for different values of $K_{\mathrm{max}}$ over the 300 sequences in our validation set. The initial annotated sample is always included in our training set, in addition to $K_{\mathrm{max}} - 1$ previous frames. Thus, the setting $K_{\mathrm{max}}=1$ corresponds to using only the initial frame, while $K_{\mathrm{max}}=2$ corresponds to using the initial frame and the previous frame for updating the target model. Using a larger training set leads to improved performance until a training set size of $K_{\mathrm{max}}=16$, at which point the performance gain begins to saturate.

\begin{table}[t!]
\caption{Impact of different evaluation modes during inference. We compare a version in which the network operates on a local search region with a version in which the network operates on the full image. The search region in the first version is obtained by using the estimate of the target mask in the previous frame.}\vspace{0mm}
\centering
\resizebox{0.40\columnwidth}{!}{%
\renewcommand{\yes}{\text{\ding{51}}}
\renewcommand{\no}{-}
\centering
\begin{tabular}{l@{~~~~}c}
\toprule
 & $\mathcal{J} (\%)$ \\
\midrule
Local search region evaluation&81.2 \\
Full frame evaluation&80.2 \\
\bottomrule
\end{tabular}
}\vspace{0mm}
\label{tab:ablation_cropping}
\end{table}

\parsection{Impact of evaluation mode} We analyze the impact of different modes for running the network on the test frames during inference. We compare our approach in which the network operates on a local search region with a baseline version in which the network operates on the full image. The search region in our approach is obtained by using the estimate of the target mask in the previous frame, as described in Section~\ref{sec:sup_inference}. The results of this comparison is shown in Table~\ref{tab:ablation_cropping}. Operating on a local search region allows the network to better handle small objects, leading to an improvement of $+1.0$ points in $\mcJ$ score over the baseline operating on the full image.

\section{Qualitative Results}
\label{sec:sup_qual}
\begin{figure}
    \centering
\tabcolsep=0.01cm
\renewcommand{\arraystretch}{0.3}
\renewcommand{\image}{\includegraphics[width=0.24\columnwidth]}
\centering
\begin{tabular}{cccc}
 & {\footnotesize Ground truth} & {\footnotesize {\bf Ours}} & {\footnotesize STM}  \\ \midrule
\image{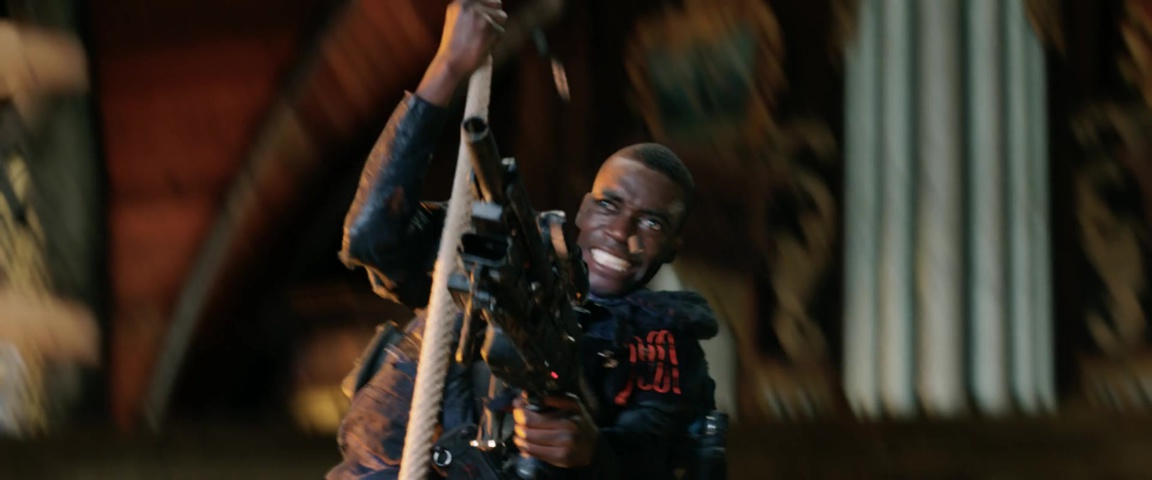} &
\image{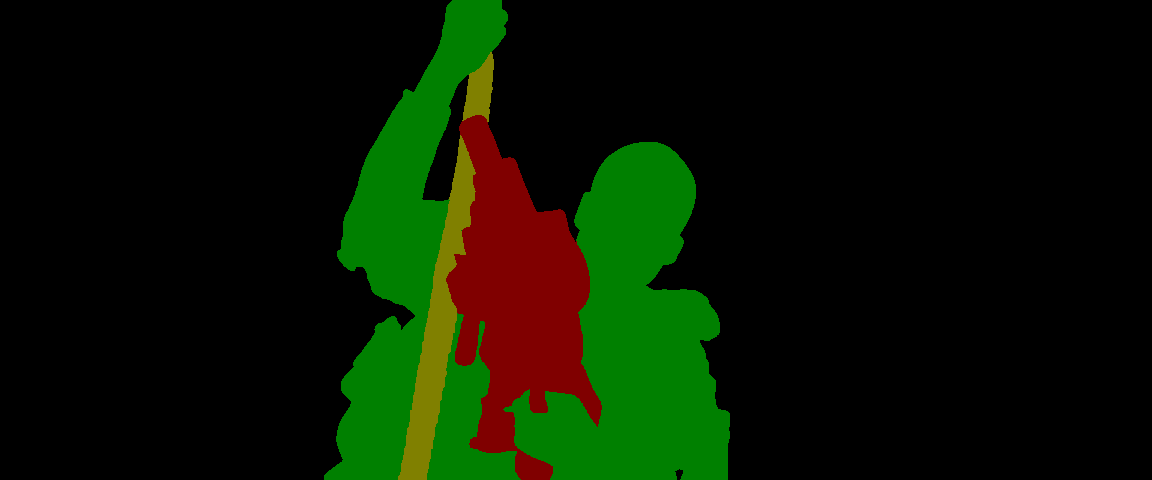} &
\image{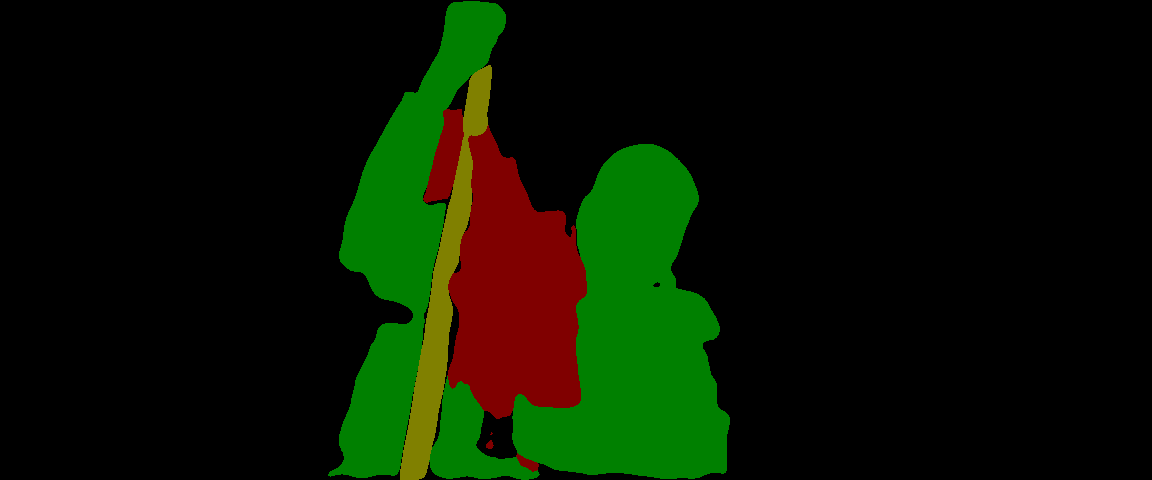} &
\image{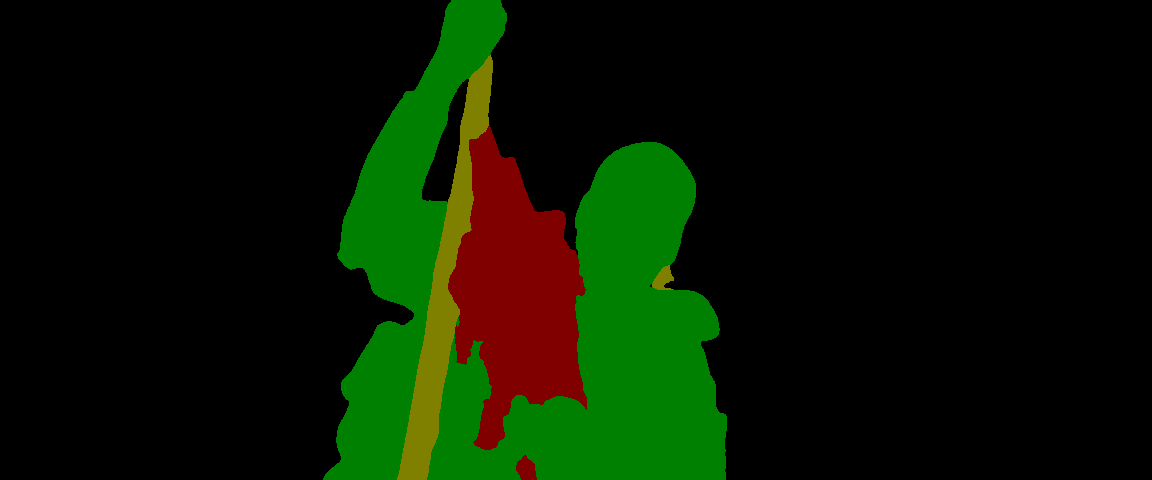} \\
\image{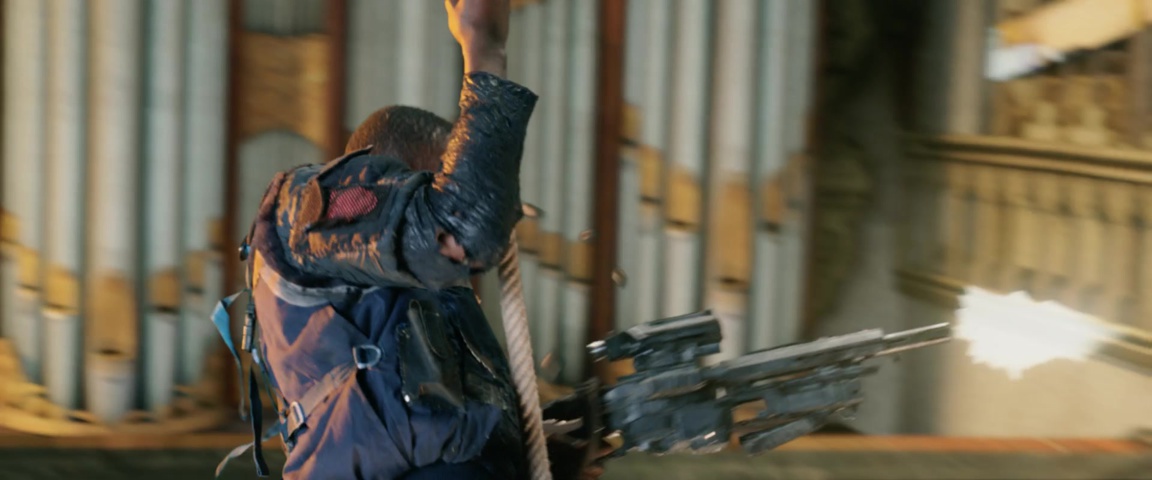} &
\image{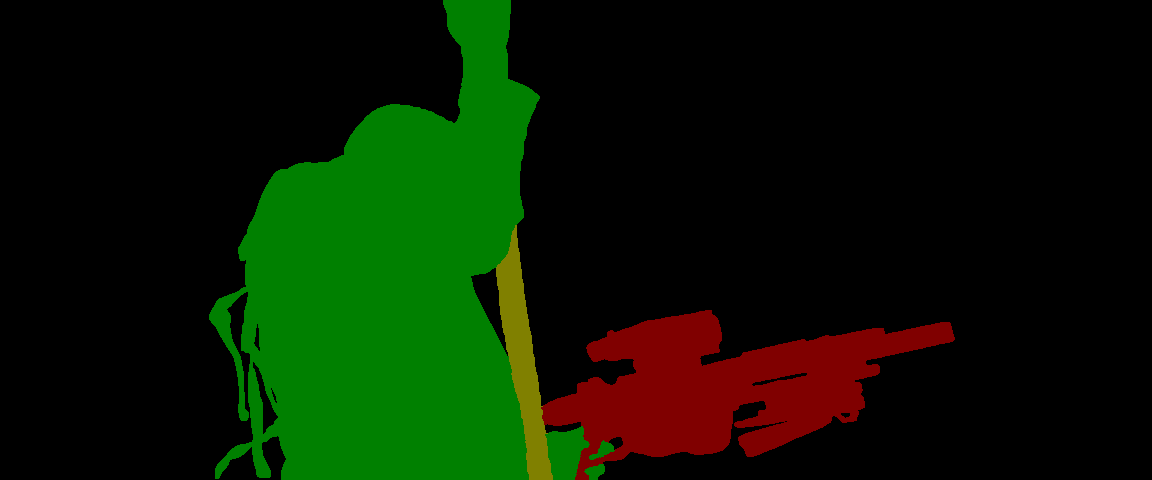} &
\image{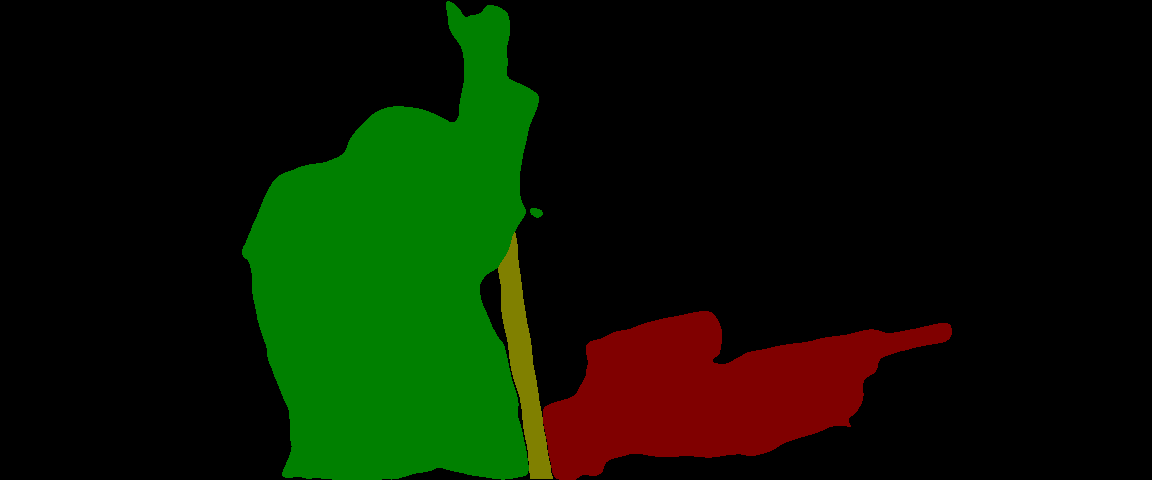} &
\image{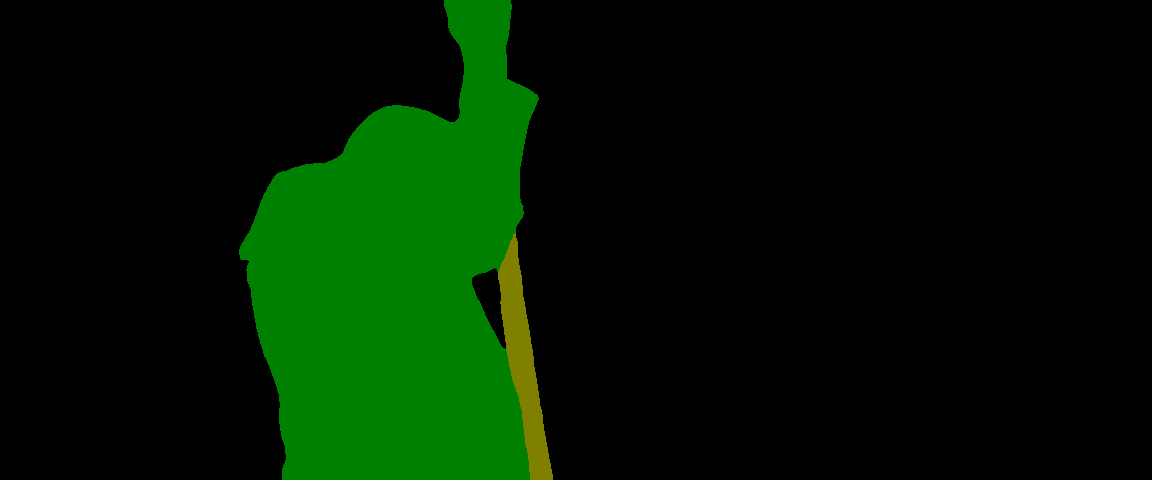}  \\
\midrule
\image{} &
\image{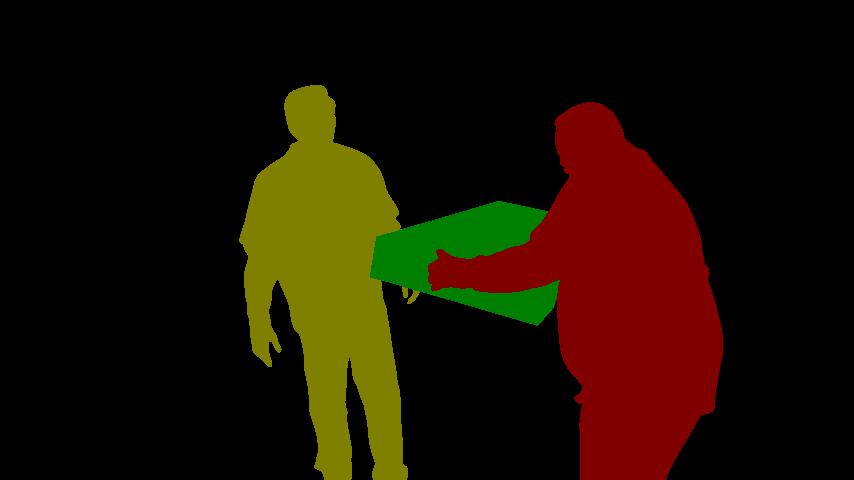} &
\image{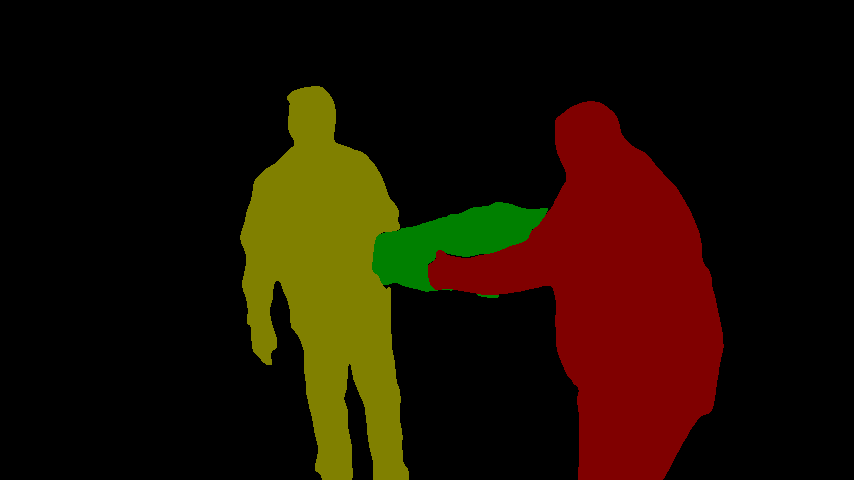} &
\image{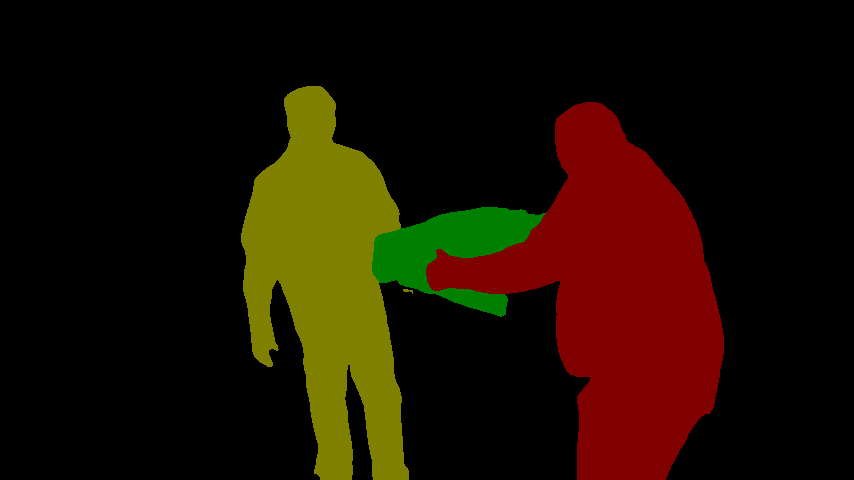} \\
\image{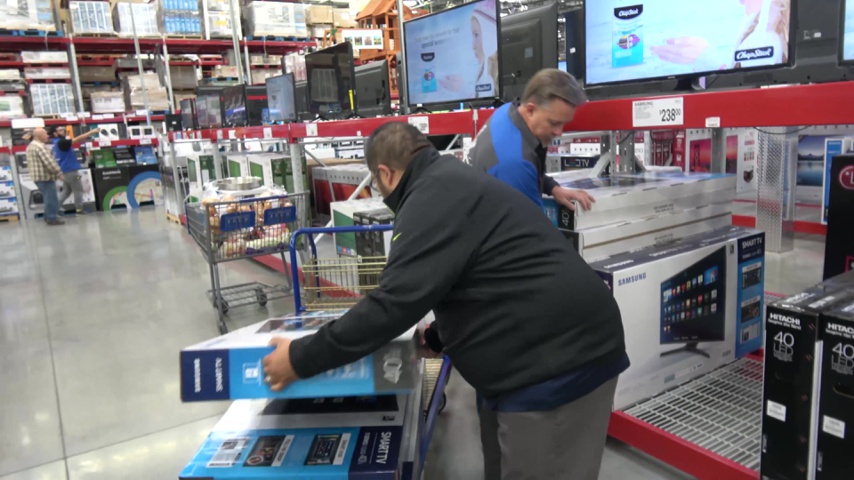} &
\image{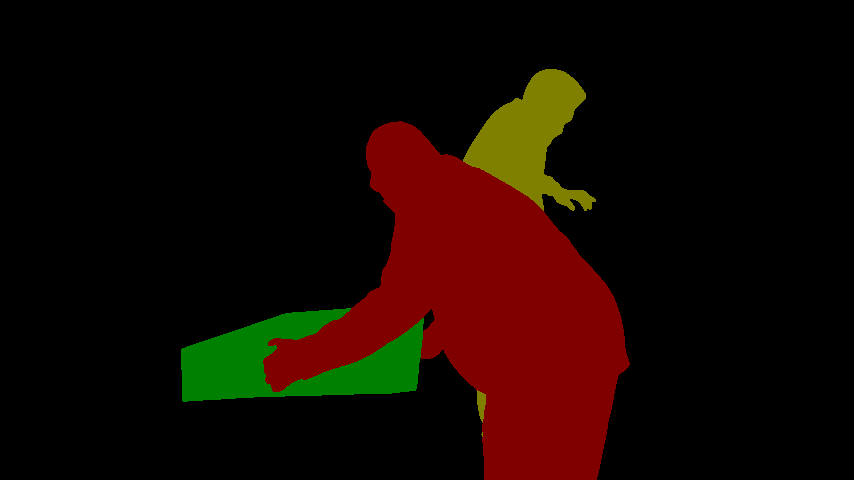} &
\image{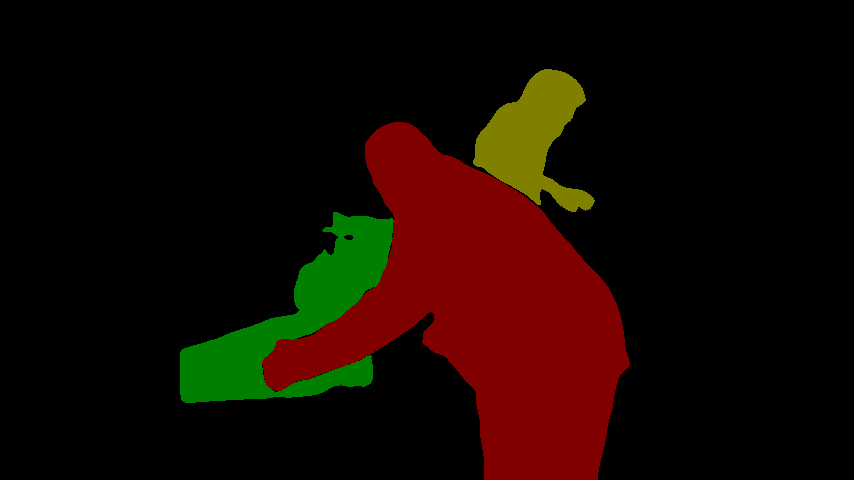} &
\image{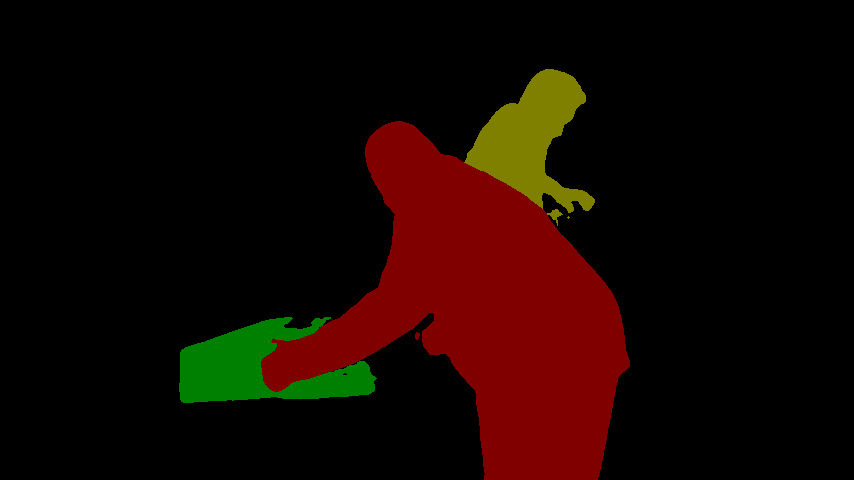}  \\
\midrule
\image{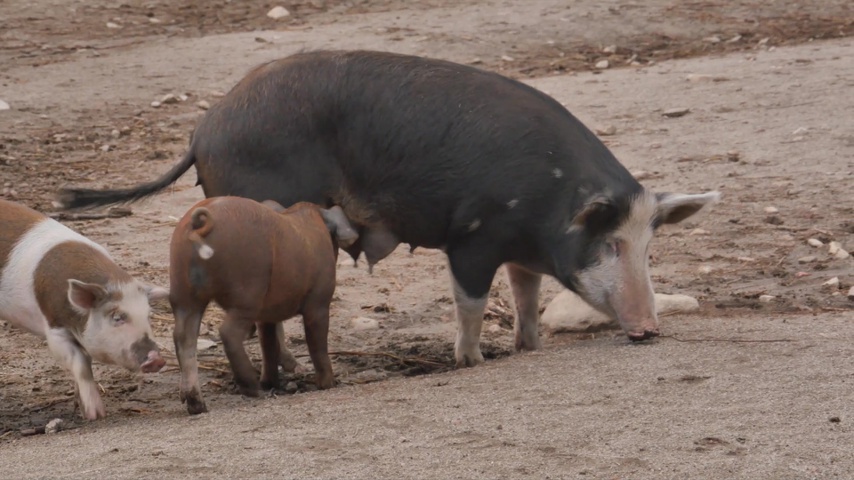} &
\image{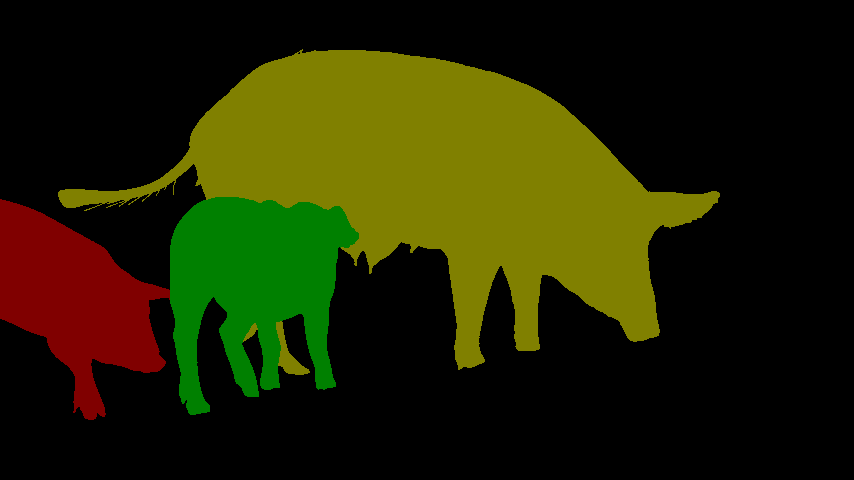} &
\image{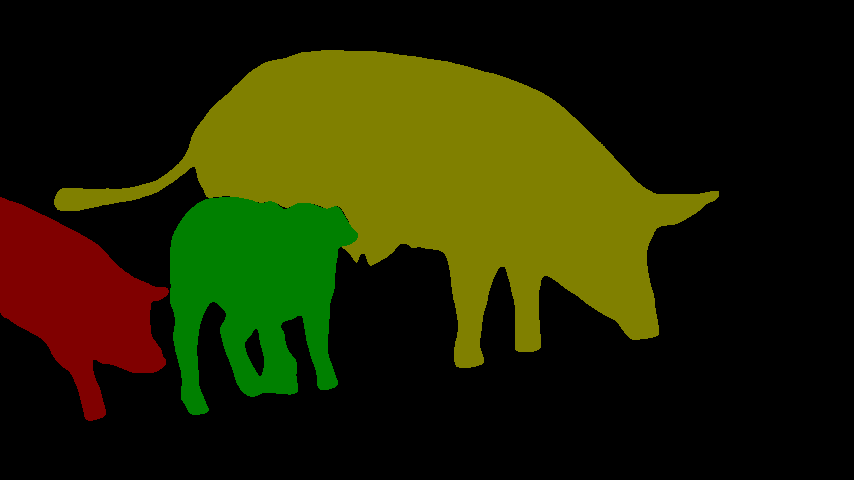} &
\image{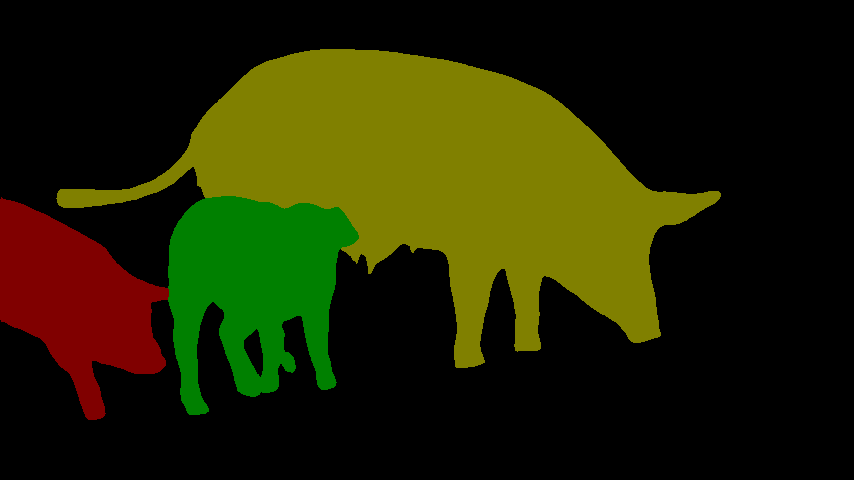} \\
\image{} &
\image{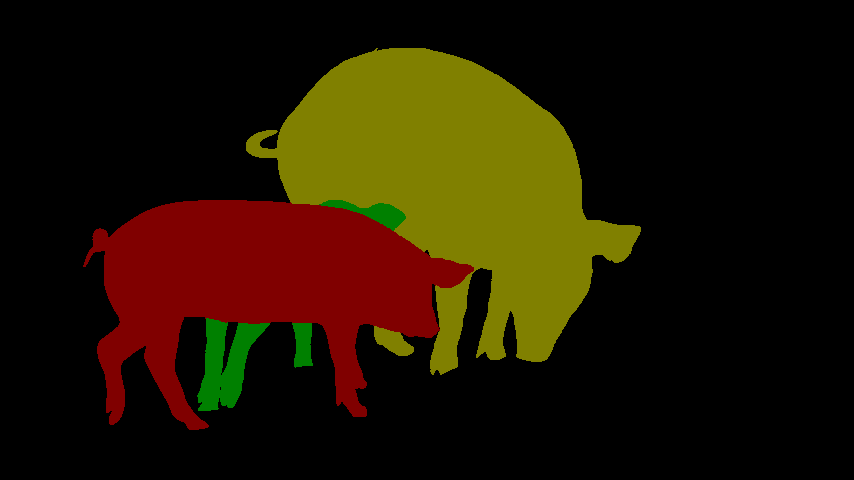} &
\image{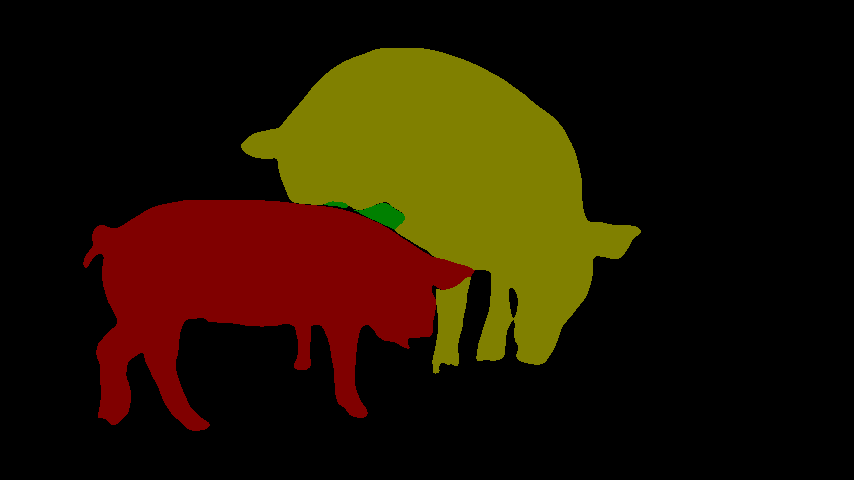} &
\image{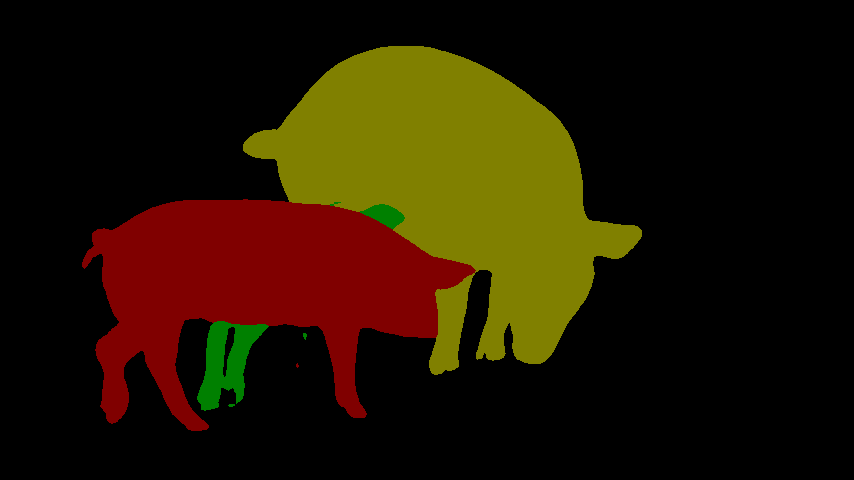}  \\
\midrule
\image{} &
\image{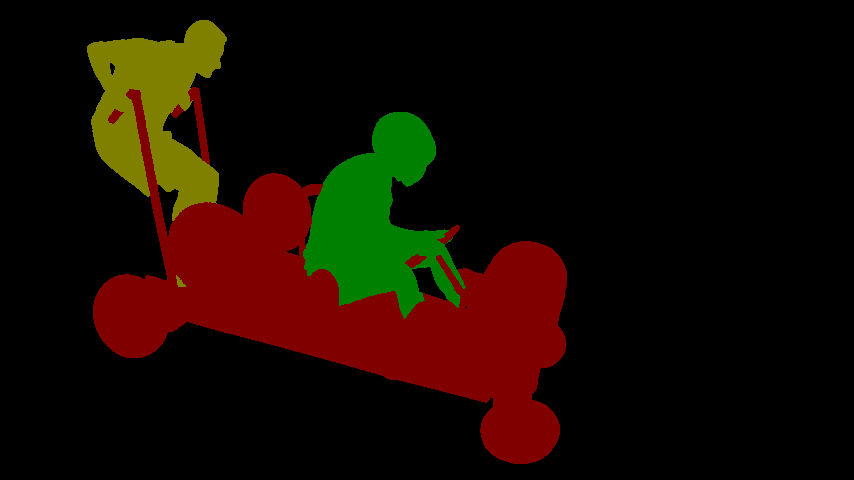} &
\image{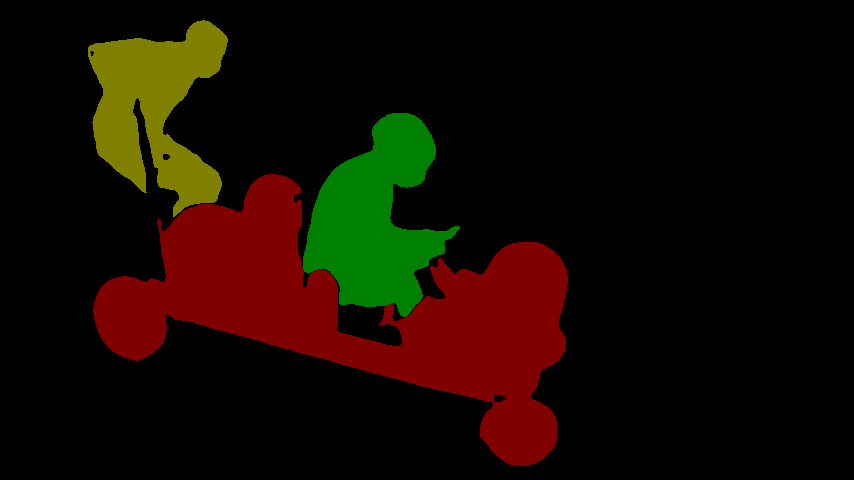} &
\image{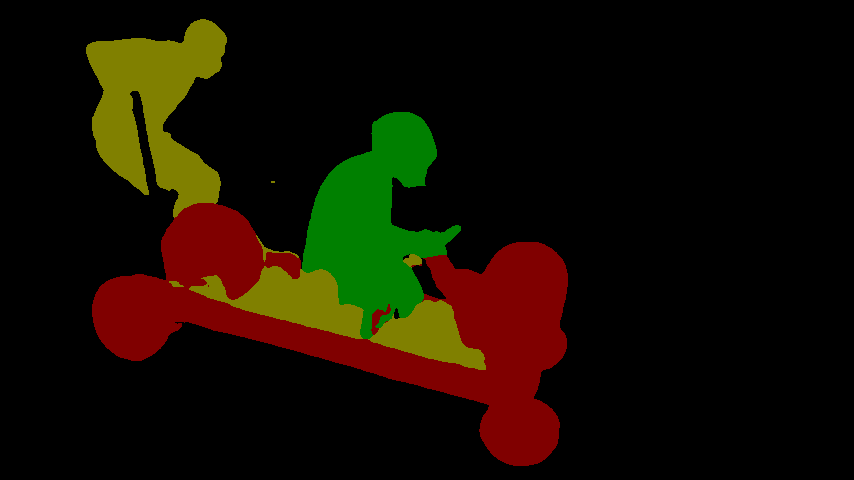} \\
\image{} &
\image{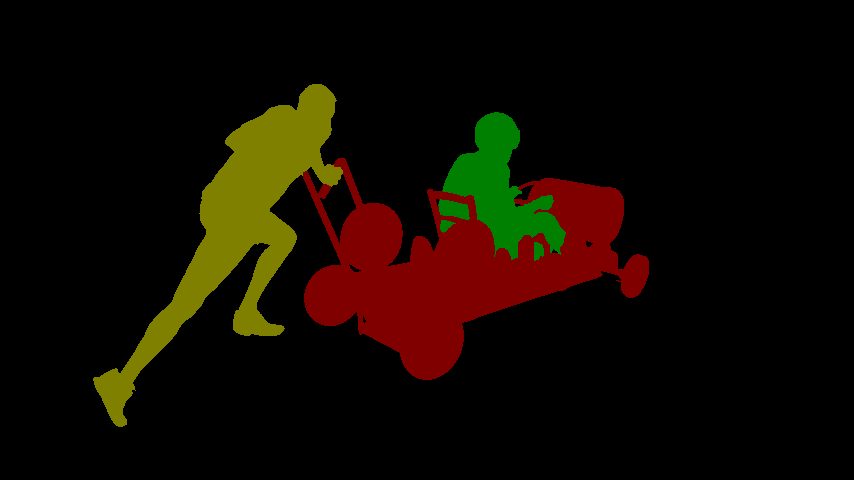} &
\image{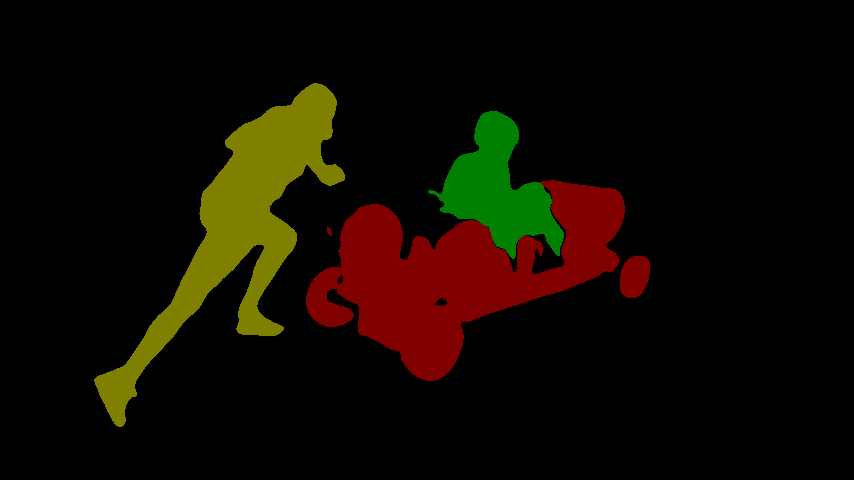} &
\image{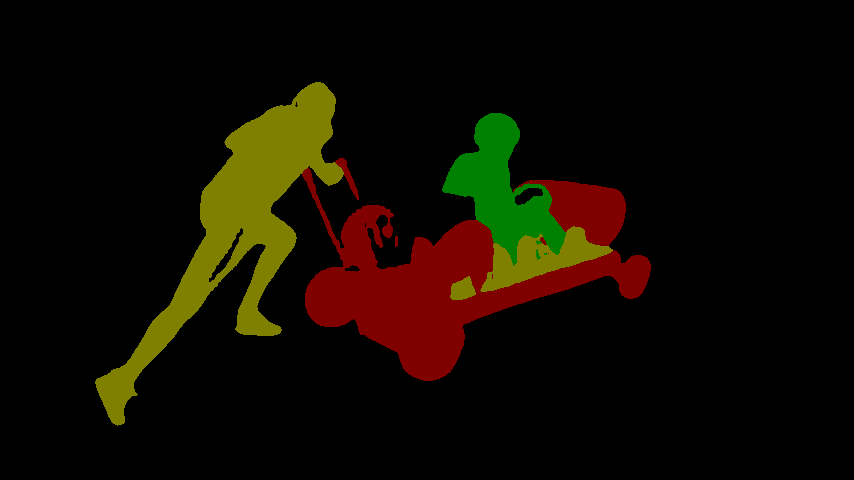}  \\
\end{tabular}
\caption{Qualitative comparisons between our approach and STM~\cite{oh2019video} on the sequences shooter, loading, pigs and soapbox from the DAVIS 2017 validation set. For each sequence, we have selected two specific frames with distinct scene appearance.}
\label{fig:sup-q-comparison}
\end{figure}

\parsection{Qualitative Comparison} We show a qualitative comparison between our approach and STM~\cite{oh2019video} in Figure~\ref{fig:sup-q-comparison}. In the comparison we include two frames from each of the sequences shooter, loading, pigs and soapbox from the DAVIS 2017 validation set. These sequences contain several challenges such as occlusions, changes in appearance and distractor objects. We observe that these challenges are handled very differently by STM and our approach. In the shooter sequence, STM fails to segment the gun in the late frame, while our approach successfully segments all targets. Further, our approach struggles with segmenting the box in the loading sequence and one of the piglets in the pigs sequence. Finally, in contrast to our approach, STM manages to segment the thin structured handles of the wagon in the soapbox sequence. On the other hand, STM falsely predicts segments on the wagon with the label of the running man, while our approach segments all targets without any major false predictions.

\begin{figure}
    \centering
\tabcolsep=0.01cm
\renewcommand{\arraystretch}{0.3}
\renewcommand{\image}{\includegraphics[width=0.12\columnwidth]}
\centering
\begin{tabular}{cccccccc}
\image{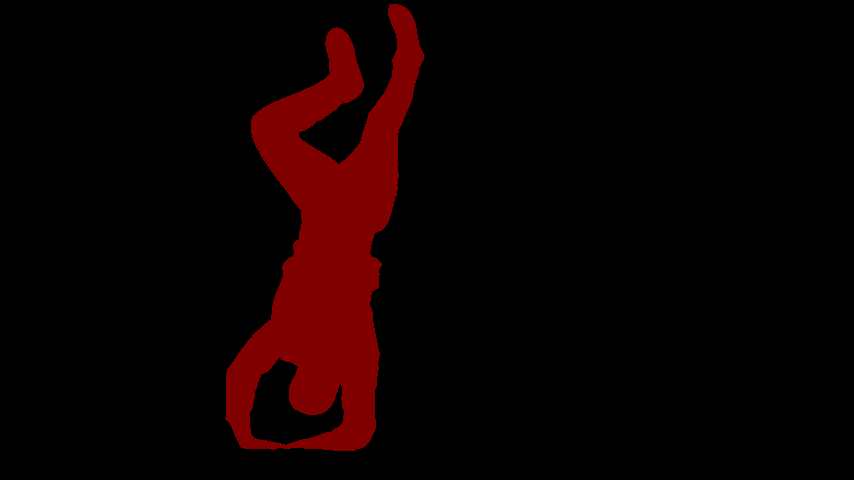} &  \image{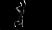} &
\image{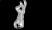} &
\image{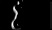} &
\image{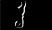}&
\image{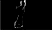} &
\image{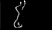} &
\image{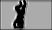} \\
\image{} & \image{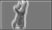} &
\image{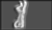} &
\image{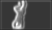} &
\image{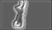} &
\image{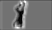} &
\image{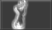} &
\image{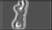} \\ \midrule
\image{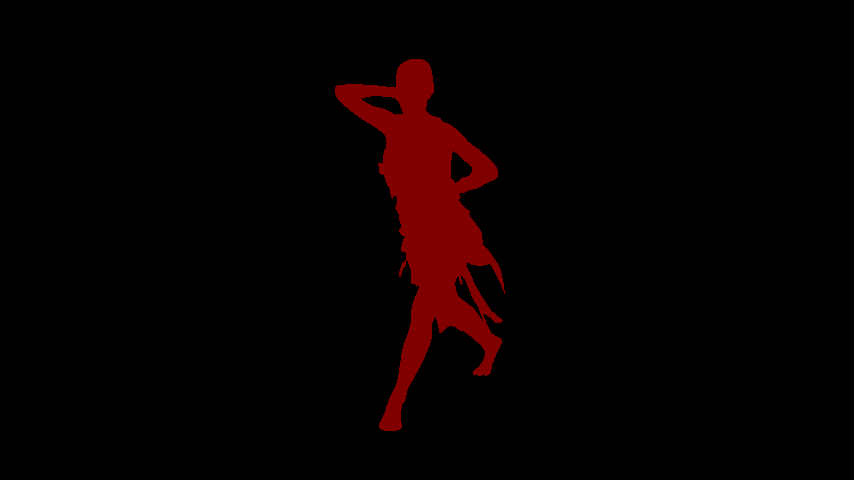} & \image{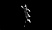} &
\image{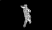} &
\image{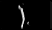} &
\image{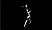} &
\image{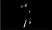} &
\image{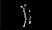} &
\image{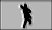} \\
\image{} & \image{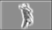} &
\image{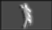} &
\image{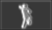} &
\image{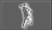} &
\image{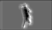} &
\image{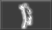} &
\image{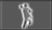} \\ \midrule
\image{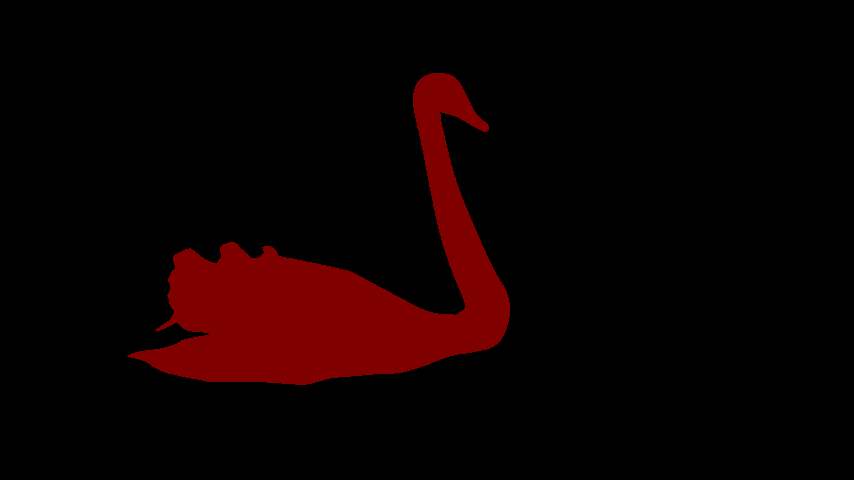} & \image{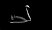} &
\image{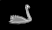} &
\image{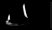} &
\image{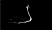} &
\image{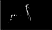} &
\image{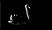} &
\image{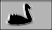} \\
\image{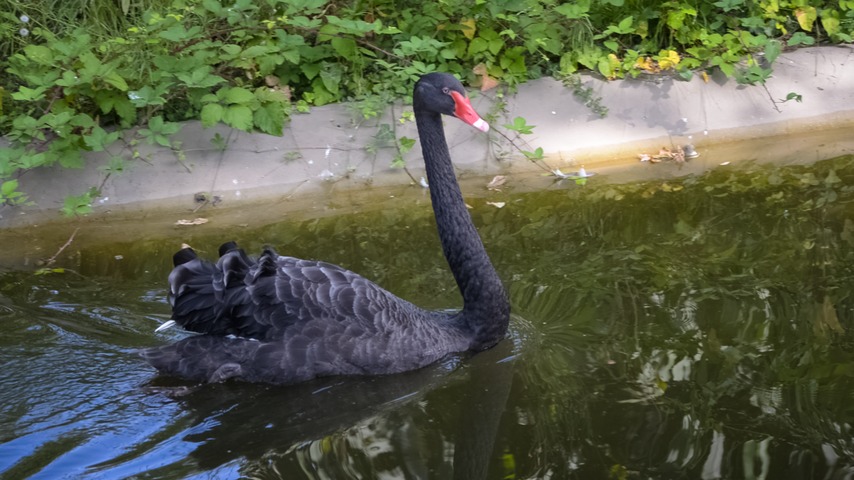} & \image{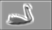} &
\image{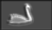} &
\image{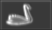} &
\image{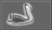} &
\image{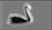} &
\image{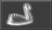} &
\image{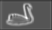} \\
\end{tabular}
\caption{Visualization of learned mask encodings. The left most column contain the ground truth masks with the corresponding image below. In the other columns we show labels generated by the label generator in the top rows and absolute values of the corresponding importance weights in the rows below.}
\label{fig:sup-encodings}
\end{figure}

\parsection{Label encoding} In Figure~\ref{fig:sup-encodings} we show a number of selected channels from the labels produced by the label generator and the absolute values of the corresponding importance weights. As we can see, the channels in labels model different aspects of the target mask, such as boundaries, background and a low resolution approximation of the mask. The weights alternate between being higher or lower on the background compared to the foreground regions. However, there seems to be consistently higher weighting around the edges of the target.

\begin{figure}
    \centering
\tabcolsep=0.01cm
\renewcommand{\arraystretch}{0.06}
\renewcommand{\image}{\includegraphics[width=0.16\columnwidth]}
\centering
\begin{tabular}{cccccc}
\image{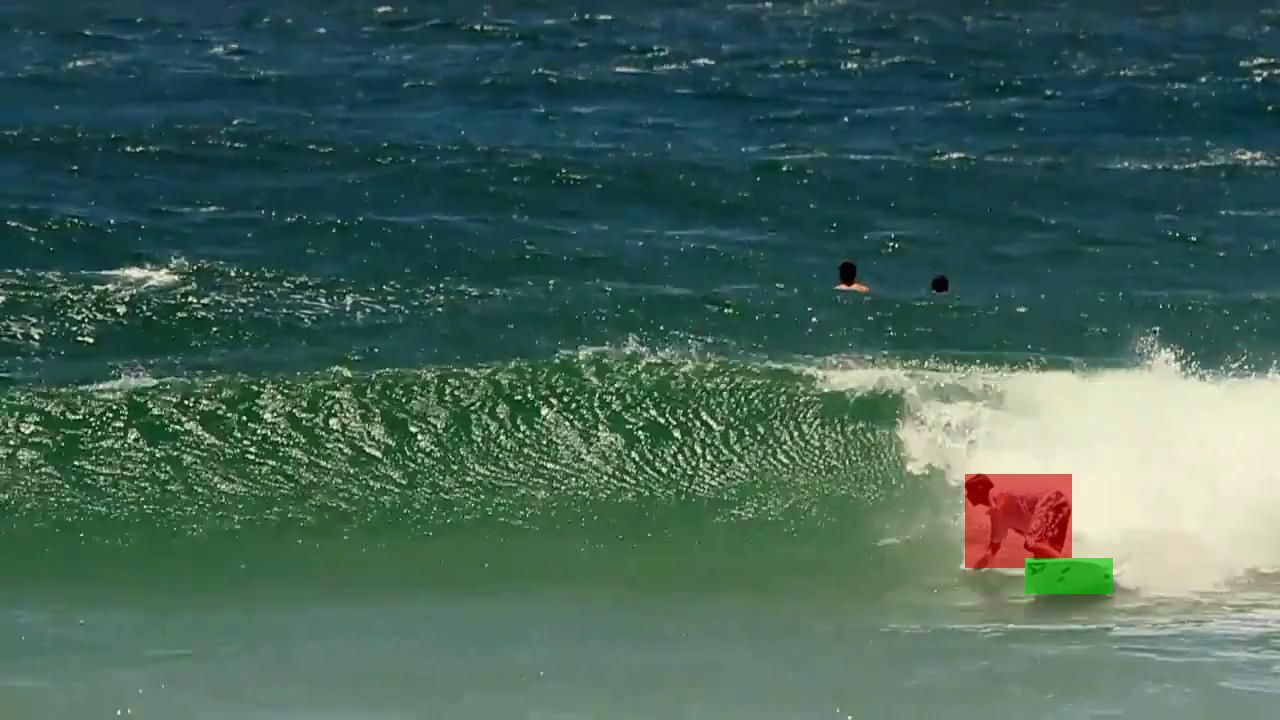} &
\image{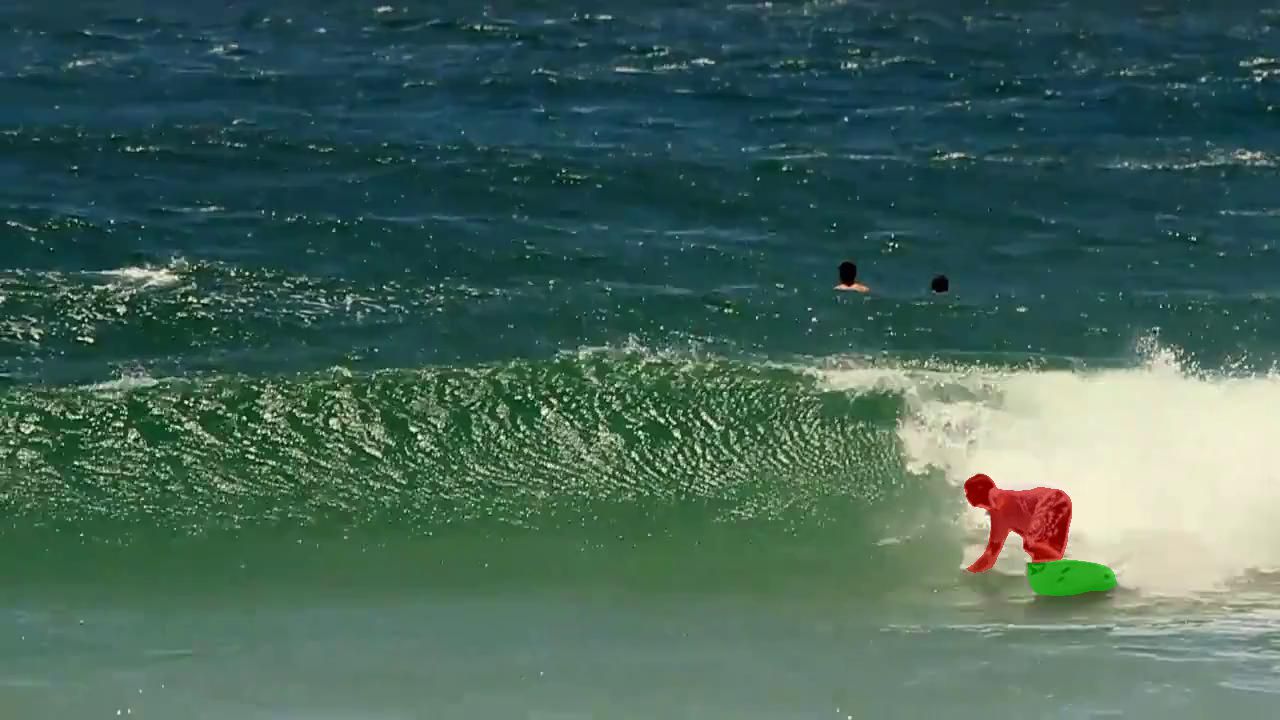} &
\image{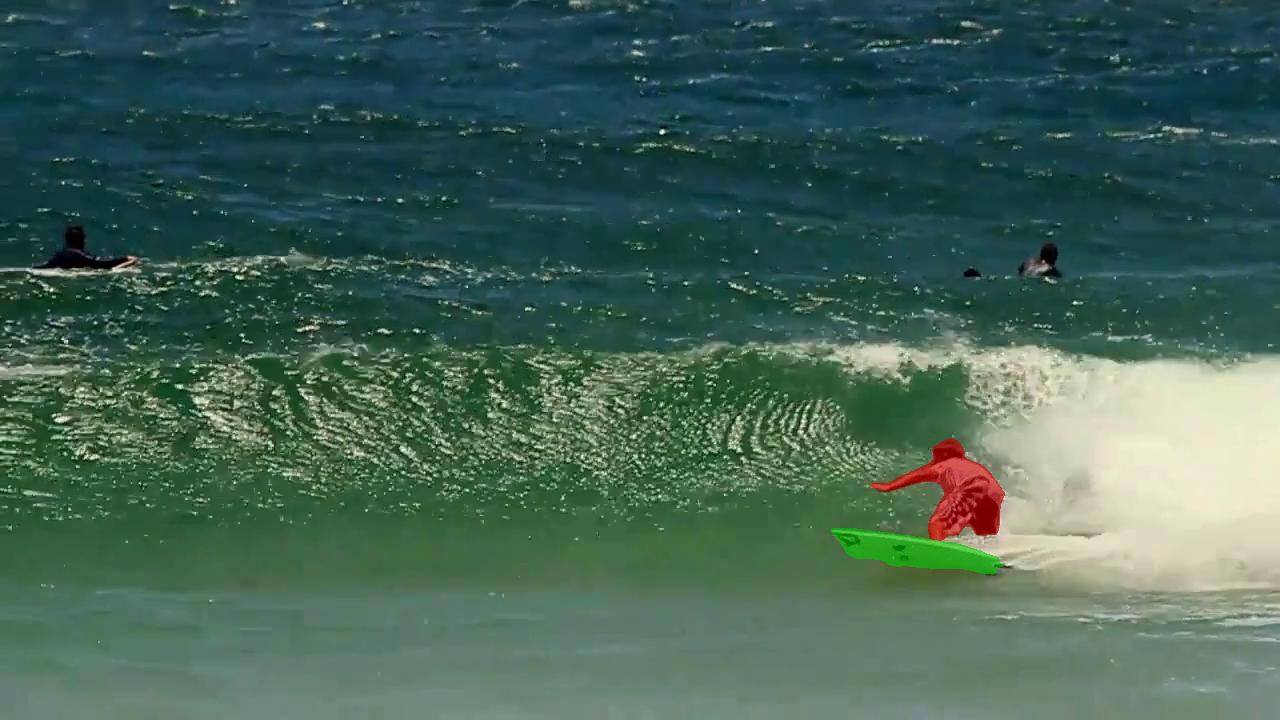} &
\image{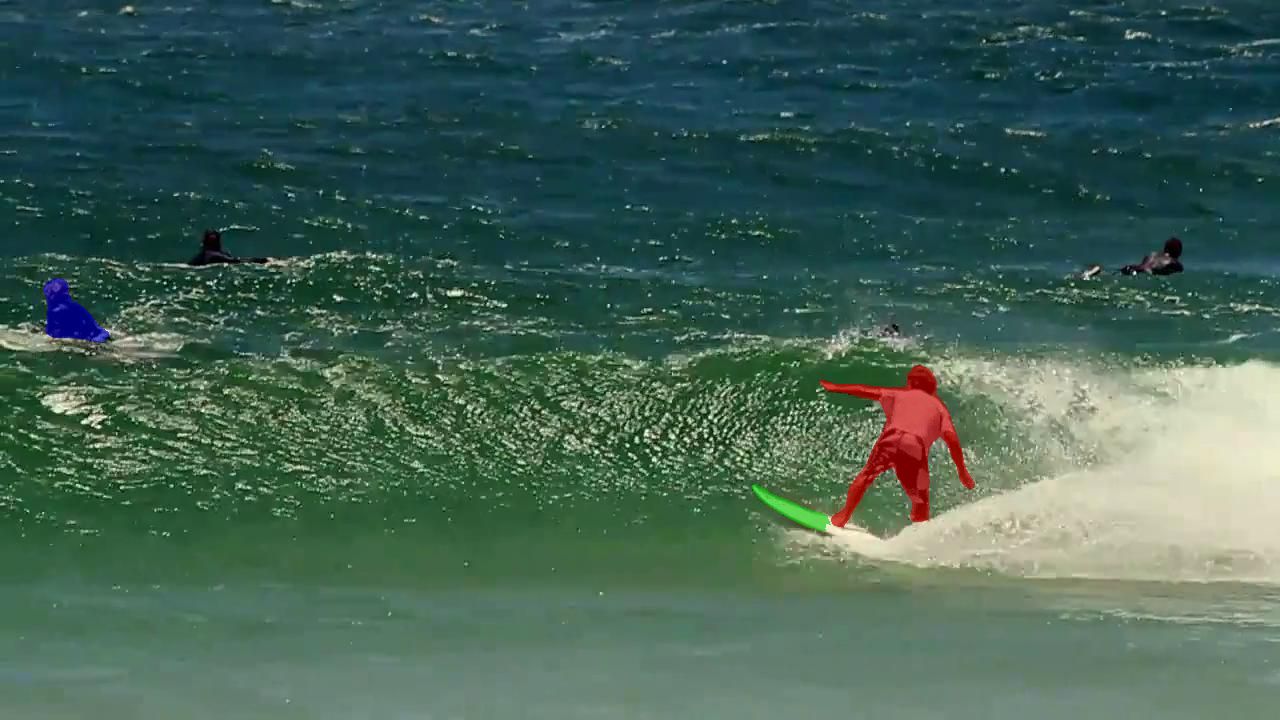} &
\image{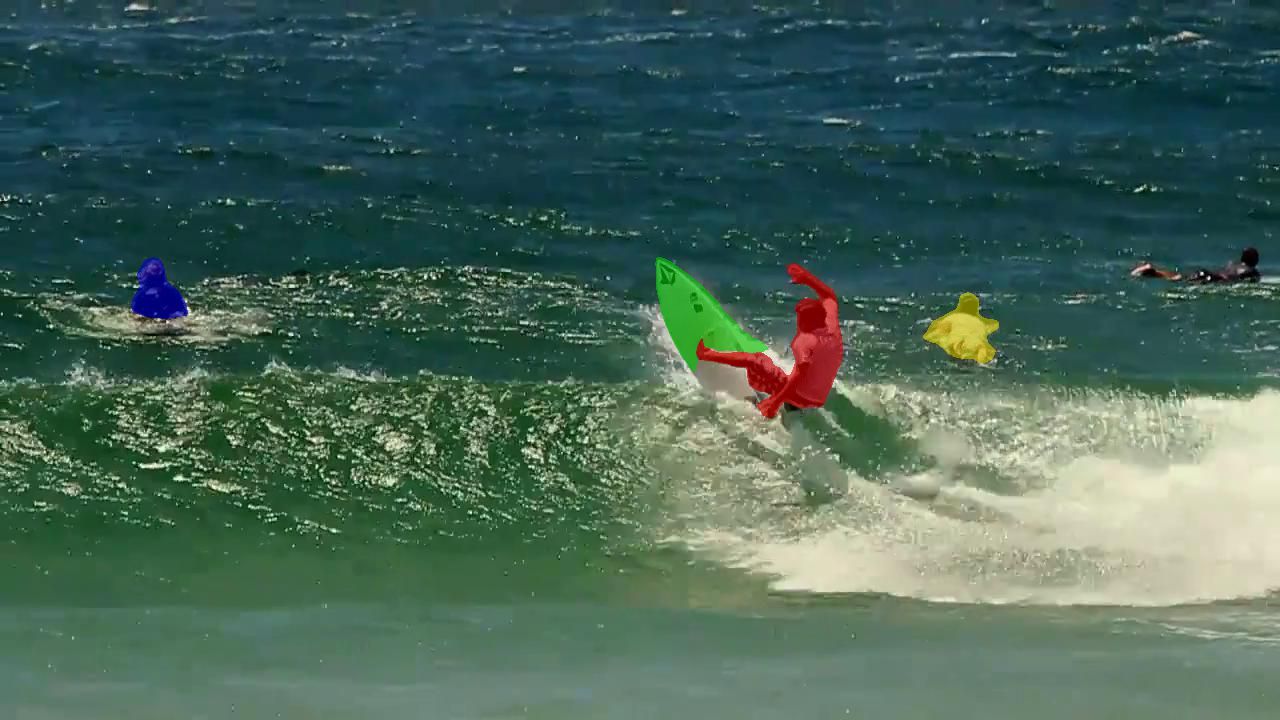} &
\image{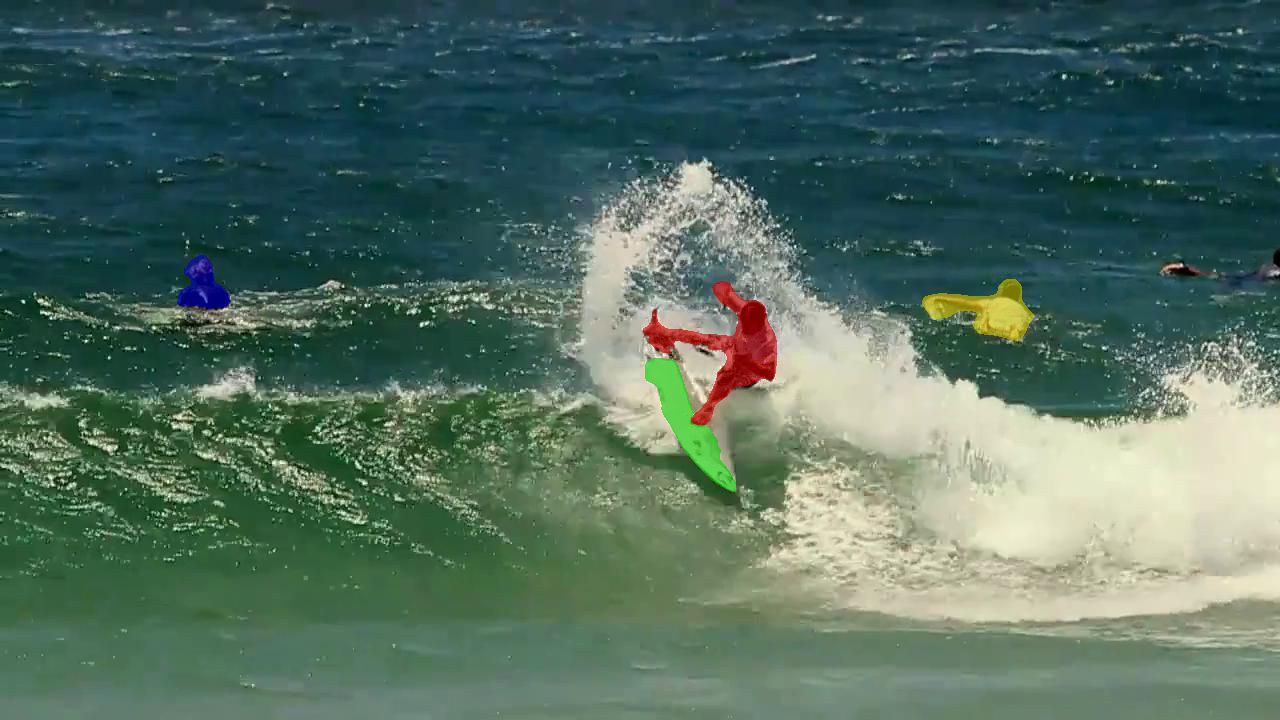}  \\
\image{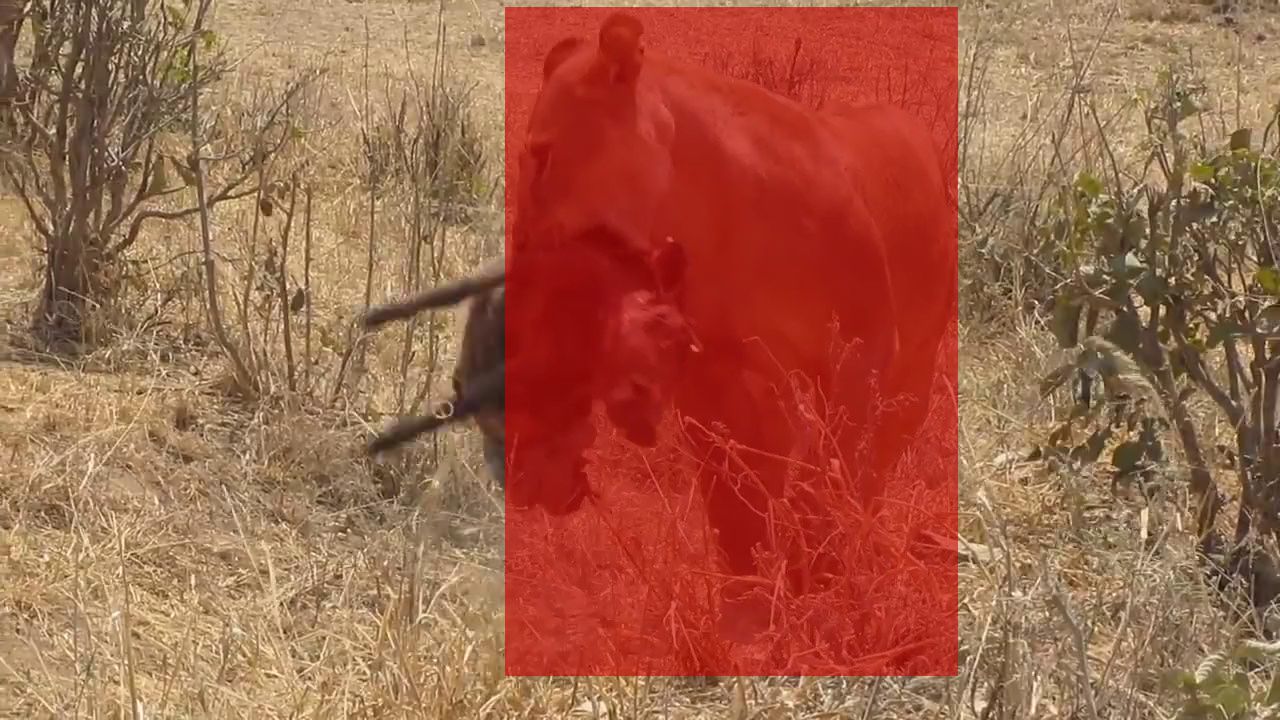} &
\image{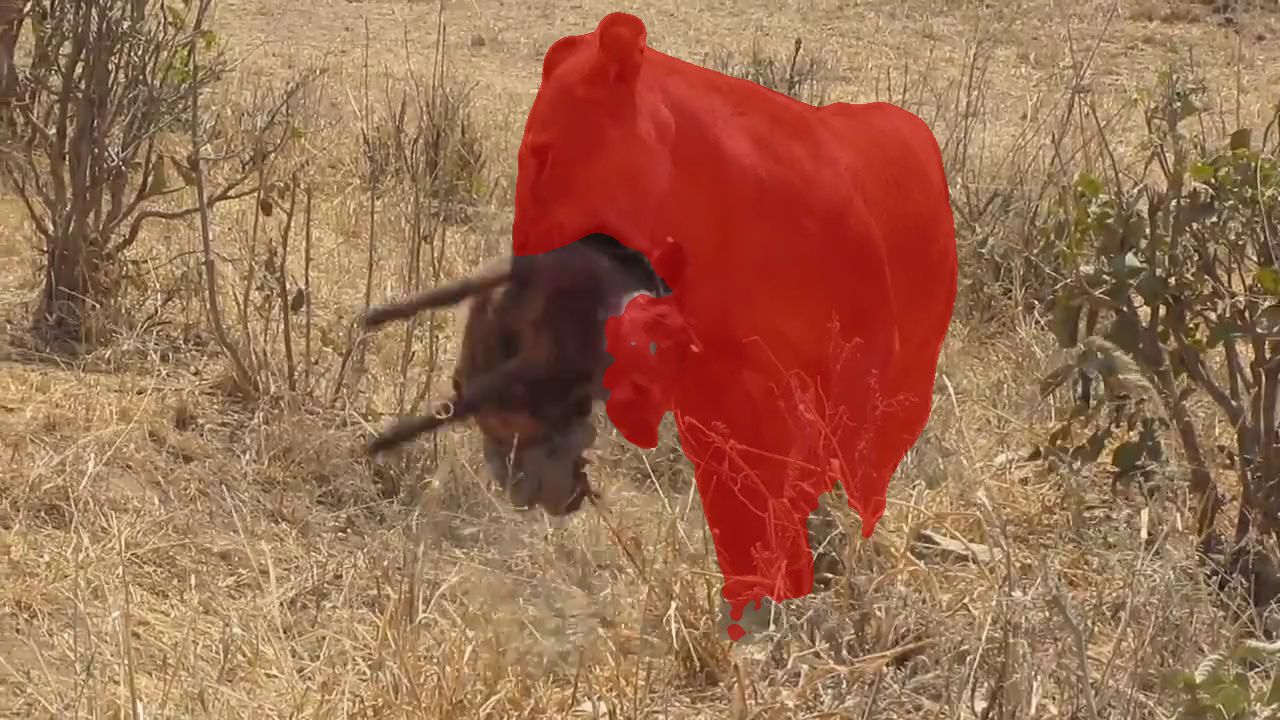} &
\image{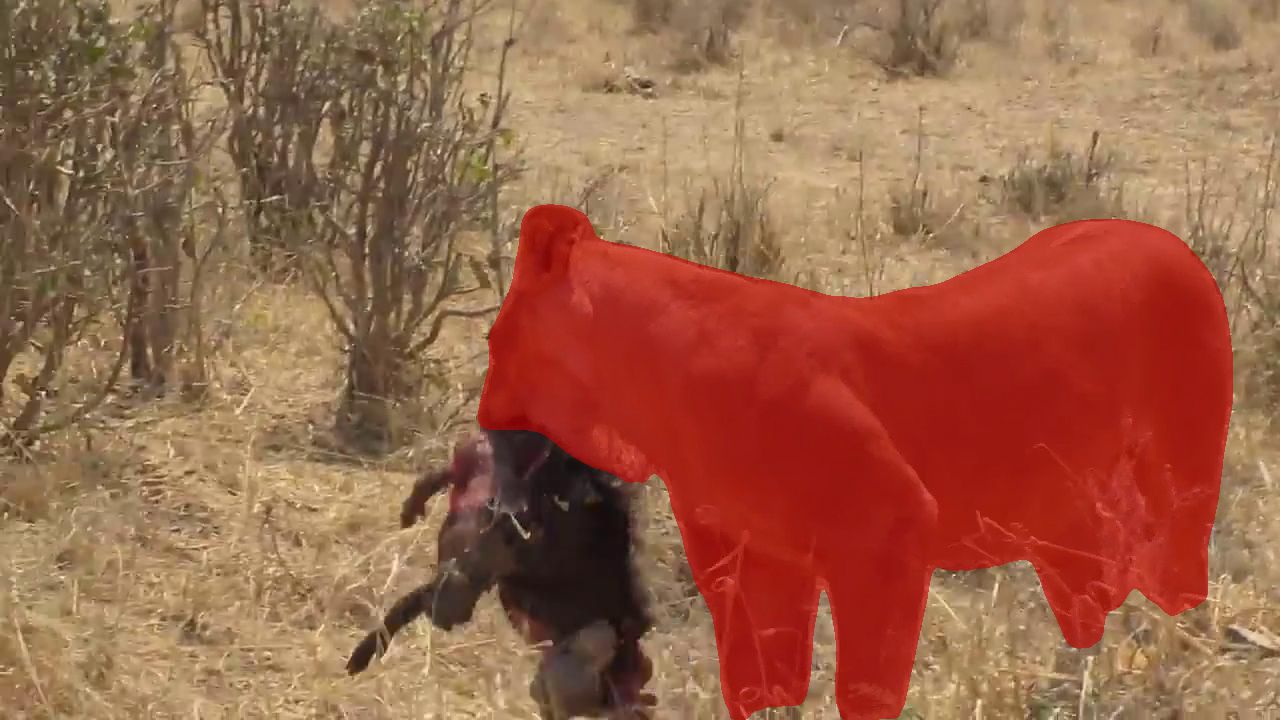} &
\image{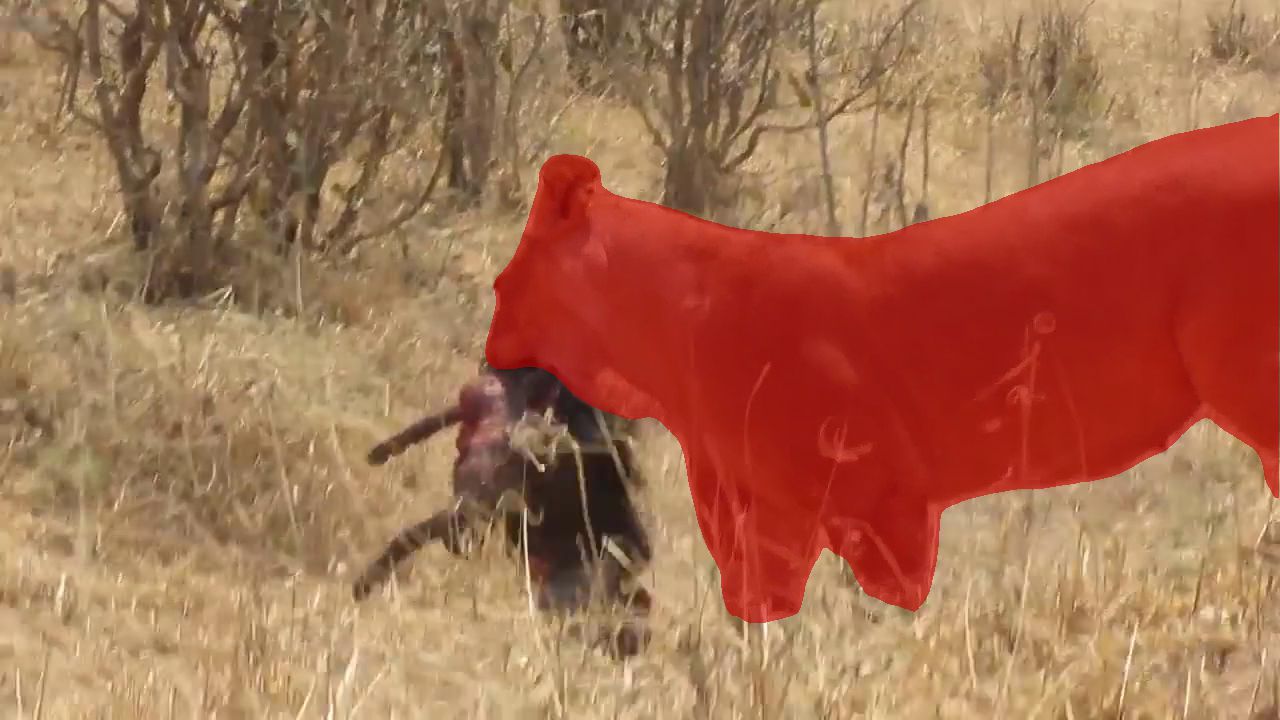} &
\image{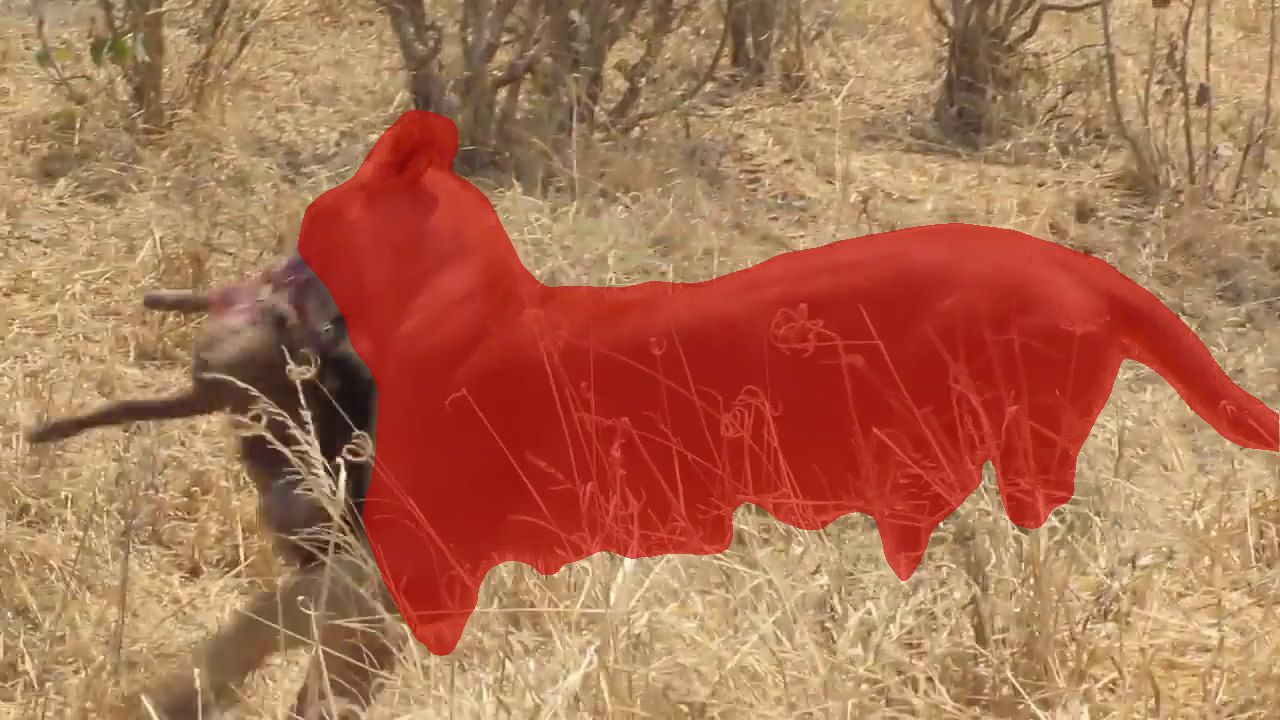} &
\image{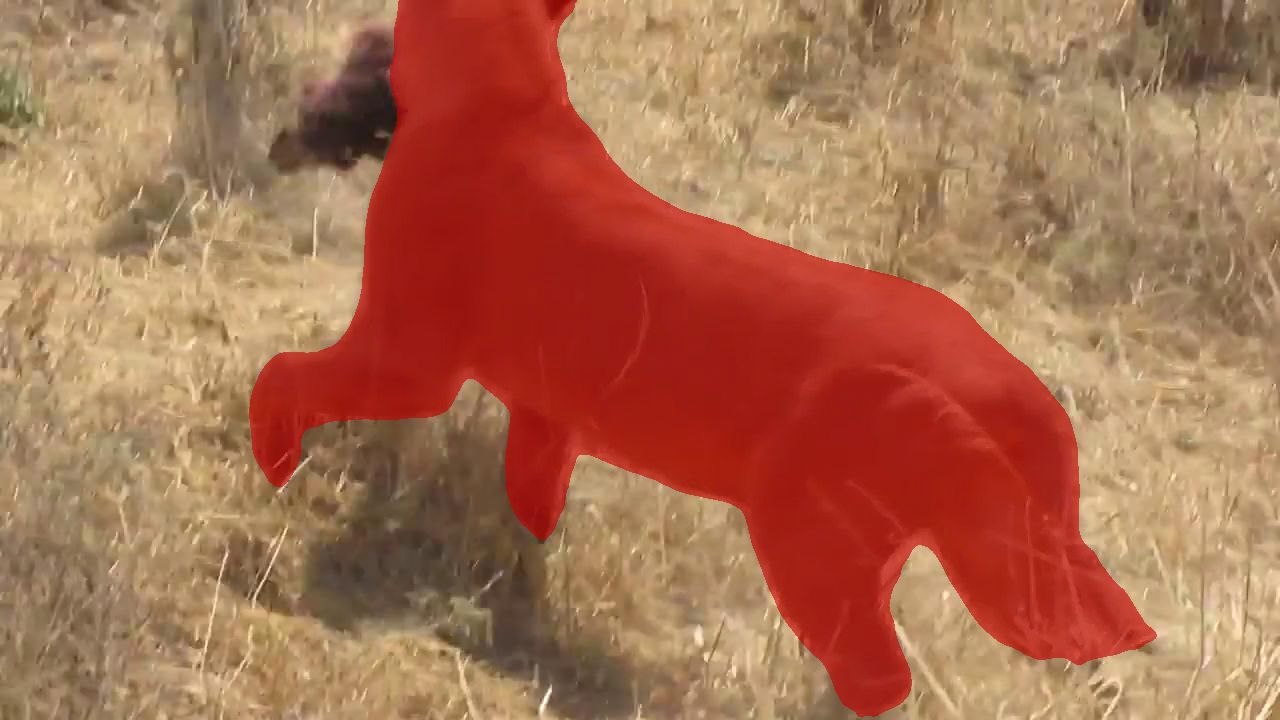} \\
\image{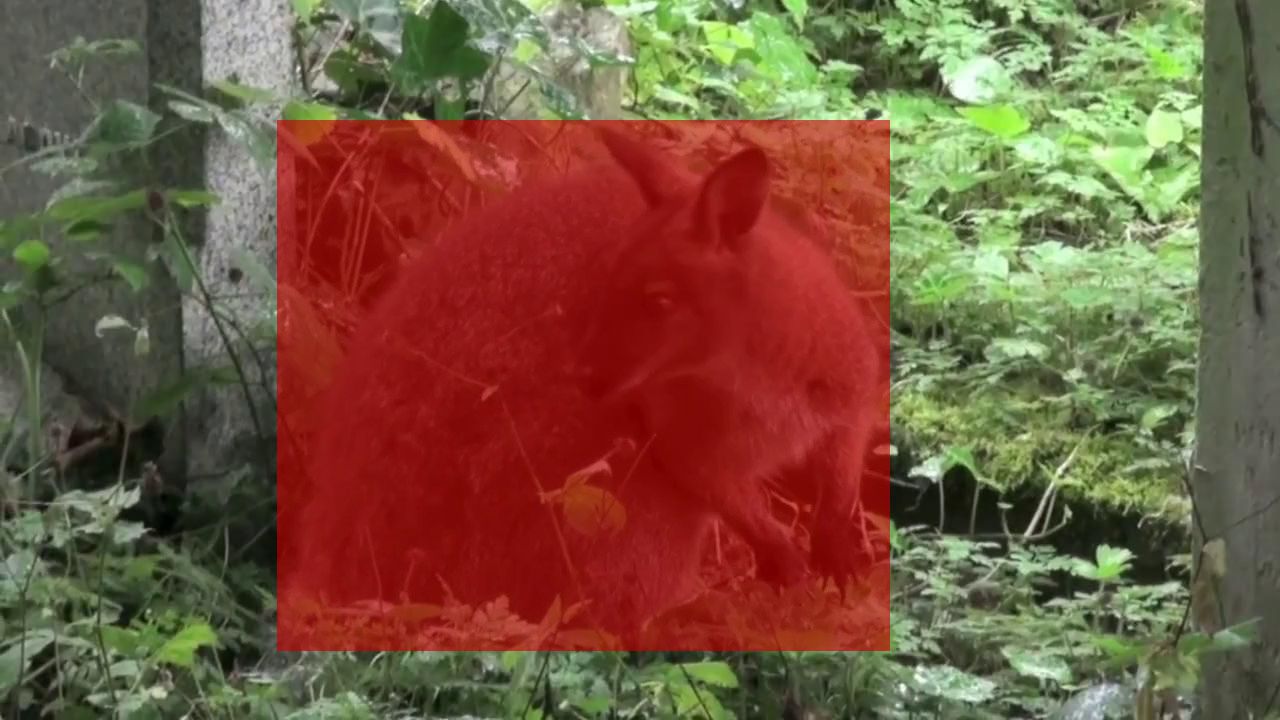} &
\image{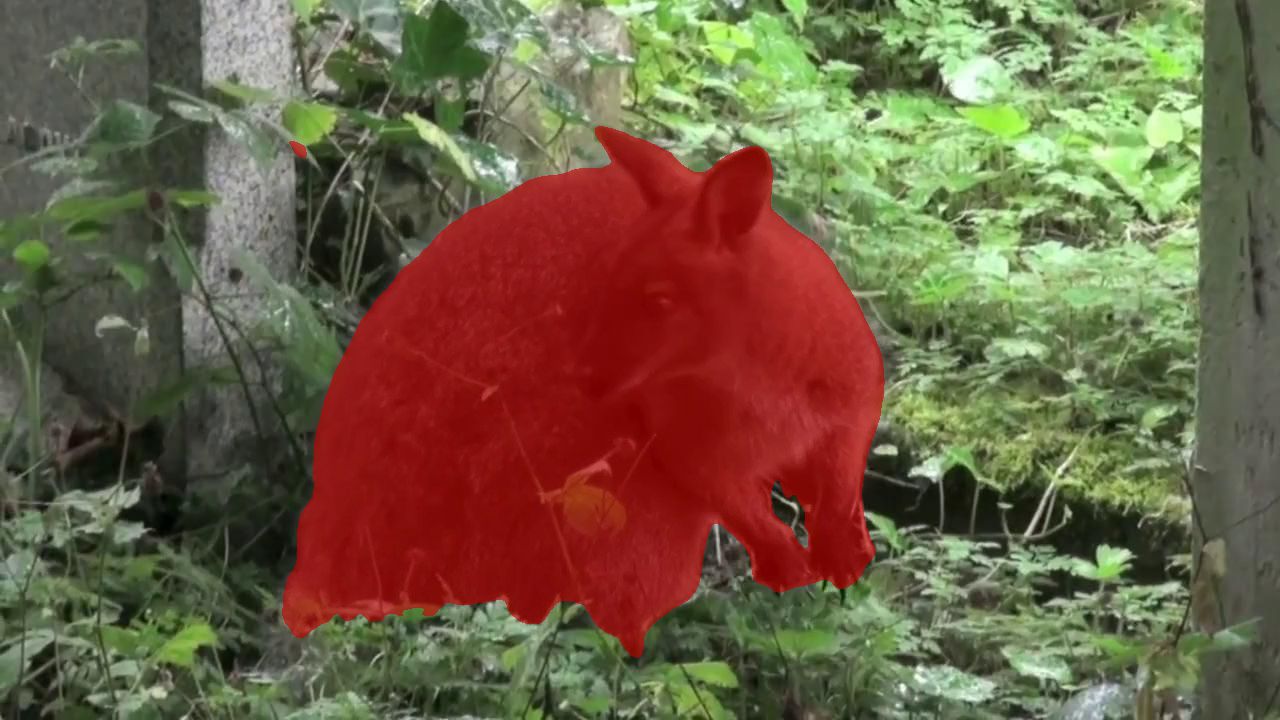} &
\image{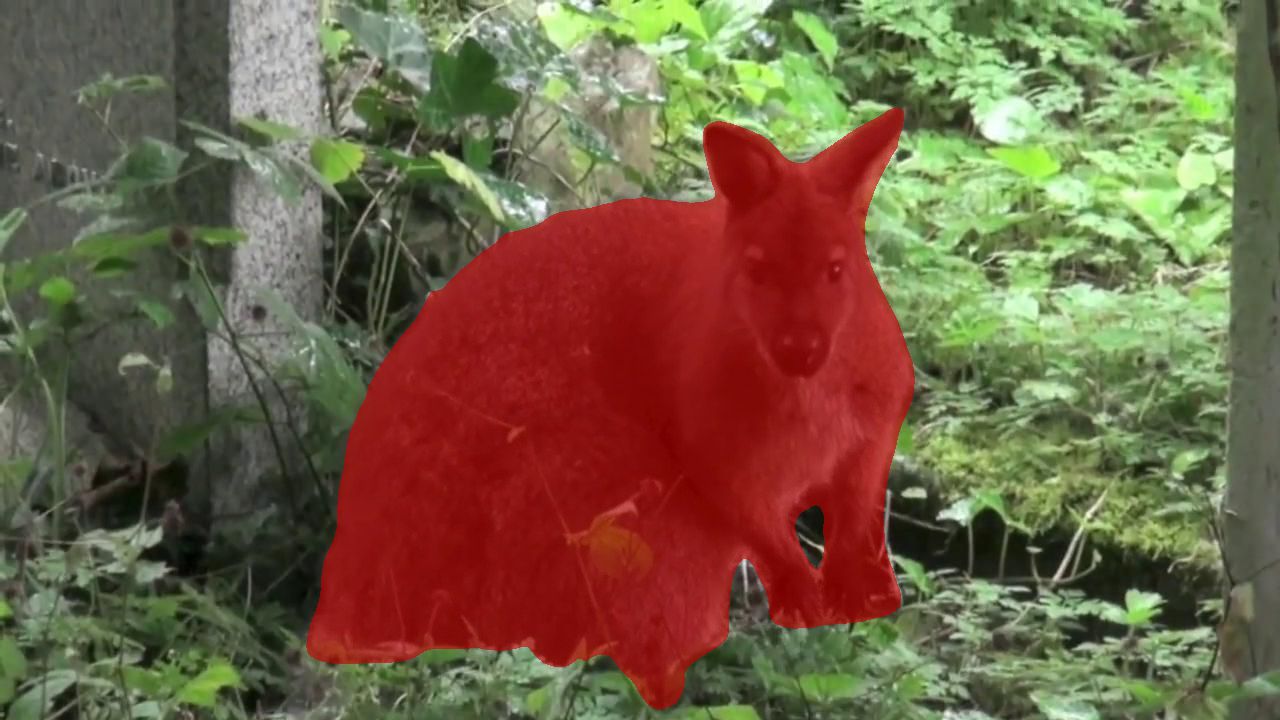} &
\image{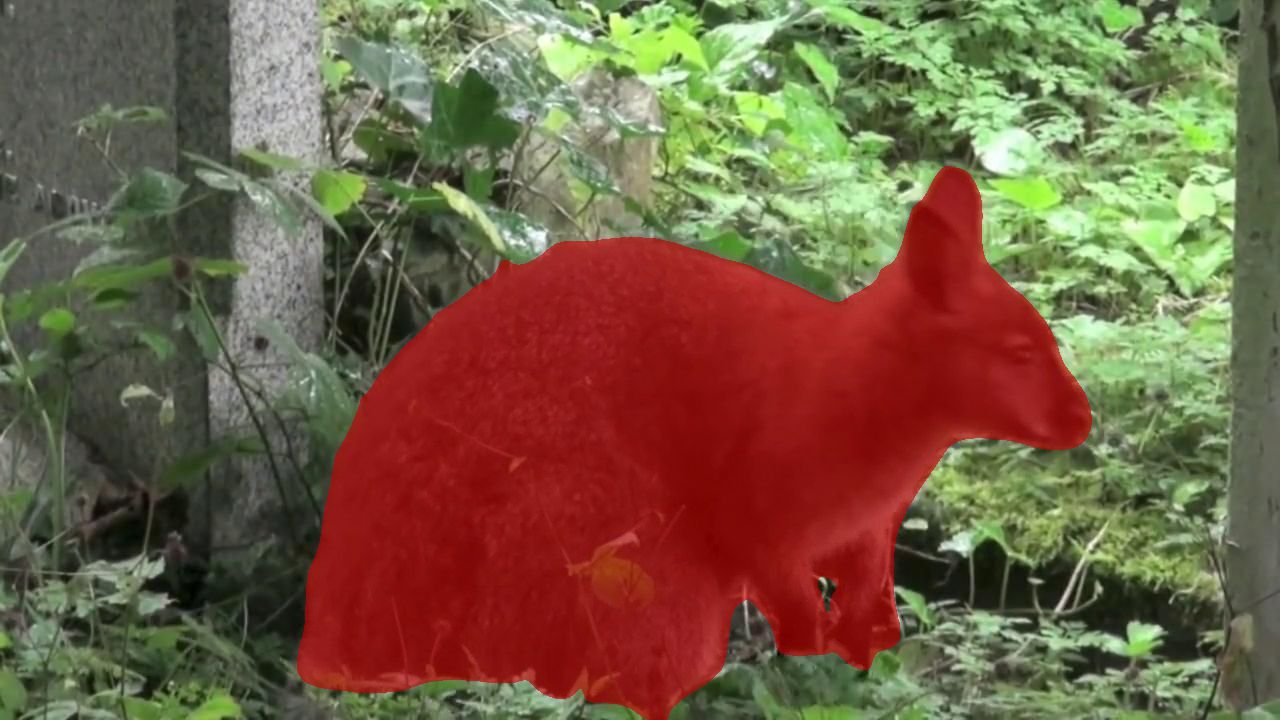} &
\image{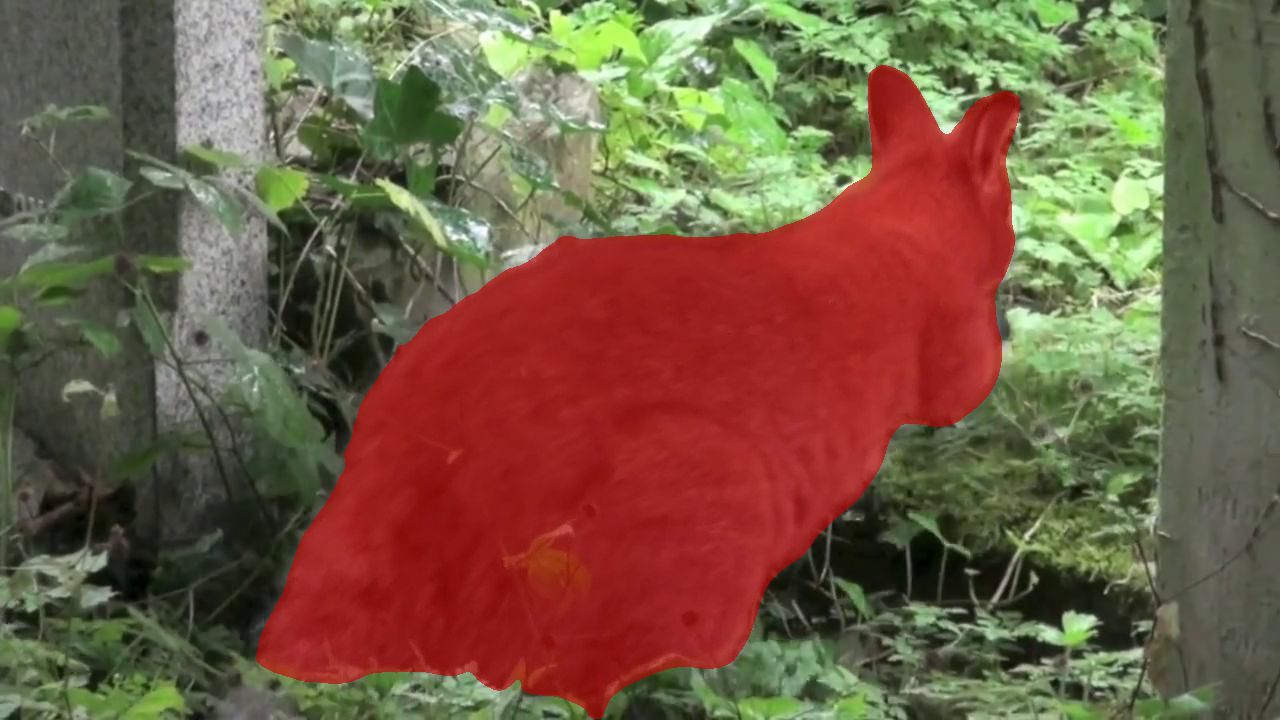} &
\image{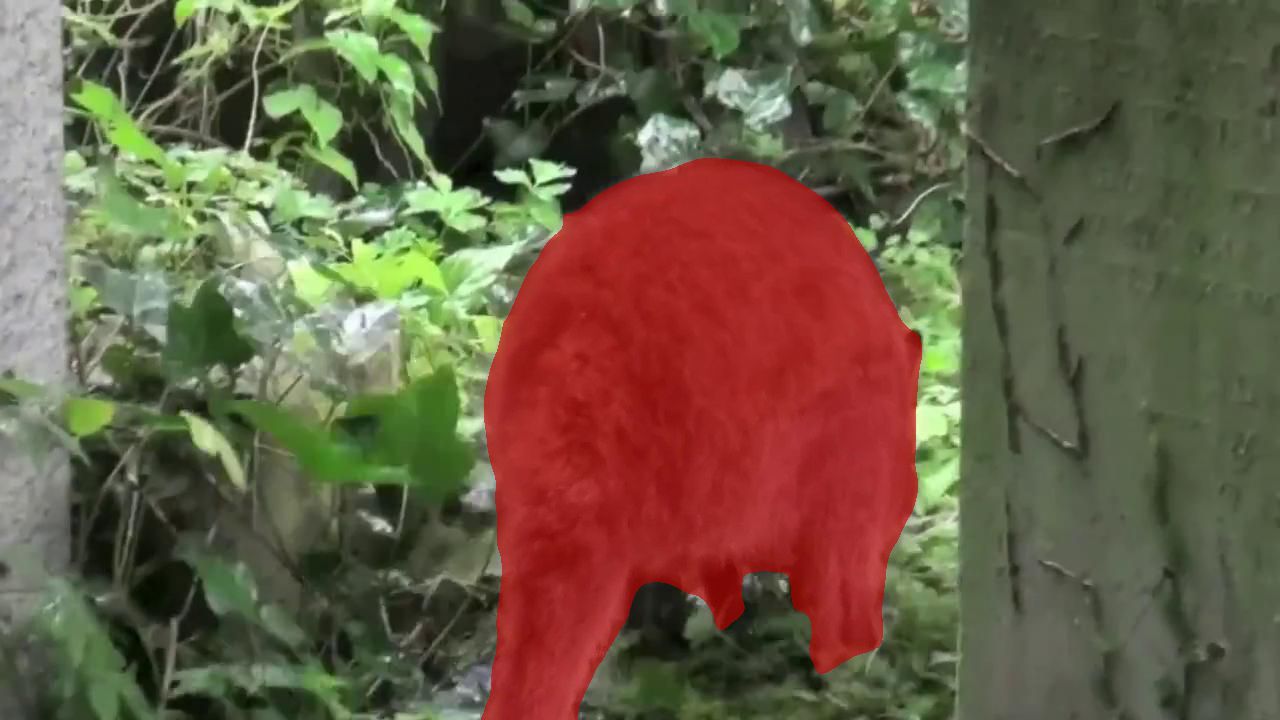} \\
\image{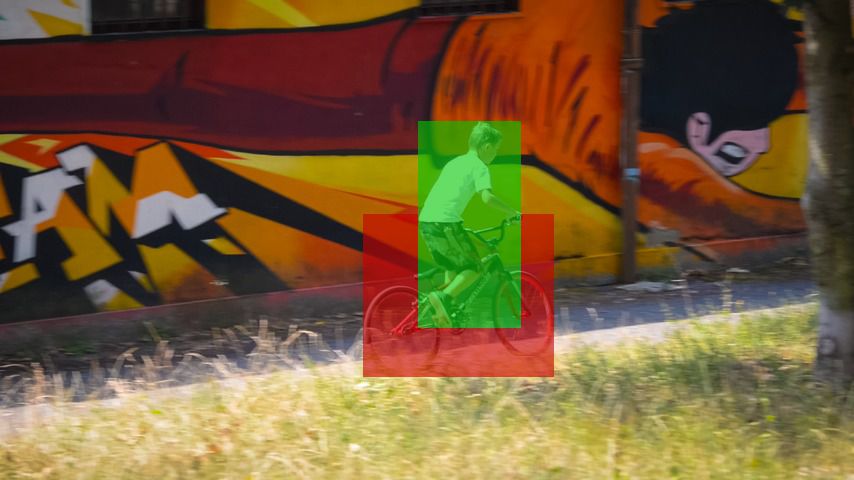} &
\image{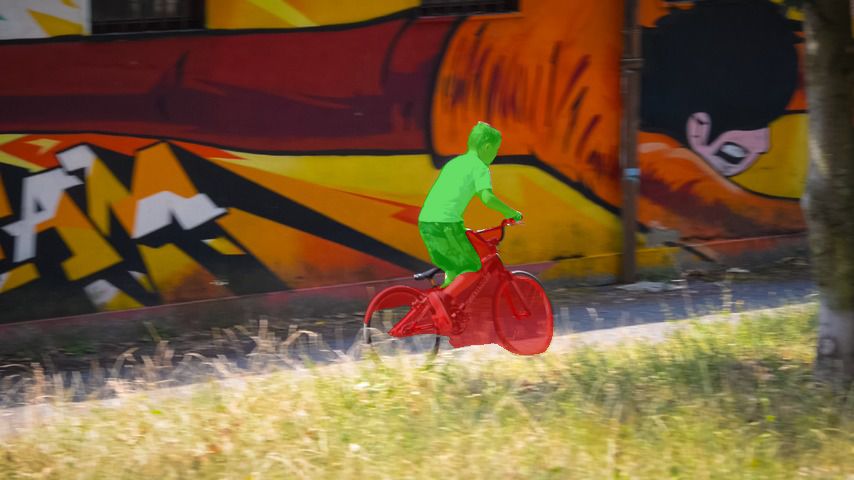} &
\image{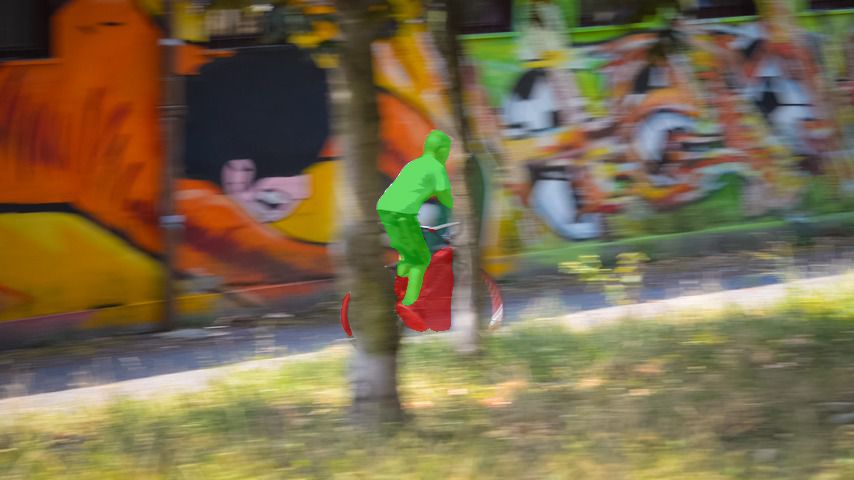} &
\image{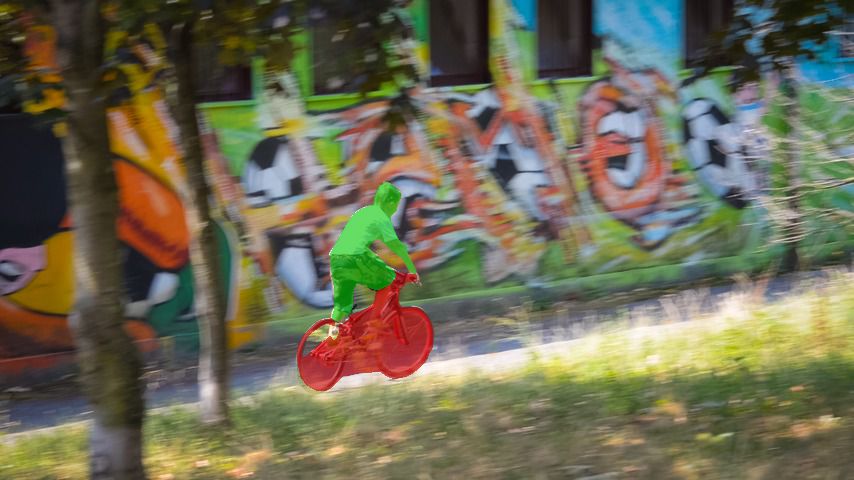} &
\image{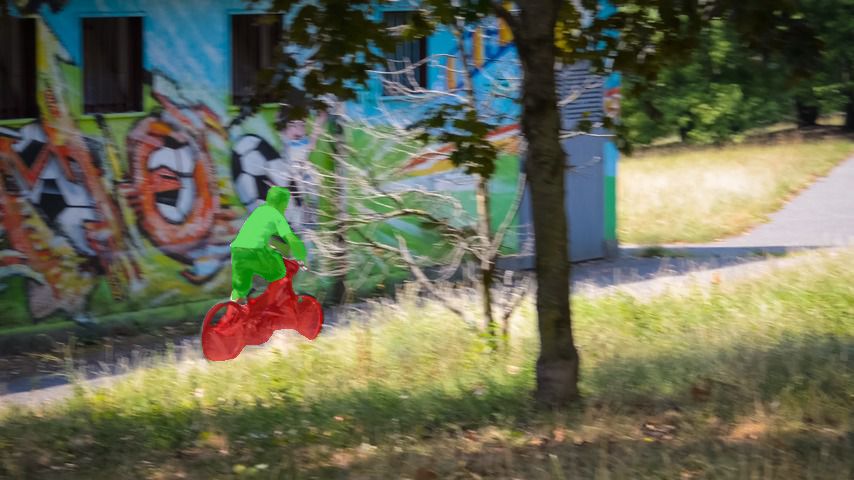} &
\image{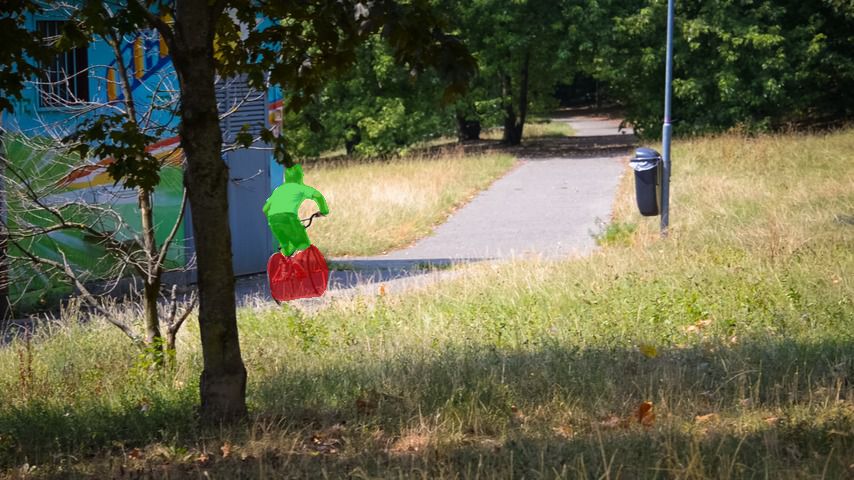} \\
\image{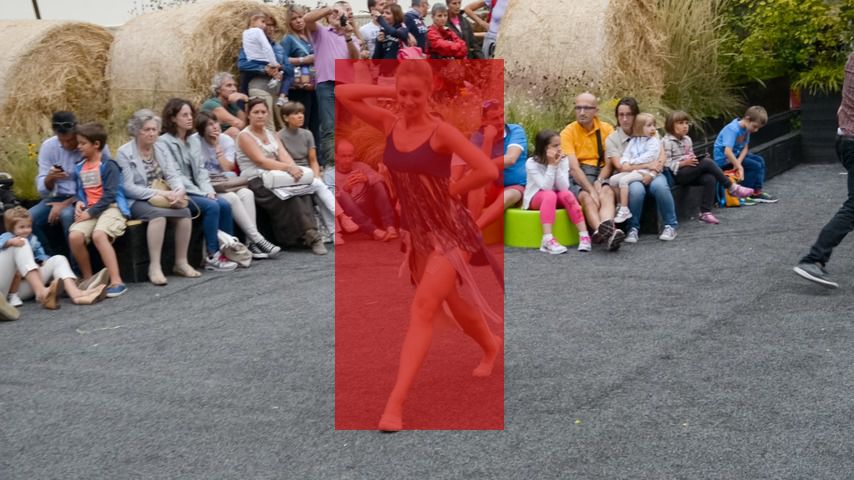} &
\image{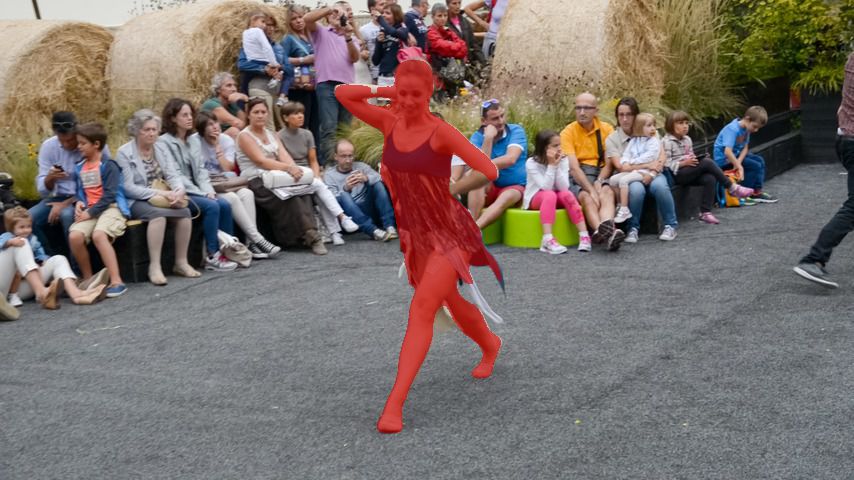} & 
\image{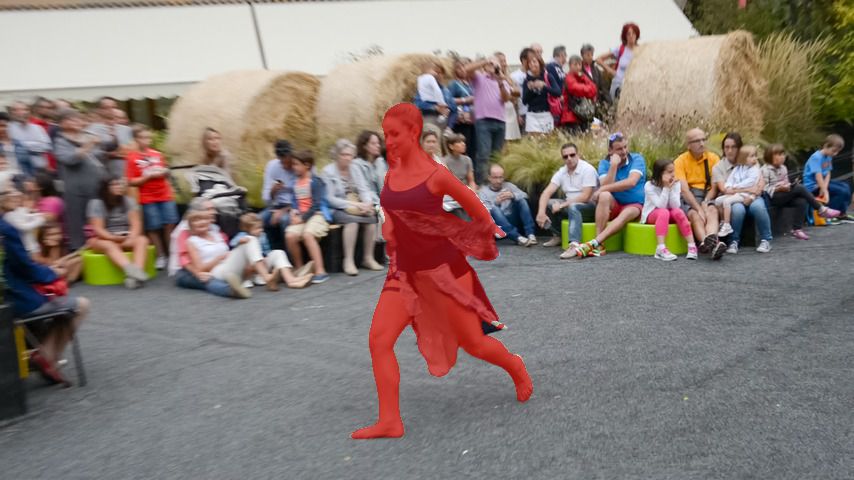} &
\image{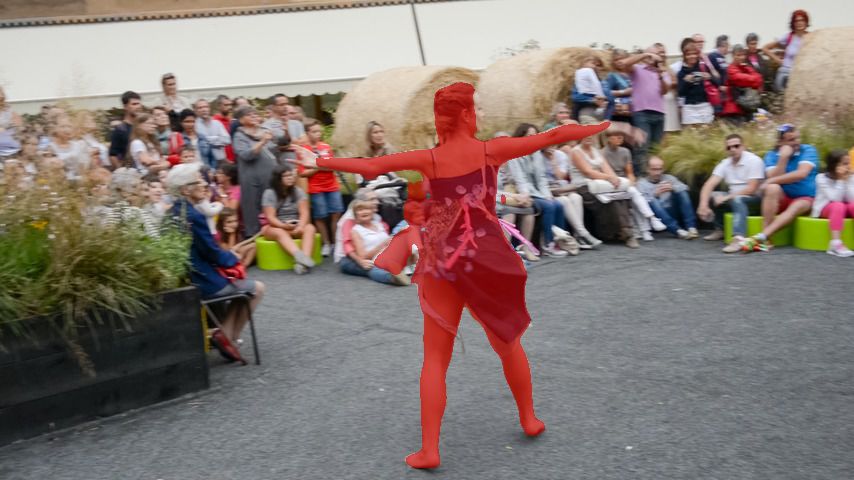} & 
\image{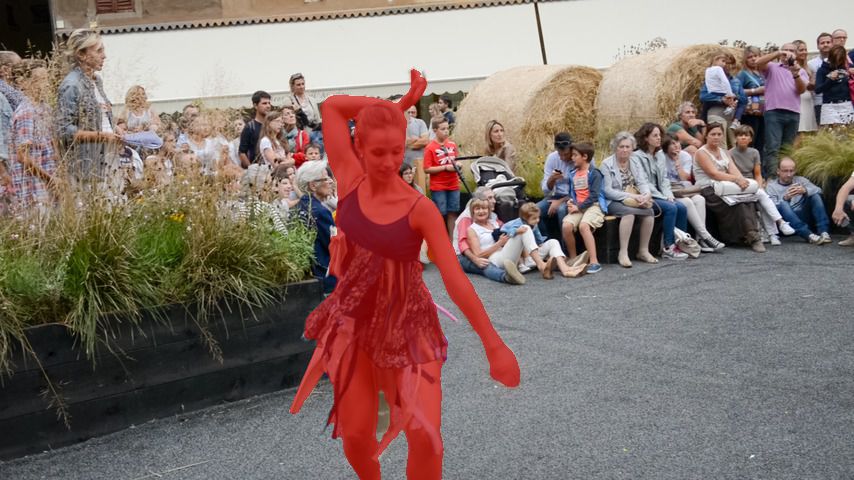} &
\image{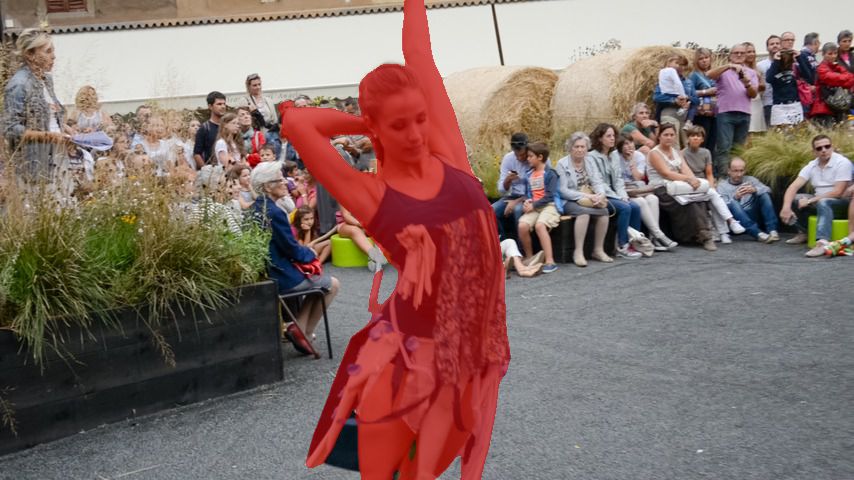} \\
\image{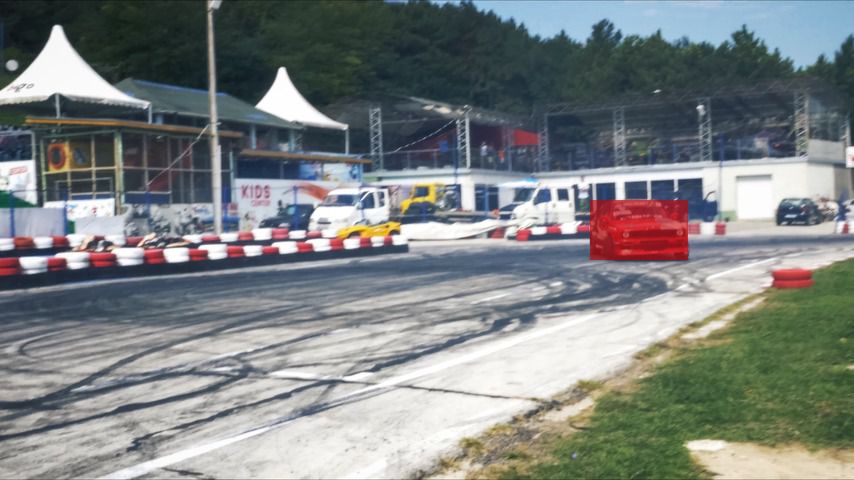} &
\image{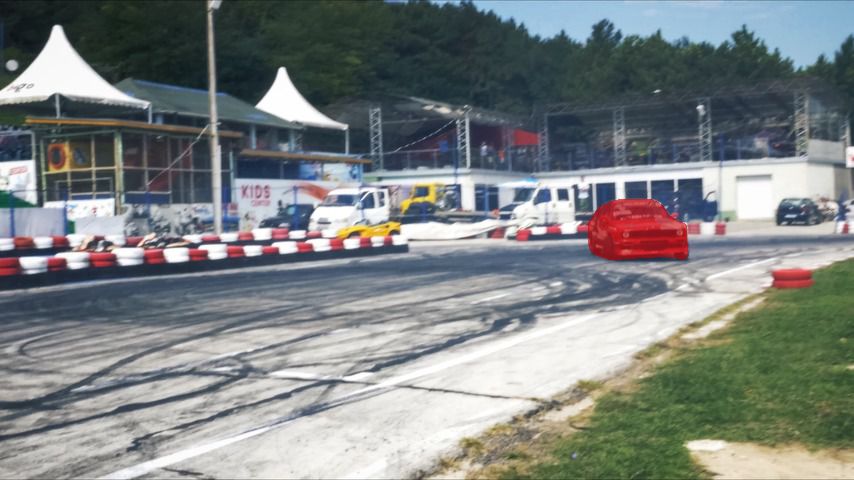}&
\image{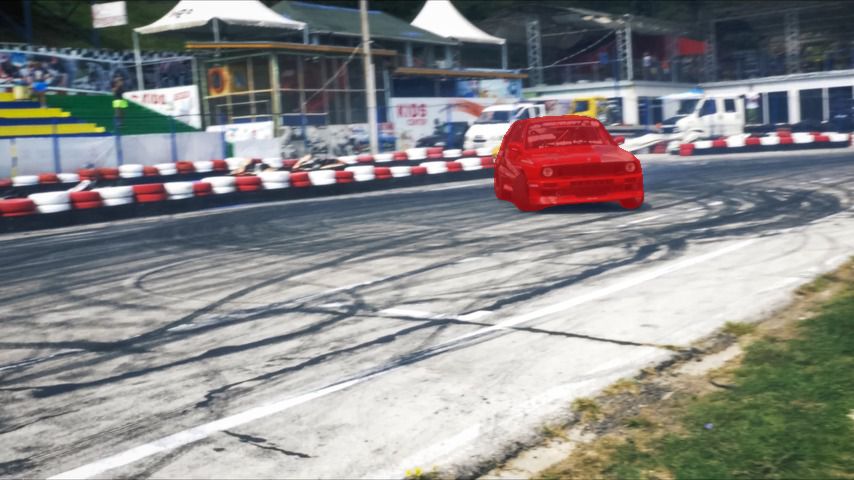}&
\image{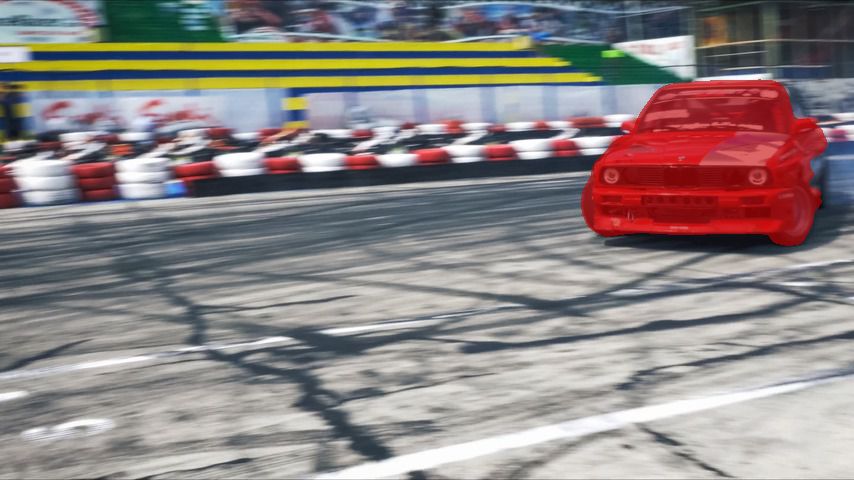}&
\image{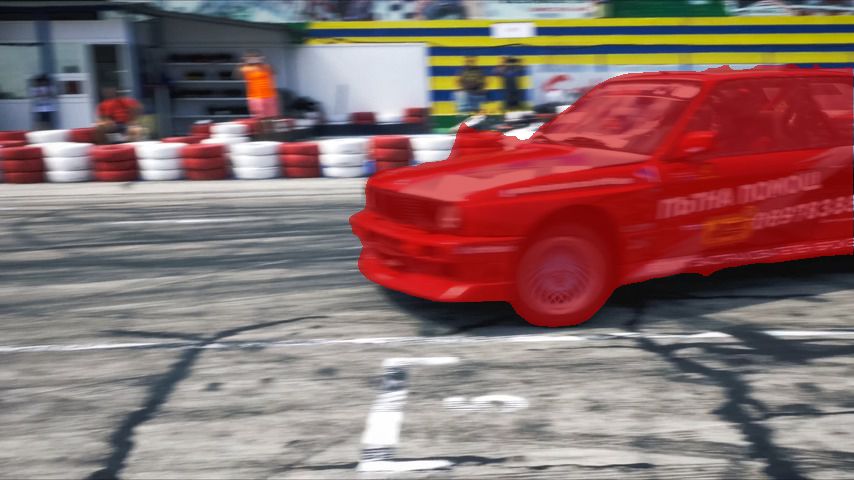}&
\image{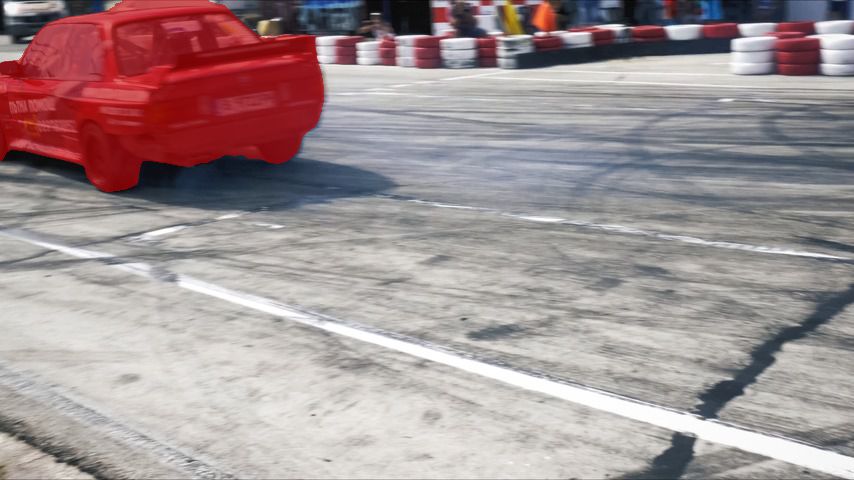}\\
\image{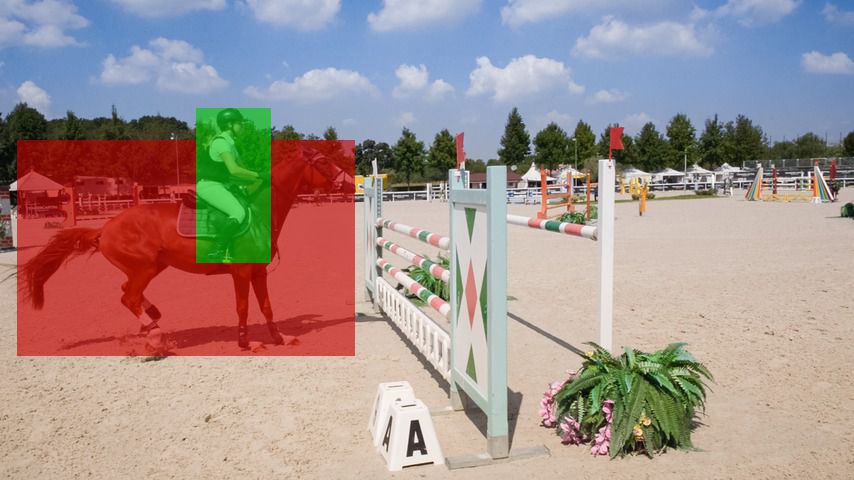} &
\image{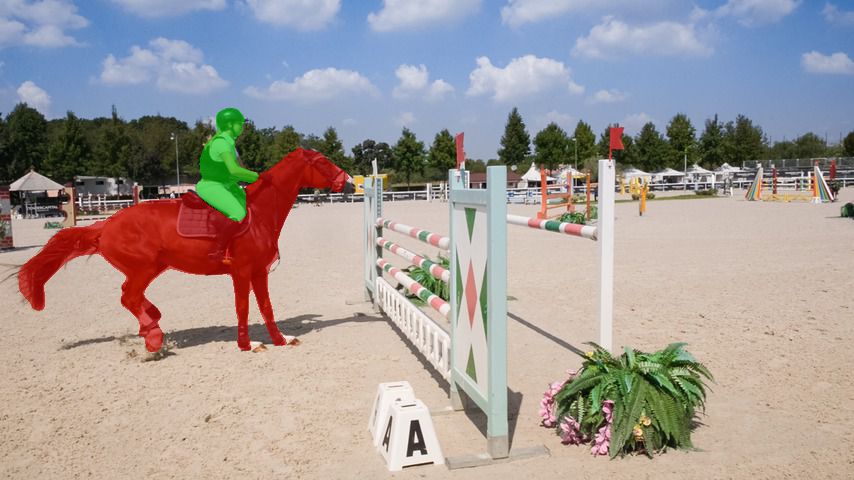}&
\image{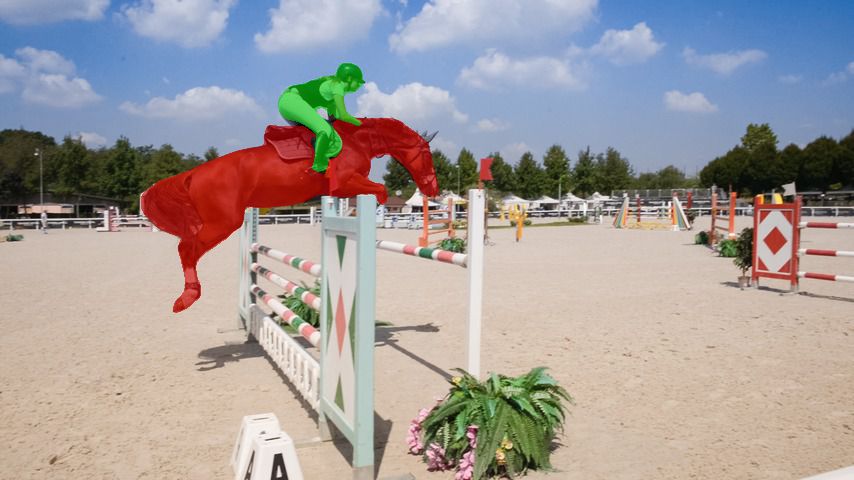}&
\image{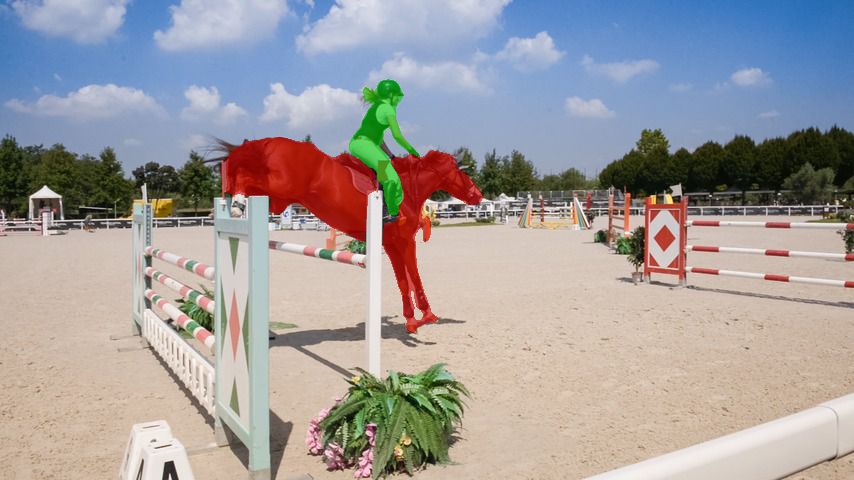}&
\image{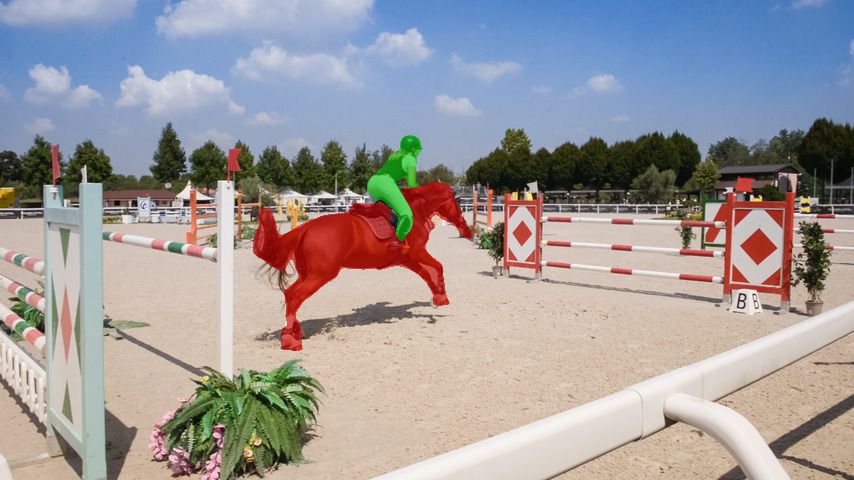}&
\image{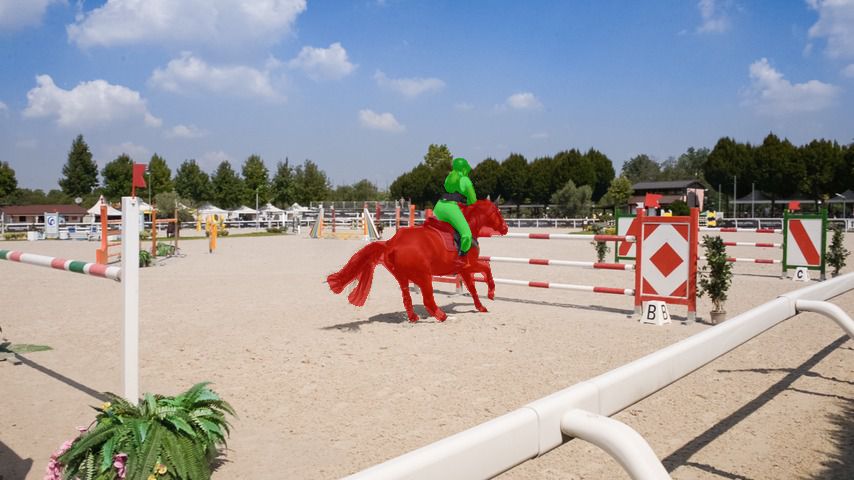}
\end{tabular}
\caption{Box initialization example results. The first column shows the initial frame with the given bounding box annotation. The second column shows the first frame segmentation, predicted by the decoder network given a mask representation generated by our box encoder module. The right four columns show predicted segmentation masks from our approach.}
\label{fig:sup-box-examples}
\end{figure}

\parsection{Bounding box initialization} In Figure~\ref{fig:sup-box-examples} we show some example outputs generated by our approach with bounding box initialization. The sequences are sampled from YouTube-VOS 2018 and DAVIS 2017 validation sets. As these examples demonstrate, our decoder network manages to segment the target in the first frame (second column), given a mask representation generated by our bounding box encoder module. This mask prediction is then used as a pseudo ground truth annotation of the initial target mask in our VOS approach. Although the predicted initial mask is less accurate than the ground truth annotation, our approach has the ability to generate high quality segmentation masks in the subsequent frames.

\clearpage

\section{Internal Validation Set} Here, we describe the internal validation set used for our ablation study in Section 4.1 of the main paper. The validation set was constructed by uniformly sampling 300 videos from the YouTube-VOS 2019 training set. The sequences included in our validation set are listed below.
\\
\noindent\footnotesize
\begin{verbatim}
d82a0aa15b, 691a111e7c, 97ab569ff3, d4a607ad81, f46c364dca, 4743bb84a7,  
1295e19071, 267964ee57, df59cfd91d, c557b69fbf, 927647fe08, 88f345941b, 
8ea6687ab0, 444aa274e7, ae93214fe6, b6e9ec577f, de30990a51, acb73e4297, 
6cccc985e0, ebc4ec32e6, f34a56525e, 2b351bfd7d, a43299e362, 733798921e, 
feda5ad1c2, 103f501680, da5d78b9d1, 634058dda0, 34d1b37101, 73c6ae7711, 
a8f78125b9, e1495354e4, 4fa9c30a45, c3457af795, fe3c02699d, 878a299541, 
a1193d6490, d69967143e, d6917db4be, bda224cb25, 621584cffe, 7a5f46198d,  
35195a56a1, 204a90d81f, e0de82caa7, 8c3015cccb, 4e3f346aa5, 5e418b25f9, 
4444753edd, c7bf937af5, 4da0d00b55, 48812cf33e, 35c6235b8d, 60c61cc2e5, 
9002761b41, 13ae097e20, ec193e1a01, d3987b2930, 72f04f1a38, 97e59f09fa, 
d0ab39112e, 9533fc037c, 2b88561cf2, 6c4387daf5, e1d26d35be, 0cfe974a89, 
0eefca067f, 887a93b198, 4bc8c676bb, 6f49f522ef, a9c9c1517e, 8dcfb878a8, 
1471274fa7, 53cad8e44a, 46146dfd39, 666b660284, 51e85b347b, ec3d4fac00, 
1c72b04b56, 2ba621c750, d123d674c1, bd0e9ed437, dd61d903df, 80c4a94706, 
b4d0c90bf4, 52c8ec0373, 7bc7761b8c, 25f97e926f, e72a7d7b0b, 9f913803e9, 
8bf84e7d45, a9cbf9c41b, 7abdff3086, ae13ee3d70, a68259572b, 081ae4fa44, 
8d064b29e2, 41dab05200, 6024888af8, 5110dc72c0, b0dd580a89, 2ff7f5744f, 
45c36a9eab, ec4186ce12, 72cac683e4, c2a35c1cda, 11485838c2, 5675d78833, 
55c1764e90, bfd8f6e6c9, 7ecd1f0c69, 90c7a87887, 4f414dd6e7, 211bc5d102, 
3299ae3116, 827cf4f886, 5665c024cb, 08aa2705d5, 8e1848197c, d7bb6b37a7, 
9d01f08ec6, fad633fbe1, 11ce6f452e, 644bad9729, ae3bc4a0ef, b2ce7699e3, 
f7e0c9bb83, 52c7a3d653, 7806308f33, fed208bfca, 9198cfb4ea, 8c469815cf, 
731b825695, c52bce43db, 0d2fcc0dcd, 1917b209f2, b274456ce1, d44e6acd1d, 
7e0cd25696, 8909bde9ab, 68ea4a8c3d, 69ea9c09d1, 5a4a785006, b73867d769, 
f0c34e1213, 84044f37f3, 479f5d7ef6, 3cc37fd487, f8fcb6a78c, f0ad38da27, 
d0c65e9e95, 3b6c7988f6, f9ae3d98b7, e4d4872dab, 14dae0dc93, 86a40b655d, 
4eb6fc23a2, 15617297cc, 4b67aa9ef6, 3e7d2aeb07, 4ea77bfd15, 2719b742ab, 
f04cf99ee6, 75285a7eb1, 74ef677020, c9b3a8fbda, 62d6ece152, 536096501f, 
3355e056eb, 6a48e4aea8, 04259896e2, 189ac8208a, ba98512f97, 223bd973ab, 
a3f51855c3, 8b4fb018b7, 0ea68d418b, 6d4bf200ad, c130c3fc0c, 8a31f7bca5, 
f8b4ac12f1, f85796a921, ef45ce3035, e4f8e5f46e, d5b6c6d94a, c760eeb8b3, 
0b9d012be8, 1f4ec0563d, 2df005b843, dc32a44804, 1cada35274, 4cfdd73249, 
b8f34cf72e, 53af427bb2, 1329409f2a, 1b8680f8cd, 2bbde474ef, 2f5b0c89b1, 
6693a52081, 684bcd8812, e1f14510fa, 72a810a799, 70c3e97e41, 7c4ec17eff, 
8a75ad7924, fd77828200, 53d9c45013, 968c41829e, d39934abe3, 6e1a21ba55, 
bc4f71372d, 57246af7d1, f49e4866ac, 1e1a18c45a, a14ef483ff, d92532c7b2, 
aab33f0e2a, f3325c3338, 4cf5bc3e60, c98b6fe013, 619812a1a7, f8c8de2764, 
6dd2827fbb, f277c7a6a4, 1ca240fede, 16e8599e94, b554843889, df0638b0a0, 
d664c89912, c5ab1f09c8, d38d1679e2, 31bbd0d793, b24fe36b2a, c1c830a735, 
75504539c3, a74b9ca19c, c6bb6d2d5c, 99dc8bb20b, 92c46be756, 7a626ec98d, 
0891ac2eb6, 7f54132e48, c47d551843, 4122aba5f9, 5aeb95cc7d, 8ca1af9f3c, 
4019231330, 8f320d0e09, 5851739c15, b69926d9fa, b132a53086, 135625b53d, 
05d7715782, e3e4134877, d3069da8bb, 747c44785c, 59a6459751, 5a75f7a1cf, 
63936d7de5, d301ca58cc, 9c404cac0c, 78613981ed, d072fda75b, 390c51b987, 
571ca79c71, 67cfbff9b1, 7a8b5456ca, efe5ac6901, c4571bedc8, 57a344ab1a, 
d205e3cff5, 39befd99fb, 3b23792b84, 6a5de0535f, ced7705ab2, 06ce2b51fb, 
dd415df125, 2f710f66bd, 0f6c2163de, e470345ede, 6b2261888d, 6671643f31, 
de74a601d3, f14c18cf6a, f38e5aa5b4, 57427393e9, 6da21f5c91, 738e5a0a14, 
0f2ab8b1ff, 4a4b50571c, a263ce8a87, 031ccc99b1, ab45078265, 01e64dd36a, 
e0c478f775, b5b9da5364, 72acb8cdf6, c922365dd4, df11931ffe, ad3fada9d9
\end{verbatim}

\end{document}